\documentclass[letterpaper]{article}

\usepackage[preprint]{preprint}
\usepackage{eso-pic}
\usepackage[hyphens]{url}
\usepackage{graphicx}
\usepackage{natbib}
\usepackage{caption}
\usepackage[caption=false,font=small]{subfig}
\usepackage{cuted}
\usepackage{amsmath,amssymb,amsthm}
\usepackage{booktabs}
\usepackage{xspace}
\usepackage{array}
\usepackage{arydshln}
\usepackage{algorithm}
\usepackage{algorithmic}
\usepackage[dvipsnames]{xcolor}
\usepackage{colortbl}
\usepackage{microtype}
\usepackage{tikz}
\usetikzlibrary{arrows.meta,positioning,fit,backgrounds,calc,shapes.multipart,decorations.pathreplacing}
\usepackage[most]{tcolorbox}
\tcbuselibrary{breakable,skins}
\usepackage{mathtools}
\usepackage{pifont}
\usepackage{needspace}
\usepackage{booktabs,multirow}

\usepackage{booktabs}
\usepackage{tabularx}
\usepackage{array}
\usepackage[table]{xcolor}

\definecolor{SAHighlight}{HTML}{FFF2A8}
\definecolor{SAAccent}{HTML}{087EA4}
\definecolor{LocalOrange}{HTML}{B45309}

\newcolumntype{Y}{>{\raggedright\arraybackslash}X}

\definecolor{saBlue}{HTML}{2B5D8A}
\definecolor{saTeal}{HTML}{147A76}
\definecolor{saGold}{HTML}{A96B00}
\definecolor{saRed}{HTML}{A94442}
\definecolor{saFill}{HTML}{F3F7FA}
\definecolor{saSoft}{HTML}{E7F0F8}
\definecolor{alchemyblue}{HTML}{EAF3FF}

\definecolor{saBlue}{HTML}{2F6090}     \definecolor{saBlueBd}{HTML}{E9F1F8}
\definecolor{saTeal}{HTML}{2E8B7F}     \definecolor{saTealBd}{HTML}{E1F1EE}
\definecolor{saOrange}{HTML}{CE8A2D}   \definecolor{saOrangeBd}{HTML}{FAF0DC}
\definecolor{saViolet}{HTML}{6C5BA6}   \definecolor{saVioletBd}{HTML}{EDEAF6}
\definecolor{saRed}{HTML}{BD4A47}      \definecolor{saRedBd}{HTML}{F8E9E8}
\definecolor{saGray}{HTML}{6B7177}     \definecolor{saGrayBd}{HTML}{F0F1F2}
\definecolor{saInk}{HTML}{222629}

\tikzset{
  hbox/.style 2 args={
    rectangle split, rectangle split parts=2, rectangle split draw splits=false,
    rectangle split part fill={#1,#2},
    draw=#1, line width=0.65pt, rounded corners=2.2pt,
    text=white, align=center, inner xsep=4pt, inner ysep=3pt,
    minimum width=1.55cm, text width=1.45cm,
    font=\sffamily,
  },
  core/.style ={hbox={saBlue}{saBlueBd}},
  accent/.style={hbox={saTeal}{saTealBd}},
  gate/.style  ={hbox={saOrange}{saOrangeBd}},
  audit/.style ={hbox={saViolet}{saVioletBd}},
  risk/.style  ={hbox={saRed}{saRedBd}},
  neutral/.style={hbox={saGray}{saGrayBd}},
  flow/.style ={-{Stealth[length=2.4mm,width=1.8mm]}, line width=0.9pt, draw=saBlue},
  oldflow/.style={-{Stealth[length=2.4mm,width=1.8mm]}, line width=0.85pt, draw=saRed!85},
  aux/.style  ={-{Stealth[length=1.9mm,width=1.5mm]}, line width=0.6pt, draw=saInk!42, dashed},
  curve/.style={-{Stealth[length=2.2mm,width=1.6mm]}, line width=0.8pt},
  clab/.style  ={font=\sffamily\bfseries\scriptsize, text=saInk!85},
  sub/.style  ={font=\sffamily\scriptsize, text=saInk!62, align=center},
  ptag/.style ={font=\sffamily\scriptsize\bfseries, fill=saInk!6, draw=saInk!35,
                rounded corners=1.5pt, inner xsep=4pt, inner ysep=2pt, text=saInk!80},
}

\tikzset{
  tbox/.style 2 args={hbox={#1}{#2}, text width=1.16cm, minimum width=1.22cm,
                      inner xsep=3pt, inner ysep=2.6pt},
  tcore/.style   ={tbox={saBlue}{saBlueBd}},
  taccent/.style ={tbox={saTeal}{saTealBd}},
  trisk/.style   ={tbox={saRed}{saRedBd}},
  tneutral/.style={tbox={saGray}{saGrayBd}},
  chip/.style={rounded corners=3pt, draw=#1, fill=#1!10, line width=0.6pt,
               font=\sffamily\scriptsize\bfseries, text=#1!88, inner xsep=5pt, inner ysep=3.4pt},
}

\newcounter{procedure}

\newcommand{\eg}{\textit{e.g.},\xspace}
\newcommand{\ie}{\textit{i.e.},\xspace}

\newcommand\figref[1]{\figurename~\ref{#1}}

\newcommand\tabref[1]{Table~\ref{#1}}

\newcommand\equref[1]{Eq.~(\ref{#1})}
\newcommand\appref[1]{Appendix~\ref{#1}}

\newcommand{\fakeparagraph}[1]{\vspace{1mm}\noindent\textbf{#1.}}

\definecolor{AuditBlue}{HTML}{2D5F87}
\definecolor{AuditBlueBG}{HTML}{EFF5F9}
\definecolor{AuditOrange}{HTML}{B45F06}
\definecolor{AuditGrayBG}{HTML}{F7F7F7}

\newtcolorbox{auditcard}[2][]{%
  enhanced,
  width=\linewidth,
  height=32mm,
  valign=top,
  colback=AuditGrayBG,
  colframe=black!28,
  colbacktitle=black!5,
  coltitle=black,
  boxrule=0.45pt,
  titlerule=0.35pt,
  arc=1.0mm,
  left=1.3mm,
  right=1.3mm,
  top=0.9mm,
  bottom=0.8mm,
  before skip=0pt,
  after skip=0pt,
  fonttitle=\bfseries\footnotesize,
  fontupper=\footnotesize,
  title={#2},
  #1
}

\newcolumntype{C}{>{\centering\arraybackslash}p{0.052\textwidth}}

\newcommand{\sysname}{\textsc{SkillAlchemy}}

\definecolor{dc1}{HTML}{B4FFA4}
\definecolor{dc2}{HTML}{FBFB0C}
\definecolor{dc3}{HTML}{F57171}
\definecolor{dc4}{HTML}{F7F6F6}

\newtcolorbox{examplecard}[1][]{
  colback=gray!3, colframe=gray!50!black,
  colbacktitle=gray!18, coltitle=black,
  boxrule=0.5pt, titlerule=0pt, arc=2pt,
  left=6pt, right=6pt, top=5pt, bottom=5pt,
  fonttitle=\bfseries\small,
  title={#1},
}

\ifodd 1
\newcommand{\TODO}[1]{\textbf{\color{red}{TODO: #1} }}

\else
\newcommand{\TODO}[1]{}

\fi

\graphicspath{{figures/}}
\ifdefined\BuildSupplementary
\title{\sysname{}: Supplementary Material}
\else
\title{\sysname: Open-World Agent Skill Creation}
\fi
\author{
    Hengjun Wang\textsuperscript{\rm 1},
    Shuyue Wei\textsuperscript{\rm 2*},
    Boyi Liu\textsuperscript{\rm 1},
    Jun Yang\textsuperscript{\rm 3},
    Yongxin Tong\textsuperscript{\rm 1*}
}
\affiliations{
    \textsuperscript{\rm 1}State Key Laboratory of Complex \& Critical Software Environment, Beihang University, Beijing, China\\
    \textsuperscript{\rm 2}Joint SDU--NTU Centre for Artificial Intelligence Research (C-FAIR), Shandong University, Jinan, China\\
    \textsuperscript{\rm 3}School of Automation, Northwestern Polytechnical University, Xi'an, China
}

\newcommand{\placecorrespondenceblock}{%
    \AddToShipoutPictureFG*{%
        \AtPageLowerLeft{%
            \hspace*{0.75in}%
            \raisebox{0.62in}[0pt][0pt]{%
                \parbox[b]{\columnwidth}{%
                    \footnotesize\raggedright
                    \rule{5pc}{0.4pt}\\[-0.25em]
                    \textsuperscript{*}Corresponding authors:
                    \texttt{weishuyue@sdu.edu.cn},
                    \texttt{yxtong@buaa.edu.cn}%
                }%
            }%
        }%
    }%
}

\date{}

\begin{document}

\maketitle
\placecorrespondenceblock
\ifdefined\BuildSupplementary

\appendix
\setcounter{table}{0}
\setcounter{figure}{0}
\setcounter{equation}{0}
\setcounter{principle}{0}
\setcounter{observation}{0}
\setcounter{proposition}{0}
\setcounter{procedure}{0}
\renewcommand{\thetable}{A.\arabic{table}}
\renewcommand{\thefigure}{A.\arabic{figure}}
\renewcommand{\theequation}{A.\arabic{equation}}
\renewcommand{\theprinciple}{A.\arabic{principle}}
\renewcommand{\theobservation}{A.\arabic{observation}}
\renewcommand{\theproposition}{A.\arabic{proposition}}

This appendix provides the implementation and analysis details supporting the main paper.
Appendix~\ref{app:exp-details} documents the evaluation protocol.
Appendix~\ref{app:qualitative-analysis} analyzes generated skill packages.
Appendix~\ref{app:skill-grammar} describes how the skill grammar is derived and used.
Appendix~\ref{app:case-study} and Appendix~\ref{app:research-planning-case} present matched execution and process-level case studies.
Finally, Appendix~\ref{app:representative-skill-files} reproduces representative \texttt{SKILL.md} files for inspection.

\section{Experimental Protocol}
\label{app:analysis-records}
\label{app:exp-details}

This section documents baseline reproduction, evaluation isolation, and the shared runtime configuration behind the reported experimental results.

\subsection{Baseline Implementation Details}
\label{subsubsec:baseline-detail}
\fakeparagraph{OpenSkill}
We reproduce the main OpenSkill procedure described in the main paper.
The method first reads the visible task information and performs a creation search $D$ and an independent verification search $D_v$.
It then plans and creates the skill and evaluates it with a virtual verifier.
After a failed virtual test, the method determines whether the failure arises from a skill defect or a knowledge gap and refines the skill for up to three rounds.
We follow the reported settings of at most four skills, three refinement rounds, three targeted searches, a pass threshold of 1.0, and at most 60 virtual-verifier tests.
For each configuration, we use the corresponding main-paper model for skill creation and downstream execution.

\fakeparagraph{MUSE-Autoskill}
We reproduce MUSE-Autoskill following the skill-distillation procedure described in the main paper.
For each task, the method distills reusable procedures, key operations, validation steps, and common errors into a task-level skill, which is then installed without further modification for downstream evaluation.

\subsection{Evaluation Isolation Protocol}
\label{subsec:evaluation-protocol}
\fakeparagraph{Isolation During Skill Creation}
Our evaluation separates two stages.
In the \emph{skill-creation stage}, a skill-creation method takes a task brief together with the source materials permitted by its protocol and produces an installable skill package, \ie a \texttt{SKILL.md} artifact and its bundled resources.
In the \emph{evaluation stage}, the produced skill is loaded by a fresh downstream agent that attempts the task, and a benchmark verifier scores the resulting submission.
\textbf{Evaluation-only assets are kept separate from skill creation.}

Table~\ref{tab:isolation} distinguishes the four evaluation-only asset types from the information available during skill creation.
The SkillsBench runtime does not expose held-out inputs, oracle artifacts, or verifier logic to the evaluated agent.
Our creation protocol excludes all four asset types from automated skill-creation methods.
The Human-Curated Skill is mounted only for its evaluation condition and is never provided as a source during creation.

\begin{table}[H]
\centering
\footnotesize
\renewcommand{\arraystretch}{1.02}
\setlength{\tabcolsep}{3pt}
\begin{tabular}{@{}>{\raggedright\arraybackslash}m{0.37\columnwidth} m{0.57\columnwidth}@{}}
\toprule
\textbf{Evaluation-only asset} & \textbf{Role in evaluation and isolation} \\
\midrule
Held-Out Task inputs & Task inputs reserved for downstream evaluation. Creators receive only the visible task description and context. Kept separate by the benchmark runtime and our protocol. \\
Oracle/Reference Assets & Reference solutions, gold outputs, or expected artifacts used to establish task success. Isolated by the benchmark runtime and our protocol. \\
Verifier Implementations & Scoring Programs whose task-specific criteria map a submission to a pass/fail signal. Isolated by the benchmark runtime and our protocol. \\
Human-Curated Skill & Human-Authored Packages used only in the curated evaluation condition, not as creation sources. \\
\bottomrule
\end{tabular}
\caption{\textbf{Evaluation-only assets excluded during skill creation.} The benchmark runtime isolates evaluation-time state from the downstream agent, while our protocol applies the stated exclusions to all skill-creation methods.}
\label{tab:isolation}
\end{table}

For Web-enabled creation, we exclude SkillsBench-related pages from retrieval.
We apply this exclusion consistently across all automated skill-creation methods.
This keeps Web access under a common source policy and preserves the same evaluation setup across methods.


\subsection{Runtime Configuration}
\label{app:runtime-config}
Model endpoints are supplied through our API layer using the exact identifiers \url{gpt5.5-2026.4.23}, \url{claude-opus-4.8}, and \url{deepseek-v4-pro}.
The \url{deepseek-v4-pro} endpoint serves the preview checkpoint, which we denote as DeepSeek-V4-Pro-Preview in the appendix text.
Agent and adapter versions are pinned by the SkillsBench runtime: Codex v0.128.0 with codex-acp v0.0.45, and Claude Code v2.1.160 with claude-agent-acp v0.40.0.
All runs use SkillsBench commit \texttt{34256d1}.
Experiments are executed on Ubuntu 22.04.5 LTS with an Intel Xeon Platinum 8360Y CPU (144 logical CPUs), 1.0 TiB of system memory, and eight NVIDIA A100-SXM4-80GB GPUs.
We set the temperature to 0.2 during skill creation, while downstream execution uses the default decoding configuration of the benchmark runner.

\section{Generated Skill Artifacts Analysis}
\label{app:qualitative-analysis}

This section analyzes how skill-creation methods distribute executable guidance and supporting material across their generated skill packages.
We compare main-file scope and package composition to determine whether the runtime-facing instructions remain focused while detailed evidence is organized in supporting resources.

\begin{figure*}[t]
\centering
\setlength{\tabcolsep}{0pt}
\begin{tabular}{@{}cccc@{}}
\multicolumn{4}{c}{%
\includegraphics[width=0.98\textwidth]{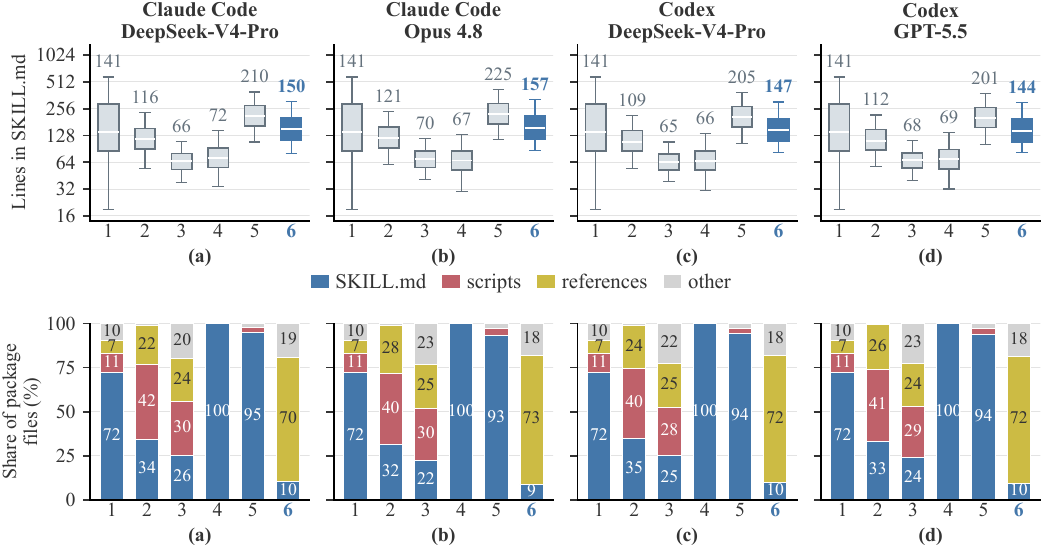}}\\[-0.25em]
\multicolumn{4}{c}{%
\footnotesize
\begin{tabular}{@{}r@{\,}l@{\qquad}r@{\,}l@{\qquad}r@{\,}l@{}}
1: & Human-Curated Skill & 2: & Anthropic Skill-Creator & 3: & OpenAI Skill-Creator \\
4: & OpenSkill & 5: & MUSE-Autoskill & 6: & SkillAlchemy
\end{tabular}}
\end{tabular}
\caption{\textbf{Skill package anatomy across four agent--model configurations.}
Columns (a)--(d) correspond to Claude Code with DeepSeek-V4-Pro-Preview, Claude Code with Opus 4.8, Codex with DeepSeek-V4-Pro-Preview, and Codex with GPT-5.5.
The upper row reports main-file length distributions, and the lower row reports the proportions of \texttt{SKILL.md}, scripts, references, and other files.}
\label{fig:skill-anatomy}
\end{figure*}

\fakeparagraph{Main-File Scope}
The upper row of Figure~\ref{fig:skill-anatomy} shows that the median top-level \texttt{SKILL.md} length is approximately 144--157 lines across configurations, although longer task-specific packages occur.
In the packaging design of \sysname{}, the top-level file retains the triggers, procedures, decision boundaries, expected outputs, and task bindings required during execution.
Detailed evidence and scoped examples not needed in the initial context are placed in package-relative reference files.

\fakeparagraph{Package Organization}
The lower row of Figure~\ref{fig:skill-anatomy} shows that the skill-creator baselines tend to package reusable operations as generated scripts, whereas \sysname{} represents them as procedures in \texttt{SKILL.md} and places supporting details in scoped references.
OpenSkill and MUSE-Autoskill instead concentrate most of their user-facing files in the main \texttt{SKILL.md}.
By acquiring information from open-world sources and organizing detailed supporting knowledge in reference files, \sysname{} preserves broad external support without placing all source-specific details in the core runtime instructions, while achieving performance comparable to human-curated skills.
Complete representative \texttt{SKILL.md} files for all six skill-bearing conditions are provided in \appref{app:representative-skill-files}.

\raggedbottom
\section{Details of Skill Grammar}
\label{app:skill-grammar}

This section expands the grammar-guided compilation introduced in Section~3.5 of the main paper.
Section~\ref{app:skill-grammar-construction} details grammar construction, Section~\ref{app:skill-grammar-specification} makes its operational rules explicit, and Section~\ref{app:skill-grammar-compilation} explains how the grammar guides package rendering.

\subsection{Skill Grammar Construction}
\label{app:skill-grammar-construction}
We derive the skill grammar from public skills indexed by \url{https://skills.sh/}, prioritizing first-party collections from Anthropic, Vercel, Microsoft, Supabase, and Remotion.
We add community-contributed skills across topic groups, remove duplicates, and retain packages with parseable metadata, nonempty instructions, and auditable source identifiers.
The resulting quality-filtered set corresponds to the public qualified skills referred to in the main paper.

We extract recurrent package-level patterns instead of task-specific content.
They cover specific triggers, executable and conditional procedures, applicability boundaries, safeguards, verifiable outputs, scoped examples, and progressive disclosure through package-relative references.
A mechanical rubric records metadata, trigger specificity, executable steps, explicit boundaries, examples, and reference sections, and retains the top-scoring portion for deriving quality-weighted patterns.
The filtered corpus identifies presentation patterns rather than certifying skill quality, and the main-paper ablation evaluates their contribution to \sysname{}.

\subsection{Operational Skill Grammar}
\label{app:skill-grammar-specification}
We use \emph{grammar} to mean a corpus-derived operational schema, not a token-level formal grammar.
Following the main paper, its role is to guide how admitted content is expressed in the final artifact.
It supplies recurrent presentation patterns for descriptions, executable sequences or conditional structures, applicability conditions, safeguards, verifiable outputs, and progressive disclosure through package-relative references.
Table~\ref{tab:skill-grammar} makes these patterns explicit.
Each component offers a small set of organization choices and guards against a recurrent skill-writing failure.
\sysname{} selects among these choices according to the admitted content and task structure rather than imposing one fixed template.

\begin{table}[t]
\centering
\small
\setlength{\tabcolsep}{3pt}
\renewcommand{\arraystretch}{1.08}
\begin{tabular}{@{}>{\raggedright\arraybackslash\bfseries}m{0.26\columnwidth}
                    >{\raggedright\arraybackslash}m{0.68\columnwidth}@{}}
\toprule
Component & \multicolumn{1}{c}{Choices and enforced property} \\
\midrule
Activation metadata
& \textit{Choices:} keyword, context, explicit, hybrid, or always-on activation. \textit{Purpose:} state what the skill does and when it should be invoked, avoiding missed or unrelated activation. \\
Workflow form
& \textit{Choices:} ordered steps, a decision tree, template filling, or operation cards. \textit{Purpose:} match the structure to procedure dependencies and expose branch conditions. \\
Executable procedure
& \textit{Form:} condition, action, observable output, and a supported failure route. \textit{Purpose:} replace vague advice with instructions an agent can execute and inspect. \\
Applicability boundary
& \textit{Choices:} source, version, capability, scope, legal, or confidence limits. \textit{Purpose:} preserve scope and prevent unsupported generalization across tasks and contexts. \\
Verification form
& \textit{Choices:} a postcondition, checklist, or artifact check. \textit{Purpose:} express an admitted verification component as a directly observable output. \\
Output contract
& \textit{Choices:} an analysis report, executable code, conversation, checklist, or task-dependent mixture. \textit{Purpose:} make the output usable by its downstream consumer. \\
Progressive disclosure
& \textit{Placement:} core instructions in \texttt{SKILL.md}, details in \texttt{references/}, routines in \texttt{scripts/}, and static resources in \texttt{assets/}. \textit{Purpose:} control initial context while keeping supporting material reachable. \\
Package organization
& \textit{Form:} complete metadata, concrete runtime instructions, and package-relative links to optional resources. \textit{Purpose:} keep the core instructions executable while supporting progressive disclosure. \\
\bottomrule
\end{tabular}
\caption{\textbf{The operational skill grammar used for package organization.}
The grammar constrains organization and execution form; it does not supply task knowledge or override the evidence-based admission process.}
\label{tab:skill-grammar}
\end{table}

\subsection{Grammar-Guided Compilation}
\label{app:skill-grammar-compilation}
During package rendering, \sysname{} uses the grammar to select an executable and inspectable presentation for the admitted content.
The selected patterns make the intended use distinguishable from nearby tasks, organize procedures as an appropriate sequence or conditional structure, preserve applicability conditions and safeguards, and place optional resources behind package-relative references.
This guidance changes only organization and presentation: it cannot add procedures, broaden their scope, or alter the supporting evidence.
Task-specific verification appears in the package only when it is an admitted component of a procedure; the grammar can express that component as a postcondition, checklist, or artifact check but does not invent one.

\Needspace{9\baselineskip}
\fakeparagraph{Illustrative Instantiation}
Suppose the admitted content requires a workbook to be recalculated after formula edits and reopened to confirm delivered values.
The grammar renders the procedure as an ordered sequence and expresses the admitted delivered-value verification as a postcondition.
It places any admitted engine-specific details in a package-relative reference when they are not needed in the initially loaded instructions.
The recalculation requirement, verification step, and engine details must still come from admitted evidence.

\fakeparagraph{Takeaways}
The grammar separates the presentation of skill-writing decisions from the admission of task-specific knowledge.
Any skill-creation method can use the same patterns to make activation explicit, procedures executable, boundaries visible, outputs verifiable, and supporting resources loadable on demand.
Its transferable contribution is therefore not a fixed template, but a compact compilation schema for turning admitted knowledge into an agent-usable package for downstream use.

\section{Case Study}
\subsection{Execution Case Study}
\label{app:case-study}

This case study uses matched runs to show why outputs that satisfy most task criteria can still fail when a specific numerical requirement is missed.

\begin{figure*}[t]
\centering
\makebox[\textwidth][c]{%
\subfloat[Matched-filter conditioning]{%
\includegraphics[width=0.485\textwidth]{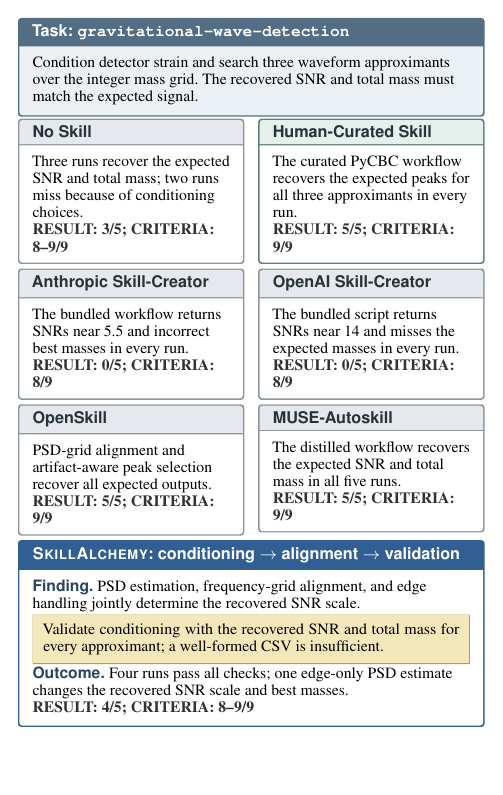}}%
\hspace{0.01\textwidth}%
\subfloat[Objective-quality audit]{%
\includegraphics[width=0.485\textwidth]{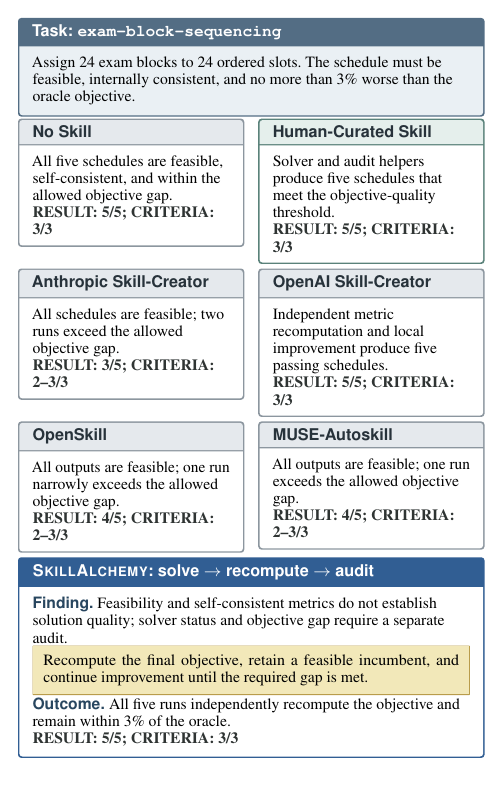}}}
\caption{\textbf{Case-level comparison across all seven conditions.} Results use Codex with GPT-5.5.
\textsc{Result} counts successes over five matched runs, and \textsc{Criteria} gives verifier-criterion coverage.
Success requires all criteria.}
\label{fig:aligned-cases}
\end{figure*}

\fakeparagraph{Matched Comparison}
SkillsBench v1.1 contains 6 easy, 53 medium, and 28 hard tasks.
We identify 33 tasks for which all seven conditions produce valid task-level and verifier-criterion results.
This matched set contains 3 easy, 21 medium, and 9 hard tasks.
The seven conditions are No Skill, Human-Curated Skill, Anthropic Skill-Creator, OpenAI Skill-Creator, OpenSkill, MUSE-Autoskill, and \sysname{}.
All case-study runs use Codex with GPT-5.5.
For each displayed case, we compare the same five runs with the same benchmark version and task inputs, then identify the verifier criterion that determines the pass/fail outcome.
Verifier-criterion coverage supports this diagnosis and is not an additional benchmark metric.
Figure~\ref{fig:aligned-cases} shows one medium task and one hard task selected from the matched set.
The pale-gold inset summarizes the task requirement highlighted by the \sysname{} artifact.

\paragraph{Case 1: Numerical conditioning.}
The medium task requires conditioning detector strain and searching an integer mass grid with three waveform approximants.
A submission must report the expected signal-to-noise ratio (SNR) and total mass for each approximant.
All seven conditions satisfy at least eight of nine verifier criteria in every matched run.
Human-Curated Skill, OpenSkill, and MUSE-Autoskill pass 5/5 runs, followed by \sysname{} at 4/5 and No Skill at 3/5, while both skill-creator conditions remain at 0/5.
The failures arise from an incorrect recovered SNR scale or best mass despite otherwise well-formed outputs.
In the unsuccessful \sysname{} run, an edge-only power spectral density estimate similarly changes the recovered SNR scale and selected masses.
This case isolates numerical conditioning rather than an artifact-format failure.

\paragraph{Case 2: Objective quality.}
The hard task assigns 24 exam blocks to 24 ordered slots and requires a feasible schedule whose objective is within 3\% of the oracle value.
All outputs in the matched runs are feasible and internally consistent.
No Skill, Human-Curated Skill, OpenAI Skill-Creator, and \sysname{} pass 5/5 runs.
Anthropic Skill-Creator passes 3/5, while OpenSkill and MUSE-Autoskill each pass 4/5 because their remaining runs exceed the allowed objective gap.
Thus, feasibility and self-consistent reported metrics do not by themselves establish that the final solution is acceptable for benchmark evaluation.

\fakeparagraph{Takeaways}
The two cases expose a shared failure pattern: structural validity and near-complete criterion coverage do not guarantee task success.
Numerical pipelines require checks on task-determining computations, while optimization tasks require objective-quality checks beyond feasibility.
Skill artifacts should therefore translate decisive task requirements into explicit, task-specific validation steps.

\subsection{Implicit-Requirement Web Search}
\label{app:research-planning-case}

This case study examines how search framing and evidence synthesis determine whether Web access improves generated skills in practice.

\paragraph{Compared workflows.}
We compare a direct creator (OpenAI Skill-Creator), a retrieval-based method (OpenSkill), and \sysname{}, which combines implicit requirement discovery with contrastive evidence acquisition.
For each workflow, we trace its questions, retrieved evidence, and final-artifact safeguards.

\paragraph{Representative task.}
\texttt{weighted-gdp-calc} requires exports, imports, and GDP for six GCC countries over 2019--2023.
The agent must add auditable formulas to an existing workbook, compute country statistics and a GDP-weighted regional mean, recalculate the file, and preserve the workbook structure and formatting.

\paragraph{Process comparison.}
The OpenAI Skill-Creator focuses on formula semantics, omitting engine compatibility and cached-value validation.
OpenSkill retrieves these risks but does not connect them to preserving and delivering the workbook.
\sysname{} turns interacting execution risks into safeguards on the delivered artifact.

\paragraph{From evidence to executable checks.}
Retrieved documentation helps only when it becomes checks on the submitted artifact.
Function semantics do not ensure engine compatibility, recalculation, or current cached values, while separately retrieved risks do not ensure whole-workbook validation.
\sysname{} connects these concerns into one execution path that preserves structure, writes compatible formulas, recalculates, reopens the saved file, and verifies values in the final delivered file.

\fakeparagraph{Takeaways}
Queries derived from implicit requirement discovery make Web evidence actionable by converting execution risks into artifact checks.
Here, direct and broad search leave critical safeguards incomplete, whereas \sysname{} connects them to checks on the final delivered artifact.

\begin{table}[H]
\centering
\footnotesize
\setlength{\tabcolsep}{3pt}
\renewcommand{\arraystretch}{1.05}
\begin{tabular}{@{}>{\raggedright\arraybackslash}m{0.27\columnwidth}
                    >{\raggedright\arraybackslash}m{0.65\columnwidth}@{}}
\toprule
Method & Search design, evidence, and outcome \\
\midrule
OpenAI Skill-Creator
&
\textbf{Framing:} task-relevant spreadsheet functions.
\textbf{Evidence:} Microsoft function documentation and openpyxl formula-writing guidance.
\textbf{Outcome:} calculation-engine compatibility and cached-value validation remain uncovered; 0/5 runs pass. \\
\addlinespace
OpenSkill
&
\textbf{Framing:} technologies for general spreadsheet construction.
\textbf{Evidence:} workbook properties, recalculation engines, Excel functions, and IMF interfaces.
\textbf{Outcome:} the risks are retrieved but not connected to a delivered-file requirement; 0/5 runs pass. \\
\addlinespace
\sysname{}
&
\textbf{Framing:} omitted execution risks identified through implicit requirement discovery and boundary probes.
\textbf{Evidence:} formula compatibility, recalculation, cached values, array alignment, reference stability, economic semantics, and workbook preservation.
\textbf{Outcome:} the evidence becomes explicit final-workbook checks; 5/5 runs pass. \\
\bottomrule
\end{tabular}
\caption{\textbf{Search strategy, retrieved evidence, and downstream result on \texttt{weighted-gdp-calc}.}
Retrieved content is summarized from the skill-creation search records.}
\label{tab:research-planning-case}
\end{table}

\section{Examples for \texttt{SKILL.md}}
\label{app:representative-skill-files}

For all six skill-bearing conditions, we reproduce the complete runtime-facing \texttt{SKILL.md} for \texttt{glm-lake-mendota}, treating the aligned module from modular Human-Curated Skill and OpenSkill outputs as their \texttt{SKILL.md}.
All numerical checks in this section use public or creator-generated artifacts; skill creation accesses no SkillsBench held-out inputs, oracle assets, or verifier implementations.
In E.6, paths, dates, and targets bind the workflow to visible task context, while the candidate parameters are explicitly scoped to Lake Mendota evidence; none are transferable defaults.
We preserve source line numbers and procedures and transliterate symbols for pdfLaTeX.
Selected spans preserve the surrounding text while providing a reading guide.
Blue marks an \emph{explicit requirement} stated by the task, while purple marks an \emph{implicit requirement} that is necessary for successful execution but must be discovered beyond the prompt.
Orange marks a scoped \emph{local example}, and green marks \emph{generalized content} intended to transfer beyond the current instance.
A red \ding{55} identifies an \emph{over-specific} use of local evidence beyond its supported scope.
Brief assessments appear immediately after the relevant source block, while neutral gray assessments describe other limitations without introducing another annotation category.
The guide below summarizes these colors.

\definecolor{skillExplicitBg}{HTML}{E1ECF3}
\definecolor{skillExplicitFg}{HTML}{2E5F7D}
\definecolor{skillImplicitBg}{HTML}{EAE4EF}
\definecolor{skillImplicitFg}{HTML}{67447D}
\definecolor{skillLocalBg}{HTML}{F3E6D5}
\definecolor{skillLocalFg}{HTML}{8D5727}
\definecolor{skillGeneralBg}{HTML}{E3ECE1}
\definecolor{skillGeneralFg}{HTML}{467044}
\definecolor{skillOverBg}{HTML}{F1DFDC}
\definecolor{skillOverFg}{HTML}{A6483E}
\definecolor{skillNeutralBg}{HTML}{ECEEEF}
\definecolor{skillNeutralFg}{HTML}{59636B}
\definecolor{skillAnnotationInk}{HTML}{20242A}
\definecolor{skillAnnotationRule}{HTML}{AAB0B5}
\definecolor{skillAnnotationPanelBg}{HTML}{FCFCFB}

\newcommand{\SkillExplicit}[1]{%
  \colorbox{skillExplicitBg}{\textcolor{skillExplicitFg}{#1}}}
\newcommand{\SkillImplicit}[1]{%
  \colorbox{skillImplicitBg}{\textcolor{skillImplicitFg}{#1}}}
\newcommand{\SkillLocal}[1]{%
  \colorbox{skillLocalBg}{\textcolor{skillLocalFg}{#1}}}
\newcommand{\SkillGeneral}[1]{%
  \colorbox{skillGeneralBg}{\textcolor{skillGeneralFg}{#1}}}
\newcommand{\SkillOverMark}{\textcolor{skillOverFg}{\ding{55}}}

\newcommand{\SkillLegendItem}[3]{%
  \colorbox{#1}{%
    \strut\hspace{0.42em}\textcolor{#2}{\textbf{#3}}\hspace{0.42em}}}

\newcommand{\SkillAnnotatedLine}[2]{%
  \par\noindent
  \makebox[2.45em][r]{\textcolor{gray}{\scriptsize #1}}\hspace{0.55em}%
  {\ttfamily\fontsize{7}{8.1}\selectfont #2}\par
}
\newcommand{\SkillAnnotatedContinuation}[1]{%
  \par\noindent
  \makebox[2.45em][r]{}\hspace{0.55em}%
  {\ttfamily\fontsize{7}{8.1}\selectfont #1}\par
}

\newtcolorbox{SkillAnnotatedFileBox}[2]{%
  enhanced,
  breakable,
  colback=skillAnnotationPanelBg,
  colframe=skillAnnotationRule,
  boxrule=0.35pt,
  arc=0.8pt,
  left=4pt,
  right=4pt,
  top=4.5mm,
  bottom=3pt,
  before skip=4pt,
  after skip=7pt,
  title={#1},
  title after break={#1\ \textnormal{(continued)}},
  coltitle=skillAnnotationInk,
  fonttitle=\ttfamily\bfseries\small,
  colbacktitle=black!4,
  boxed title style={boxrule=0pt,arc=0pt},
  attach boxed title to top left={xshift=3pt,yshift=-1.2mm},
  label={#2},
}

\newtcolorbox{SkillAnnotationBox}[2]{%
  enhanced,
  breakable,
  width=0.93\linewidth,
  grow to left by=-0.07\linewidth,
  colback=#1!18!white,
  colframe=#1,
  boxrule=0pt,
  leftrule=1.7pt,
  arc=0.5pt,
  left=4pt,
  right=4pt,
  top=1.3pt,
  bottom=1.3pt,
  before skip=1.5pt,
  after skip=2pt,
  fontupper=\scriptsize,
  title={#2},
  coltitle=skillAnnotationInk,
  fonttitle=\bfseries\scriptsize,
  colbacktitle=#1!18!white,
  boxed title style={boxrule=0pt},
  attach title to upper={\quad},
}

\begingroup
\setlength{\fboxsep}{1.35pt}
\emergencystretch=1em
\clearpage
\onecolumn
\begin{center}
\textbf{Annotation guide.}\par
\vspace{2pt}
{\footnotesize
\SkillLegendItem{skillExplicitBg}{skillExplicitFg}{Explicit requirement}
\hspace{0.35em}
\SkillLegendItem{skillImplicitBg}{skillImplicitFg}{Implicit requirement}
\hspace{0.35em}
\SkillLegendItem{skillLocalBg}{skillLocalFg}{Local example}
\hspace{0.35em}
\SkillLegendItem{skillGeneralBg}{skillGeneralFg}{Generalized content}
\hspace{0.35em}
\SkillLegendItem{skillOverBg}{skillOverFg}{\ding{55}\ Over-specific}}\par
\end{center}
\vspace{-2pt}

\Needspace{0.24\textheight}
\subsection{Human-Curated Skill}
Useful domain heuristics, but limited task grounding and several unscoped local rules. \textcolor{gray}{(92 source lines.)}\par
\begin{SkillAnnotatedFileBox}{SKILL.md (Human-Curated Skill)}{skill:0}
\SkillAnnotatedLine{1}{---\ }
\SkillAnnotatedLine{2}{name:\ glm-calibration\ }
\SkillAnnotatedLine{3}{description:\ \SkillExplicit{Calibrate\ GLM\ parameters\ for\ water\ temperature\ simulation}.\ Use\ when\ you\ need\ to\ }
\SkillAnnotatedContinuation{adjust\ model\ parameters\ to\ minimize\ \SkillExplicit{RMSE\ between\ simulated\ and\ observed\ temperatures}.\ }
\SkillAnnotatedLine{4}{license:\ MIT\ }
\SkillAnnotatedLine{5}{---\ }
\SkillAnnotatedLine{6}{\mbox{}\ }
\SkillAnnotatedLine{7}{\#\ GLM\ Calibration\ Guide\ }
\SkillAnnotatedLine{8}{\mbox{}\ }
\SkillAnnotatedLine{9}{\#\#\ Overview\ }
\SkillAnnotatedLine{10}{\mbox{}\ }
\SkillAnnotatedLine{11}{GLM\ calibration\ involves\ adjusting\ physical\ parameters\ to\ minimize\ the\ difference\ between\ }
\SkillAnnotatedContinuation{simulated\ and\ observed\ water\ temperatures.\ The\ goal\ is\ typically\ to\ achieve\ \SkillLocal{RMSE\ \textless{}\ 2.0\textdegree{}C}.\ \SkillOverMark}
\begin{SkillAnnotationBox}{skillOverFg}{Over-specific threshold}The benchmark's 2\textdegree{}C acceptance threshold is presented as a general definition of good GLM calibration.\end{SkillAnnotationBox}
\SkillAnnotatedLine{12}{\mbox{}\ }
\SkillAnnotatedLine{13}{\#\#\ Key\ Calibration\ Parameters\ }
\SkillAnnotatedLine{14}{\mbox{}\ }
\SkillAnnotatedLine{15}{|\ Parameter\ |\ Section\ |\ Description\ |\ Default\ |\ Range\ |\ }
\SkillAnnotatedLine{16}{|-----------|---------|-------------|---------|-------|\ }
\SkillAnnotatedLine{17}{|\ \SkillLocal{\textasciigrave{}Kw\textasciigrave{}\ |\ \textasciigrave{}\&light\textasciigrave{}\ |\ Light\ extinction\ coefficient\ (m\ensuremath{^{-}}\ensuremath{^{1}})\ |\ 0.3\ |\ 0.1\ -\ 0.5}\ |\ \SkillOverMark}
\SkillAnnotatedLine{18}{|\ \SkillLocal{\textasciigrave{}coef\_mix\_hyp\textasciigrave{}\ |\ \textasciigrave{}\&mixing\textasciigrave{}\ |\ Hypolimnetic\ mixing\ coefficient\ |\ 0.5\ |\ 0.3\ -\ 0.7}\ |\ \SkillOverMark}
\SkillAnnotatedLine{19}{|\ \SkillLocal{\textasciigrave{}wind\_factor\textasciigrave{}\ |\ \textasciigrave{}\&meteorology\textasciigrave{}\ |\ Wind\ speed\ scaling\ factor\ |\ 1.0\ |\ 0.7\ -\ 1.3}\ |\ \SkillOverMark}
\SkillAnnotatedLine{20}{|\ \SkillLocal{\textasciigrave{}lw\_factor\textasciigrave{}\ |\ \textasciigrave{}\&meteorology\textasciigrave{}\ |\ Longwave\ radiation\ scaling\ |\ 1.0\ |\ 0.7\ -\ 1.3}\ |\ \SkillOverMark}
\SkillAnnotatedLine{21}{|\ \SkillLocal{\textasciigrave{}ch\textasciigrave{}\ |\ \textasciigrave{}\&meteorology\textasciigrave{}\ |\ Sensible\ heat\ transfer\ coefficient\ |\ 0.0013\ |\ 0.0005\ -\ 0.002}\ |\ \SkillOverMark}
\begin{SkillAnnotationBox}{skillLocalFg}{Unscoped local defaults}The fixed values and ranges are potentially useful examples, but the skill gives no lake, GLM configuration, or evidence boundary for them.\end{SkillAnnotationBox}
\SkillAnnotatedLine{22}{\mbox{}\ }
\SkillAnnotatedLine{23}{\#\#\ Parameter\ Effects\ }
\SkillAnnotatedLine{24}{\mbox{}\ }
\SkillAnnotatedLine{25}{|\ Parameter\ |\ Increase\ Effect\ |\ Decrease\ Effect\ |\ }
\SkillAnnotatedLine{26}{|-----------|-----------------|-----------------|\ }
\SkillAnnotatedLine{27}{|\ \SkillGeneral{\textasciigrave{}Kw\textasciigrave{}\ |\ Less\ light\ penetration,\ cooler\ deep\ water\ |\ More\ light\ penetration,\ warmer\ deep\ water}\ |\ }
\SkillAnnotatedLine{28}{|\ \SkillGeneral{\textasciigrave{}coef\_mix\_hyp\textasciigrave{}\ |\ More\ deep\ mixing,\ weaker\ stratification\ |\ Less\ mixing,\ stronger\ stratification}\ }
\SkillAnnotatedContinuation{|\ }
\SkillAnnotatedLine{29}{|\ \SkillGeneral{\textasciigrave{}wind\_factor\textasciigrave{}\ |\ More\ surface\ mixing\ |\ Less\ surface\ mixing}\ |\ }
\SkillAnnotatedLine{30}{|\ \SkillGeneral{\textasciigrave{}lw\_factor\textasciigrave{}\ |\ More\ heat\ input\ |\ Less\ heat\ input}\ |\ }
\SkillAnnotatedLine{31}{|\ \SkillGeneral{\textasciigrave{}ch\textasciigrave{}\ |\ More\ sensible\ heat\ exchange\ |\ Less\ heat\ exchange}\ |\ }
\begin{SkillAnnotationBox}{skillGeneralFg}{Reusable abstraction}Parameter-effect relations transfer more safely than fixed values and can support residual-driven calibration.\end{SkillAnnotationBox}
\SkillAnnotatedLine{32}{\mbox{}\ }
\SkillAnnotatedLine{33}{\#\#\ Calibration\ with\ Optimization\ }
\SkillAnnotatedLine{34}{\textasciigrave{}\textasciigrave{}\textasciigrave{}python\ }
\SkillAnnotatedLine{35}{from\ scipy.optimize\ import\ minimize\ }
\SkillAnnotatedLine{36}{\mbox{}\ }
\SkillAnnotatedLine{37}{def\ objective(x):\ }
\SkillAnnotatedLine{38}{\ \ \ \ \SkillLocal{Kw,\ coef\_mix\_hyp,\ wind\_factor,\ lw\_factor,\ ch}\ =\ x\ }
\SkillAnnotatedLine{39}{\mbox{}\ }
\SkillAnnotatedLine{40}{\ \ \ \ \#\ Modify\ parameters\ }
\SkillAnnotatedLine{41}{\ \ \ \ params\ =\ \{\ }
\SkillAnnotatedLine{42}{\ \ \ \ \ \ \ \ 'Kw':\ round(Kw,\ 4),\ }
\SkillAnnotatedLine{43}{\ \ \ \ \ \ \ \ 'coef\_mix\_hyp':\ round(coef\_mix\_hyp,\ 4),\ }
\SkillAnnotatedLine{44}{\ \ \ \ \ \ \ \ 'wind\_factor':\ round(wind\_factor,\ 4),\ }
\SkillAnnotatedLine{45}{\ \ \ \ \ \ \ \ 'lw\_factor':\ round(lw\_factor,\ 4),\ }
\SkillAnnotatedLine{46}{\ \ \ \ \ \ \ \ 'ch':\ round(ch,\ 6)\ }
\SkillAnnotatedLine{47}{\ \ \ \ \}\ }
\SkillAnnotatedLine{48}{\ \ \ \ modify\_nml('glm3.nml',\ params)\ }
\SkillAnnotatedLine{49}{\mbox{}\ }
\SkillAnnotatedLine{50}{\ \ \ \ \#\ Run\ GLM\ }
\SkillAnnotatedLine{51}{\ \ \ \ subprocess.run(['glm'],\ capture\_output=True)\ }
\SkillAnnotatedLine{52}{\mbox{}\ }
\SkillAnnotatedLine{53}{\ \ \ \ \#\ Calculate\ RMSE\ }
\SkillAnnotatedLine{54}{\ \ \ \ rmse\ =\ calculate\_rmse(sim\_df,\ obs\_df)\ }
\SkillAnnotatedLine{55}{\ \ \ \ return\ rmse\ }
\begin{SkillAnnotationBox}{skillNeutralFg}{Implementation gap}The displayed code is not directly executable: modify\_nml, calculate\_rmse, sim\_df, and obs\_df are undefined.\end{SkillAnnotationBox}
\SkillAnnotatedLine{56}{\mbox{}\ }
\SkillAnnotatedLine{57}{\#\ \SkillLocal{Initial\ values\ (defaults)}\ }
\SkillAnnotatedLine{58}{\SkillLocal{x0\ =\ [0.3,\ 0.5,\ 1.0,\ 1.0,\ 0.0013]}\ }
\SkillAnnotatedLine{59}{\mbox{}\ }
\SkillAnnotatedLine{60}{\#\ Run\ optimization\ }
\SkillAnnotatedLine{61}{result\ =\ minimize(\ }
\SkillAnnotatedLine{62}{\ \ \ \ objective,\ }
\SkillAnnotatedLine{63}{\ \ \ \ x0,\ }
\SkillAnnotatedLine{64}{\ \ \ \ \SkillLocal{method='Nelder-Mead'},\ }
\SkillAnnotatedLine{65}{\ \ \ \ \SkillLocal{options=\{'maxiter':\ 150\}}\ }
\SkillAnnotatedLine{66}{)\ }
\SkillAnnotatedLine{67}{\textasciigrave{}\textasciigrave{}\textasciigrave{}\ }
\SkillAnnotatedLine{68}{\mbox{}\ }
\SkillAnnotatedLine{69}{\#\#\ Manual\ Calibration\ Strategy\ }
\SkillAnnotatedLine{70}{\mbox{}\ }
\SkillAnnotatedLine{71}{1.\ \SkillGeneral{Start\ with\ default\ parameters,\ run\ GLM,\ calculate\ RMSE}\ }
\SkillAnnotatedLine{72}{2.\ \SkillGeneral{Adjust\ one\ parameter\ at\ a\ time}\ }
\SkillAnnotatedLine{73}{3.\ \SkillGeneral{If\ surface\ too\ warm\ \ensuremath{\rightarrow}\ increase\ \textasciigrave{}wind\_factor\textasciigrave{}}\ }
\SkillAnnotatedLine{74}{4.\ \SkillGeneral{If\ deep\ water\ too\ warm\ \ensuremath{\rightarrow}\ increase\ \textasciigrave{}Kw\textasciigrave{}}\ }
\SkillAnnotatedLine{75}{5.\ \SkillGeneral{If\ stratification\ too\ weak\ \ensuremath{\rightarrow}\ decrease\ \textasciigrave{}coef\_mix\_hyp\textasciigrave{}}\ }
\SkillAnnotatedLine{76}{6.\ Iterate\ until\ \SkillLocal{RMSE\ \textless{}\ 2.0\textdegree{}C}\ \SkillOverMark}
\SkillAnnotatedLine{77}{\mbox{}\ }
\SkillAnnotatedLine{78}{\#\#\ Common\ Issues\ }
\SkillAnnotatedLine{79}{\mbox{}\ }
\SkillAnnotatedLine{80}{|\ Issue\ |\ Likely\ Cause\ |\ Solution\ |\ }
\SkillAnnotatedLine{81}{|-------|--------------|----------|\ }
\SkillAnnotatedLine{82}{|\ \SkillGeneral{Surface\ too\ warm\ |\ Low\ wind\ mixing\ |\ Increase\ \textasciigrave{}wind\_factor\textasciigrave{}}\ |\ }
\SkillAnnotatedLine{83}{|\ \SkillGeneral{Deep\ water\ too\ warm\ |\ Too\ much\ light\ penetration\ |\ Increase\ \textasciigrave{}Kw\textasciigrave{}}\ |\ }
\SkillAnnotatedLine{84}{|\ \SkillGeneral{Weak\ stratification\ |\ Too\ much\ mixing\ |\ Decrease\ \textasciigrave{}coef\_mix\_hyp\textasciigrave{}}\ |\ }
\SkillAnnotatedLine{85}{|\ \SkillGeneral{Overall\ warm\ bias\ |\ Heat\ budget\ too\ high\ |\ Decrease\ \textasciigrave{}lw\_factor\textasciigrave{}\ or\ \textasciigrave{}ch\textasciigrave{}}\ |\ }
\SkillAnnotatedLine{86}{\mbox{}\ }
\SkillAnnotatedLine{87}{\#\#\ Best\ Practices\ }
\SkillAnnotatedLine{88}{\mbox{}\ }
\SkillAnnotatedLine{89}{-\ \SkillGeneral{Change\ one\ parameter\ at\ a\ time\ when\ manually\ calibrating}\ }
\SkillAnnotatedLine{90}{-\ \SkillGeneral{Keep\ parameters\ within\ physical\ ranges}\ }
\SkillAnnotatedLine{91}{-\ \SkillGeneral{Use\ optimization\ for\ fine-tuning\ after\ manual\ adjustment}\ }
\SkillAnnotatedLine{92}{-\ Target\ \SkillLocal{RMSE\ \textless{}\ 2.0\textdegree{}C\ for\ good\ calibration}\ \SkillOverMark}
\begin{SkillAnnotationBox}{skillOverFg}{Repeated local rule}The same task-specific threshold is reused as both a universal stopping rule and a universal quality judgment.\end{SkillAnnotationBox}
\end{SkillAnnotatedFileBox}
\Needspace{0.24\textheight}
\subsection{Anthropic Skill-Creator}
Task-aware and operational, but one coordinate policy selects semantics by score. \textcolor{gray}{(114 source lines.)}\par
\begin{SkillAnnotatedFileBox}{SKILL.md (Anthropic Skill-Creator)}{skill:1}
\SkillAnnotatedLine{1}{---\ }
\SkillAnnotatedLine{2}{name:\ glm-lake-mendota-skill\ }
\SkillAnnotatedLine{3}{description:\ Use\ this\ skill\ whenever\ a\ task\ asks\ you\ to\ run,\ calibrate,\ or\ evaluate\ the\ General\ }
\SkillAnnotatedContinuation{Lake\ Model\ for\ \SkillExplicit{Lake\ Mendota\ temperature\ simulations}\ using\ glm3.nml,\ bcs\ forcing\ CSVs,\ }
\SkillAnnotatedContinuation{field\_temp\_oxy.csv\ observations,\ and\ an\ \SkillExplicit{RMSE\ target}.\ It\ guides\ GLM\ execution,\ NetCDF\ temperature\ }
\SkillAnnotatedContinuation{scoring,\ and\ conservative\ parameter\ iteration\ without\ hardcoding\ a\ solution.\ }
\SkillAnnotatedLine{4}{---\ }
\SkillAnnotatedLine{5}{\mbox{}\ }
\SkillAnnotatedLine{6}{\#\ GLM\ Lake\ Mendota\ Solver\ Workflow\ }
\SkillAnnotatedLine{7}{\mbox{}\ }
\SkillAnnotatedLine{8}{This\ skill\ helps\ solve\ \SkillExplicit{Lake\ Mendota\ General\ Lake\ Model\ tasks}\ where\ the\ final\ deliverables\ are:\ }
\SkillAnnotatedLine{9}{\mbox{}\ }
\SkillAnnotatedLine{10}{-\ \SkillExplicit{\textasciigrave{}/root/output/output.nc\textasciigrave{}}\ }
\SkillAnnotatedLine{11}{-\ \SkillExplicit{final\ calibrated\ parameters\ saved\ in\ \textasciigrave{}/root/glm3.nml\textasciigrave{}}\ }
\SkillAnnotatedLine{12}{-\ \SkillExplicit{vertical\ water\ temperature\ RMSE}\ against\ \textasciigrave{}/root/field\_temp\_oxy.csv\textasciigrave{}\ below\ the\ requested\ threshold\ }
\SkillAnnotatedLine{13}{\mbox{}\ }
\SkillAnnotatedLine{14}{Do\ not\ treat\ this\ skill\ as\ a\ reference\ solution.\ It\ is\ a\ \SkillGeneral{reusable\ workflow,\ scoring\ harness,\ and}\ }
\SkillAnnotatedContinuation{\SkillGeneral{calibration\ checklist}.\ You\ still\ need\ to\ run\ GLM,\ inspect\ the\ outputs,\ and\ choose\ parameters\ based\ }
\SkillAnnotatedContinuation{on\ the\ current\ sandbox.\ }
\SkillAnnotatedLine{15}{\mbox{}\ }
\SkillAnnotatedLine{16}{\#\#\ First\ Moves\ }
\SkillAnnotatedLine{17}{\mbox{}\ }
\SkillAnnotatedLine{18}{1.\ \SkillImplicit{Work\ from\ \textasciigrave{}/root\textasciigrave{}}\ unless\ the\ user\ explicitly\ stages\ the\ model\ elsewhere.\ }
\SkillAnnotatedLine{19}{2.\ Confirm\ these\ inputs\ exist:\ }
\SkillAnnotatedLine{20}{\ \ \ -\ \SkillLocal{\textasciigrave{}/usr/local/bin/glm\textasciigrave{}}\ }
\SkillAnnotatedLine{21}{\ \ \ -\ \SkillExplicit{\textasciigrave{}/root/glm3.nml\textasciigrave{}}\ }
\SkillAnnotatedLine{22}{\ \ \ -\ \SkillLocal{\textasciigrave{}/root/bcs/meteo.csv\textasciigrave{}}\ }
\SkillAnnotatedLine{23}{\ \ \ -\ \SkillLocal{\textasciigrave{}/root/bcs/yahara.csv\textasciigrave{}}\ }
\SkillAnnotatedLine{24}{\ \ \ -\ \SkillLocal{\textasciigrave{}/root/bcs/pheasant.csv\textasciigrave{}}\ }
\SkillAnnotatedLine{25}{\ \ \ -\ \SkillLocal{\textasciigrave{}/root/bcs/outflow.csv\textasciigrave{}}\ }
\SkillAnnotatedLine{26}{\ \ \ -\ \SkillExplicit{\textasciigrave{}/root/field\_temp\_oxy.csv\textasciigrave{}}\ }
\SkillAnnotatedLine{27}{3.\ Read\ \textasciigrave{}references/task\_environment.md\textasciigrave{}\ for\ the\ \SkillLocal{data\ shape\ observed\ when\ this\ skill\ was\ created}.\ }
\SkillAnnotatedLine{28}{4.\ Read\ \textasciigrave{}references/sources.md\textasciigrave{}\ before\ changing\ model\ settings;\ it\ records\ the\ public\ }
\SkillAnnotatedContinuation{documentation\ used\ to\ ground\ this\ workflow.\ }
\SkillAnnotatedLine{29}{5.\ Run\ a\ lightweight\ environment\ check:\ }
\SkillAnnotatedLine{30}{\mbox{}\ }
\SkillAnnotatedLine{31}{\textasciigrave{}\textasciigrave{}\textasciigrave{}bash\ }
\SkillAnnotatedLine{32}{python\ /path/to/this-skill/scripts/inspect\_environment.py\ --root\ /root\ }
\SkillAnnotatedLine{33}{\textasciigrave{}\textasciigrave{}\textasciigrave{}\ }
\SkillAnnotatedLine{34}{\mbox{}\ }
\SkillAnnotatedLine{35}{The\ checker\ prints\ \SkillImplicit{paths,\ CSV\ columns,\ date\ ranges,\ observation\ depth\ ranges},\ and\ the\ key\ }
\SkillAnnotatedContinuation{\textasciigrave{}glm3.nml\textasciigrave{}\ blocks.\ It\ does\ not\ run\ a\ simulation.\ }
\begin{SkillAnnotationBox}{skillImplicitFg}{Environment requirements}The task names inputs, but it does not state that their schema, coverage, depth range, and working directory must be validated before calibration.\end{SkillAnnotationBox}
\SkillAnnotatedLine{36}{\mbox{}\ }
\SkillAnnotatedLine{37}{\#\#\ Run\ GLM\ }
\SkillAnnotatedLine{38}{\mbox{}\ }
\SkillAnnotatedLine{39}{GLM\ reads\ \textasciigrave{}glm3.nml\textasciigrave{}\ in\ the\ working\ simulation\ directory.\ Keep\ the\ task's\ \SkillImplicit{relative\ paths}\ intact\ }
\SkillAnnotatedContinuation{unless\ a\ GLM\ log\ proves\ they\ are\ wrong.\ In\ this\ task,\ the\ configuration\ points\ at\ \textasciigrave{}bcs/...\textasciigrave{}\ CSV\ }
\SkillAnnotatedContinuation{files\ and\ \SkillExplicit{writes\ \textasciigrave{}output/output.nc\textasciigrave{}}.\ }
\SkillAnnotatedLine{40}{\mbox{}\ }
\SkillAnnotatedLine{41}{\textasciigrave{}\textasciigrave{}\textasciigrave{}bash\ }
\SkillAnnotatedLine{42}{cd\ /root\ }
\SkillAnnotatedLine{43}{mkdir\ -p\ output\ }
\SkillAnnotatedLine{44}{/usr/local/bin/glm\ }
\SkillAnnotatedLine{45}{\textasciigrave{}\textasciigrave{}\textasciigrave{}\ }
\SkillAnnotatedLine{46}{\mbox{}\ }
\SkillAnnotatedLine{47}{After\ each\ run,\ \SkillImplicit{confirm\ \textasciigrave{}/root/output/output.nc\textasciigrave{}\ exists}.\ If\ GLM\ fails,\ inspect\ the\ terminal\ log\ }
\SkillAnnotatedContinuation{first;\ most\ failures\ are\ \SkillImplicit{path,\ time-format,\ or\ malformed\ namelist\ edits}.\ }
\SkillAnnotatedLine{48}{\mbox{}\ }
\SkillAnnotatedLine{49}{\#\#\ Score\ Temperature\ RMSE\ }
\SkillAnnotatedLine{50}{\mbox{}\ }
\SkillAnnotatedLine{51}{Score\ against\ the\ observation\ points,\ not\ just\ a\ regular\ grid.\ The\ bundled\ scorer\ opens\ the\ NetCDF\ }
\SkillAnnotatedContinuation{output,\ infers\ the\ temperature\ variable\ and\ vertical\ coordinate,\ \SkillImplicit{interpolates\ to\ each\ \textasciigrave{}(datetime,}\ }
\SkillAnnotatedContinuation{\SkillImplicit{depth)\textasciigrave{}\ observation},\ and\ reports\ RMSE.\ }
\SkillAnnotatedLine{52}{\mbox{}\ }
\SkillAnnotatedLine{53}{\textasciigrave{}\textasciigrave{}\textasciigrave{}bash\ }
\SkillAnnotatedLine{54}{python\ /path/to/this-skill/scripts/score\_glm\_temperature.py\ \textbackslash{}\ }
\SkillAnnotatedLine{55}{\ \ --nc\ /root/output/output.nc\ \textbackslash{}\ }
\SkillAnnotatedLine{56}{\ \ --obs\ /root/field\_temp\_oxy.csv\ \textbackslash{}\ }
\SkillAnnotatedLine{57}{\ \ --json\ /root/output/rmse.json\ }
\SkillAnnotatedLine{58}{\textasciigrave{}\textasciigrave{}\textasciigrave{}\ }
\SkillAnnotatedLine{59}{\mbox{}\ }
\SkillAnnotatedLine{60}{The\ scorer\ tries\ \SkillLocal{both\ direct-depth\ and\ elevation-to-depth\ vertical\ interpretations}\ and\ reports\ the\ }
\SkillAnnotatedContinuation{\SkillLocal{lower-RMSE\ orientation}.\ If\ it\ cannot\ infer\ the\ output\ schema,\ inspect\ the\ dataset\ with\ \textasciigrave{}python\ -\ }
\SkillAnnotatedContinuation{\textless{}\textless{}'PY'\textasciigrave{}\ and\ \textasciigrave{}xarray.open\_dataset(...)\textasciigrave{},\ then\ rerun\ with\ explicit\ \textasciigrave{}--temp-var\textasciigrave{},\ \textasciigrave{}--time-dim\textasciigrave{},\ and\ }
\SkillAnnotatedContinuation{\textasciigrave{}--depth-dim\textasciigrave{}.\ \SkillOverMark}
\begin{SkillAnnotationBox}{skillOverFg}{Score-selected coordinate semantics}Choosing whichever vertical interpretation yields lower RMSE can hide a coordinate error. Coordinate meaning should be established from schema or source evidence.\end{SkillAnnotationBox}
\SkillAnnotatedLine{61}{\mbox{}\ }
\SkillAnnotatedLine{62}{\#\#\ Calibrate\ Conservatively\ }
\SkillAnnotatedLine{63}{\mbox{}\ }
\SkillAnnotatedLine{64}{Use\ \SkillGeneral{short,\ logged\ candidate\ runs}.\ Change\ a\ small\ set\ of\ physically\ meaningful\ parameters,\ score,\ }
\SkillAnnotatedContinuation{keep\ the\ \SkillGeneral{best\ reproducible\ candidate},\ and\ \SkillGeneral{avoid\ broad\ blind\ edits}\ that\ make\ the\ nml\ hard\ to\ trust.\ }
\SkillAnnotatedLine{65}{\mbox{}\ }
\SkillAnnotatedLine{66}{\SkillLocal{Parameters\ worth\ checking\ first\ in\ this\ task\ type}:\ }
\SkillAnnotatedLine{67}{\mbox{}\ }
\SkillAnnotatedLine{68}{-\ \SkillLocal{Surface\ forcing\ multipliers}\ in\ \textasciigrave{}\&meteorology\textasciigrave{}:\ \textasciigrave{}wind\_factor\textasciigrave{},\ \textasciigrave{}sw\_factor\textasciigrave{},\ \textasciigrave{}lw\_factor\textasciigrave{},\ }
\SkillAnnotatedContinuation{\textasciigrave{}at\_factor\textasciigrave{},\ \textasciigrave{}rh\_factor\textasciigrave{}.\ }
\SkillAnnotatedLine{69}{-\ \SkillLocal{Heat/light\ penetration}\ in\ \textasciigrave{}\&light\textasciigrave{}:\ \textasciigrave{}Kw\textasciigrave{}\ and,\ only\ if\ needed,\ \textasciigrave{}light\_extc\textasciigrave{}\ /\ \textasciigrave{}energy\_frac\textasciigrave{}.\ }
\SkillAnnotatedLine{70}{-\ \SkillLocal{Vertical\ mixing}\ in\ \textasciigrave{}\&mixing\textasciigrave{}:\ \textasciigrave{}coef\_wind\_stir\textasciigrave{},\ \textasciigrave{}coef\_mix\_conv\textasciigrave{},\ \textasciigrave{}coef\_mix\_shear\textasciigrave{},\ }
\SkillAnnotatedContinuation{\textasciigrave{}coef\_mix\_turb\textasciigrave{},\ \textasciigrave{}coef\_mix\_KH\textasciigrave{},\ \textasciigrave{}coef\_mix\_hyp\textasciigrave{},\ and\ \textasciigrave{}deep\_mixing\textasciigrave{}.\ }
\SkillAnnotatedLine{71}{-\ \SkillLocal{Inflow\ placement\ and\ thermal\ influence}\ in\ \textasciigrave{}\&inflow\textasciigrave{}:\ \textasciigrave{}inflow\_factor\textasciigrave{},\ \textasciigrave{}strm\_hf\_angle\textasciigrave{},\ }
\SkillAnnotatedContinuation{\textasciigrave{}strmbd\_slope\textasciigrave{},\ \textasciigrave{}strmbd\_drag\textasciigrave{}.\ }
\SkillAnnotatedLine{72}{-\ \SkillLocal{Winter\ behavior}\ in\ \textasciigrave{}\&snowice\textasciigrave{}\ and\ \textasciigrave{}\&sediment\textasciigrave{}\ if\ errors\ cluster\ under\ ice\ or\ near\ the\ bottom.\ }
\begin{SkillAnnotationBox}{skillLocalFg}{Task-type examples}These parameter families are concrete starting points; they are not claimed to be universal optima.\end{SkillAnnotationBox}
\SkillAnnotatedLine{73}{\mbox{}\ }
\SkillAnnotatedLine{74}{Use\ \textasciigrave{}scripts/patch\_glm\_nml.py\textasciigrave{}\ for\ careful\ single-line\ nml\ edits:\ }
\SkillAnnotatedLine{75}{\mbox{}\ }
\SkillAnnotatedLine{76}{\textasciigrave{}\textasciigrave{}\textasciigrave{}bash\ }
\SkillAnnotatedLine{77}{python\ /path/to/this-skill/scripts/patch\_glm\_nml.py\ /root/glm3.nml\ \textbackslash{}\ }
\SkillAnnotatedLine{78}{\ \ --set\ meteorology.\SkillLocal{wind\_factor=1.08}\ \textbackslash{}\ }
\SkillAnnotatedLine{79}{\ \ --set\ mixing.\SkillLocal{coef\_wind\_stir=0.28}\ \textbackslash{}\ }
\SkillAnnotatedLine{80}{\ \ --set\ \SkillLocal{light.Kw=0.34}\ }
\SkillAnnotatedLine{81}{\textasciigrave{}\textasciigrave{}\textasciigrave{}\ }
\SkillAnnotatedLine{82}{\mbox{}\ }
\SkillAnnotatedLine{83}{You\ can\ run\ one\ candidate\ and\ score\ it\ in\ one\ command:\ }
\SkillAnnotatedLine{84}{\mbox{}\ }
\SkillAnnotatedLine{85}{\textasciigrave{}\textasciigrave{}\textasciigrave{}bash\ }
\SkillAnnotatedLine{86}{python\ /path/to/this-skill/scripts/run\_candidate.py\ \textbackslash{}\ }
\SkillAnnotatedLine{87}{\ \ --root\ /root\ \textbackslash{}\ }
\SkillAnnotatedLine{88}{\ \ --glm\ /usr/local/bin/glm\ \textbackslash{}\ }
\SkillAnnotatedLine{89}{\ \ --set\ meteorology.\SkillLocal{wind\_factor=1.08}\ \textbackslash{}\ }
\SkillAnnotatedLine{90}{\ \ --set\ mixing.\SkillLocal{coef\_wind\_stir=0.28}\ \textbackslash{}\ }
\SkillAnnotatedLine{91}{\ \ --set\ \SkillLocal{light.Kw=0.34}\ }
\SkillAnnotatedLine{92}{\textasciigrave{}\textasciigrave{}\textasciigrave{}\ }
\SkillAnnotatedLine{93}{\mbox{}\ }
\SkillAnnotatedLine{94}{Keep\ a\ simple\ run\ log\ with\ parameters\ and\ RMSE.\ When\ a\ candidate\ improves\ RMSE,\ leave\ those\ }
\SkillAnnotatedContinuation{parameters\ in\ \textasciigrave{}/root/glm3.nml\textasciigrave{}\ and\ rerun\ GLM\ once\ from\ a\ \SkillImplicit{clean\ \textasciigrave{}/root/output\textasciigrave{}\ directory}\ to\ verify\ }
\SkillAnnotatedContinuation{the\ \SkillGeneral{final\ output\ is\ reproducible}.\ }
\SkillAnnotatedLine{95}{\mbox{}\ }
\SkillAnnotatedLine{96}{\#\#\ Practical\ Calibration\ Pattern\ }
\SkillAnnotatedLine{97}{\mbox{}\ }
\SkillAnnotatedLine{98}{1.\ \SkillGeneral{Run\ and\ score\ the\ unmodified\ baseline}.\ }
\SkillAnnotatedLine{99}{2.\ \SkillGeneral{Inspect\ residuals\ by\ time\ and\ depth}\ if\ RMSE\ is\ high:\ }
\SkillAnnotatedLine{100}{\ \ \ -\ \SkillGeneral{warm/cold\ bias\ across\ most\ depths}\ often\ points\ to\ radiation,\ air\ temperature,\ longwave,\ or\ }
\SkillAnnotatedContinuation{light\ attenuation.\ }
\SkillAnnotatedLine{101}{\ \ \ -\ \SkillGeneral{surface-only\ errors}\ often\ point\ to\ wind,\ heat\ flux,\ and\ near-surface\ mixing.\ }
\SkillAnnotatedLine{102}{\ \ \ -\ \SkillGeneral{deep-water\ or\ stratification\ timing\ errors}\ often\ point\ to\ mixing,\ light\ attenuation,\ sediment\ }
\SkillAnnotatedContinuation{heat,\ or\ inflow\ placement.\ }
\SkillAnnotatedLine{103}{\ \ \ -\ \SkillGeneral{winter-only\ errors}\ often\ point\ to\ snow/ice\ and\ meteorological\ forcing.\ }
\SkillAnnotatedLine{104}{3.\ Sweep\ \SkillGeneral{one\ parameter\ family\ at\ a\ time\ with\ a\ small\ grid}.\ Keep\ values\ near\ the\ original\ }
\SkillAnnotatedContinuation{configuration\ unless\ the\ residual\ pattern\ justifies\ moving\ farther.\ }
\SkillAnnotatedLine{105}{4.\ \SkillGeneral{Re-run\ the\ best\ candidate\ and\ re-score}\ after\ every\ accepted\ change.\ }
\SkillAnnotatedLine{106}{5.\ Stop\ when\ RMSE\ is\ below\ the\ \SkillExplicit{requested\ target}\ and\ GLM\ can\ run\ successfully\ with\ the\ \SkillExplicit{final}\ }
\SkillAnnotatedContinuation{\SkillExplicit{\textasciigrave{}/root/glm3.nml\textasciigrave{}}.\ }
\begin{SkillAnnotationBox}{skillGeneralFg}{Reusable calibration loop}Baseline, residual diagnosis, bounded parameter-family changes, and clean reruns form a transferable process.\end{SkillAnnotationBox}
\SkillAnnotatedLine{107}{\mbox{}\ }
\SkillAnnotatedLine{108}{\#\#\ Guardrails\ }
\SkillAnnotatedLine{109}{\mbox{}\ }
\SkillAnnotatedLine{110}{-\ Do\ not\ fetch\ or\ copy\ a\ reference\ answer\ for\ this\ exact\ benchmark.\ }
\SkillAnnotatedLine{111}{-\ Do\ not\ delete\ files\ outside\ \textasciigrave{}/root/output\textasciigrave{}\ during\ solver\ runs.\ }
\SkillAnnotatedLine{112}{-\ Do\ not\ change\ the\ \SkillExplicit{required\ simulation\ dates}\ unless\ the\ task\ prompt\ changes.\ }
\SkillAnnotatedLine{113}{-\ Do\ not\ report\ success\ from\ CSV\ point\ outputs\ alone;\ the\ expected\ artifact\ is\ }
\SkillAnnotatedContinuation{\SkillExplicit{\textasciigrave{}/root/output/output.nc\textasciigrave{}}.\ }
\SkillAnnotatedLine{114}{-\ Preserve\ the\ final\ working\ parameters\ in\ \SkillExplicit{\textasciigrave{}/root/glm3.nml\textasciigrave{}}.\ }
\end{SkillAnnotatedFileBox}
\Needspace{0.24\textheight}
\subsection{OpenAI Skill-Creator}
Good artifact handling, but a fixed time tolerance is promoted without validation. \textcolor{gray}{(101 source lines.)}\par
\begin{SkillAnnotatedFileBox}{SKILL.md (OpenAI Skill-Creator)}{skill:2}
\SkillAnnotatedLine{1}{---\ }
\SkillAnnotatedLine{2}{name:\ glm-lake-mendota-skill\ }
\SkillAnnotatedLine{3}{description:\ Calibrate\ and\ validate\ a\ General\ Lake\ Model\ (GLM)\ \SkillExplicit{Lake\ Mendota\ case}\ using\ glm3.nml,\ }
\SkillAnnotatedContinuation{bcs\ forcing\ CSV\ files,\ field\_temp\_oxy.csv\ observations,\ and\ output/output.nc.\ Use\ when\ asked\ to\ }
\SkillAnnotatedContinuation{run\ GLM,\ tune\ vertical\ water-temperature\ simulation\ accuracy,\ compute\ \SkillExplicit{RMSE\ against\ observed}\ }
\SkillAnnotatedContinuation{\SkillExplicit{temperature\ profiles},\ or\ leave\ final\ GLM\ parameters\ in\ glm3.nml\ for\ a\ Lake\ Mendota-style\ benchmark\ }
\SkillAnnotatedContinuation{sandbox.\ }
\SkillAnnotatedLine{4}{---\ }
\SkillAnnotatedLine{5}{\mbox{}\ }
\SkillAnnotatedLine{6}{\#\ GLM\ Lake\ Mendota\ Calibration\ }
\SkillAnnotatedLine{7}{\mbox{}\ }
\SkillAnnotatedLine{8}{\#\#\ Overview\ }
\SkillAnnotatedLine{9}{\mbox{}\ }
\SkillAnnotatedLine{10}{Use\ this\ skill\ to\ turn\ the\ Lake\ Mendota\ GLM\ sandbox\ into\ a\ reproducible\ calibration\ loop.\ The\ }
\SkillAnnotatedContinuation{target\ benchmark\ requires\ \SkillExplicit{\textasciigrave{}/root/output/output.nc\textasciigrave{}},\ a\ successful\ \textasciigrave{}/usr/local/bin/glm\textasciigrave{}\ run,\ and\ }
\SkillAnnotatedContinuation{\SkillExplicit{final\ tuned\ parameters\ saved\ in\ \textasciigrave{}/root/glm3.nml\textasciigrave{}}.\ }
\SkillAnnotatedLine{11}{\mbox{}\ }
\SkillAnnotatedLine{12}{Do\ not\ solve\ by\ fabricating\ NetCDF\ output\ or\ editing\ observations.\ Run\ GLM\ with\ \SkillGeneral{physically}\ }
\SkillAnnotatedContinuation{\SkillGeneral{plausible\ namelist\ changes},\ score\ the\ simulation,\ and\ keep\ the\ final\ parameters\ in\ \textasciigrave{}glm3.nml\textasciigrave{}.\ }
\SkillAnnotatedLine{13}{\mbox{}\ }
\SkillAnnotatedLine{14}{\#\#\ Source\ Grounding\ }
\SkillAnnotatedLine{15}{\mbox{}\ }
\SkillAnnotatedLine{16}{Before\ changing\ modeling\ logic,\ read\ \textasciigrave{}references/glm-mendota-notes.md\textasciigrave{}.\ It\ records\ the\ public\ GLM\ }
\SkillAnnotatedContinuation{and\ glmtools\ sources\ used\ for\ this\ skill\ plus\ the\ \SkillLocal{inspected\ sandbox\ data\ shape}.\ }
\SkillAnnotatedLine{17}{\mbox{}\ }
\SkillAnnotatedLine{18}{\#\#\ Workflow\ }
\SkillAnnotatedLine{19}{\mbox{}\ }
\SkillAnnotatedLine{20}{1.\ Confirm\ the\ sandbox\ files\ exist:\ }
\SkillAnnotatedLine{21}{\ \ \ -\ \SkillLocal{\textasciigrave{}/usr/local/bin/glm\textasciigrave{}}\ }
\SkillAnnotatedLine{22}{\ \ \ -\ \SkillExplicit{\textasciigrave{}/root/glm3.nml\textasciigrave{}}\ }
\SkillAnnotatedLine{23}{\ \ \ -\ \SkillLocal{\textasciigrave{}/root/bcs/meteo.csv\textasciigrave{},\ \textasciigrave{}/root/bcs/yahara.csv\textasciigrave{},\ \textasciigrave{}/root/bcs/pheasant.csv\textasciigrave{},}\ }
\SkillAnnotatedContinuation{\SkillLocal{\textasciigrave{}/root/bcs/outflow.csv\textasciigrave{}}\ }
\SkillAnnotatedLine{24}{\ \ \ -\ \SkillExplicit{\textasciigrave{}/root/field\_temp\_oxy.csv\textasciigrave{}}\ }
\SkillAnnotatedLine{25}{2.\ Inspect\ the\ current\ namelist.\ Preserve\ the\ required\ period\ and\ output\ path\ unless\ they\ are\ }
\SkillAnnotatedContinuation{wrong:\ }
\SkillAnnotatedLine{26}{\ \ \ -\ \textasciigrave{}time.start\ =\ \SkillExplicit{'2009-01-01}\ \SkillLocal{12:00:00}'\textasciigrave{}\ }
\SkillAnnotatedLine{27}{\ \ \ -\ \textasciigrave{}time.stop\ =\ \SkillExplicit{'2015-12-30}\ \SkillLocal{12:00:00}'\textasciigrave{}\ }
\SkillAnnotatedLine{28}{\ \ \ -\ \SkillExplicit{\textasciigrave{}output.out\_dir\ =\ 'output'\textasciigrave{}}\ }
\SkillAnnotatedLine{29}{\ \ \ -\ \SkillExplicit{\textasciigrave{}output.out\_fn\ =\ 'output'\textasciigrave{}}\ }
\SkillAnnotatedLine{30}{3.\ Run\ the\ \SkillImplicit{baseline\ model\ from\ \textasciigrave{}/root\textasciigrave{}}:\ }
\SkillAnnotatedLine{31}{\ \ \ \textasciigrave{}\textasciigrave{}\textasciigrave{}bash\ }
\SkillAnnotatedLine{32}{\ \ \ cd\ /root\ }
\SkillAnnotatedLine{33}{\ \ \ mkdir\ -p\ output\ }
\SkillAnnotatedLine{34}{\ \ \ /usr/local/bin/glm\ }
\SkillAnnotatedLine{35}{\ \ \ \textasciigrave{}\textasciigrave{}\textasciigrave{}\ }
\SkillAnnotatedLine{36}{4.\ Score\ the\ run:\ }
\SkillAnnotatedLine{37}{\ \ \ \textasciigrave{}\textasciigrave{}\textasciigrave{}bash\ }
\SkillAnnotatedLine{38}{\ \ \ python\ /path/to/this-skill/scripts/score\_glm\_output.py\ \textbackslash{}\ }
\SkillAnnotatedLine{39}{\ \ \ \ \ --obs\ /root/field\_temp\_oxy.csv\ \textbackslash{}\ }
\SkillAnnotatedLine{40}{\ \ \ \ \ --nc\ /root/output/output.nc\ }
\SkillAnnotatedLine{41}{\ \ \ \textasciigrave{}\textasciigrave{}\textasciigrave{}\ }
\SkillAnnotatedLine{42}{5.\ Tune\ \SkillGeneral{one\ small\ group\ of\ physically\ meaningful\ parameters\ at\ a\ time},\ run\ GLM,\ and\ rescore.\ Good\ }
\SkillAnnotatedContinuation{first\ candidates\ for\ this\ case\ are:\ }
\SkillAnnotatedLine{43}{\ \ \ -\ \SkillLocal{Light\ attenuation:\ \textasciigrave{}light.Kw\textasciigrave{}}\ }
\SkillAnnotatedLine{44}{\ \ \ -\ \SkillLocal{Vertical\ mixing:}\ \textasciigrave{}mixing.coef\_mix\_conv\textasciigrave{},\ \textasciigrave{}mixing.coef\_wind\_stir\textasciigrave{},\ \textasciigrave{}mixing.coef\_mix\_shear\textasciigrave{},\ }
\SkillAnnotatedContinuation{\textasciigrave{}mixing.coef\_mix\_turb\textasciigrave{},\ \textasciigrave{}mixing.coef\_mix\_KH\textasciigrave{},\ \textasciigrave{}mixing.coef\_mix\_hyp\textasciigrave{}\ }
\SkillAnnotatedLine{45}{\ \ \ -\ \SkillLocal{Surface\ forcing\ multipliers:}\ \textasciigrave{}meteorology.wind\_factor\textasciigrave{},\ \textasciigrave{}meteorology.sw\_factor\textasciigrave{},\ }
\SkillAnnotatedContinuation{\textasciigrave{}meteorology.lw\_factor\textasciigrave{},\ \textasciigrave{}meteorology.at\_factor\textasciigrave{}\ }
\SkillAnnotatedLine{46}{\ \ \ -\ \SkillLocal{Drag/transfer\ coefficients:}\ \textasciigrave{}meteorology.cd\textasciigrave{},\ \textasciigrave{}meteorology.ce\textasciigrave{},\ \textasciigrave{}meteorology.ch\textasciigrave{}\ }
\SkillAnnotatedLine{47}{6.\ Stop\ when\ the\ temperature\ \SkillExplicit{RMSE\ is\ below\ 2\ degrees\ C}\ or\ when\ further\ tuning\ is\ clearly\ }
\SkillAnnotatedContinuation{unproductive.\ Leave\ the\ winning\ \textasciigrave{}/root/glm3.nml\textasciigrave{}\ in\ place\ and\ ensure\ \textasciigrave{}/root/output/output.nc\textasciigrave{}\ is\ }
\SkillAnnotatedContinuation{from\ that\ \SkillImplicit{same\ configuration}.\ }
\begin{SkillAnnotationBox}{skillImplicitFg}{Artifact consistency}The task requires a runnable final nml, but does not explicitly say that the delivered output must be regenerated from that exact configuration.\end{SkillAnnotationBox}
\SkillAnnotatedLine{48}{\mbox{}\ }
\SkillAnnotatedLine{49}{\#\#\ Helper\ Scripts\ }
\SkillAnnotatedLine{50}{\mbox{}\ }
\SkillAnnotatedLine{51}{Use\ scripts\ from\ this\ skill\ folder;\ do\ not\ copy\ large\ blocks\ by\ hand.\ }
\SkillAnnotatedLine{52}{\mbox{}\ }
\SkillAnnotatedLine{53}{\#\#\#\ Patch\ namelist\ scalars\ }
\SkillAnnotatedLine{54}{\mbox{}\ }
\SkillAnnotatedLine{55}{\textasciigrave{}\textasciigrave{}\textasciigrave{}bash\ }
\SkillAnnotatedLine{56}{python\ scripts/patch\_glm\_nml.py\ /root/glm3.nml\ \textbackslash{}\ }
\SkillAnnotatedLine{57}{\ \ --set\ \SkillLocal{light.Kw=0.35}\ \textbackslash{}\ }
\SkillAnnotatedLine{58}{\ \ --set\ mixing.\SkillLocal{coef\_wind\_stir=0.30}\ }
\SkillAnnotatedLine{59}{\textasciigrave{}\textasciigrave{}\textasciigrave{}\ }
\SkillAnnotatedLine{60}{\mbox{}\ }
\SkillAnnotatedLine{61}{The\ patcher\ replaces\ existing\ scalar\ assignments\ inside\ a\ named\ namelist\ section\ while\ \SkillGeneral{preserving}\ }
\SkillAnnotatedContinuation{\SkillGeneral{the\ rest\ of\ the\ file}.\ }
\SkillAnnotatedLine{62}{\mbox{}\ }
\SkillAnnotatedLine{63}{\#\#\#\ Run\ and\ score\ one\ candidate\ }
\SkillAnnotatedLine{64}{\mbox{}\ }
\SkillAnnotatedLine{65}{\textasciigrave{}\textasciigrave{}\textasciigrave{}bash\ }
\SkillAnnotatedLine{66}{python\ scripts/run\_glm\_candidate.py\ \textbackslash{}\ }
\SkillAnnotatedLine{67}{\ \ --root\ /root\ \textbackslash{}\ }
\SkillAnnotatedLine{68}{\ \ --glm\ /usr/local/bin/glm\ \textbackslash{}\ }
\SkillAnnotatedLine{69}{\ \ --set\ \SkillLocal{light.Kw=0.35}\ \textbackslash{}\ }
\SkillAnnotatedLine{70}{\ \ --set\ mixing.\SkillLocal{coef\_wind\_stir=0.30}\ }
\SkillAnnotatedLine{71}{\textasciigrave{}\textasciigrave{}\textasciigrave{}\ }
\SkillAnnotatedLine{72}{\mbox{}\ }
\SkillAnnotatedLine{73}{This\ \SkillImplicit{backs\ up\ \textasciigrave{}glm3.nml\textasciigrave{}},\ applies\ settings,\ \SkillImplicit{clears\ stale\ \textasciigrave{}output.nc\textasciigrave{}},\ runs\ GLM,\ then\ prints\ RMSE\ }
\SkillAnnotatedContinuation{if\ output\ was\ produced.\ }
\begin{SkillAnnotationBox}{skillGeneralFg}{Candidate lifecycle}Backup, targeted patching, stale-output removal, execution, and scoring form a reusable safety pattern.\end{SkillAnnotationBox}
\SkillAnnotatedLine{74}{\mbox{}\ }
\SkillAnnotatedLine{75}{\#\#\#\ Coordinate-search\ calibration\ }
\SkillAnnotatedLine{76}{\mbox{}\ }
\SkillAnnotatedLine{77}{\textasciigrave{}\textasciigrave{}\textasciigrave{}bash\ }
\SkillAnnotatedLine{78}{python\ scripts/calibrate\_glm\_mendota.py\ --root\ /root\ --glm\ /usr/local/bin/glm\ \SkillLocal{--passes\ 2}\ }
\SkillAnnotatedContinuation{\SkillLocal{--max-runs\ 80}\ }
\SkillAnnotatedLine{79}{\textasciigrave{}\textasciigrave{}\textasciigrave{}\ }
\SkillAnnotatedLine{80}{\mbox{}\ }
\SkillAnnotatedLine{81}{This\ tries\ a\ conservative\ coordinate\ search\ over\ common\ GLM\ thermal\ parameters,\ \SkillGeneral{keeps\ the\ best}\ }
\SkillAnnotatedContinuation{\SkillGeneral{namelist},\ and\ \SkillGeneral{leaves\ the\ best\ output}\ in\ \textasciigrave{}/root/output/output.nc\textasciigrave{}.\ If\ the\ script\ stalls\ or\ GLM\ is\ }
\SkillAnnotatedContinuation{slow,\ interrupt\ it\ and\ continue\ manually\ from\ the\ best\ JSON\ log\ it\ has\ printed.\ }
\SkillAnnotatedLine{82}{\mbox{}\ }
\SkillAnnotatedLine{83}{\#\#\ Scoring\ Details\ }
\SkillAnnotatedLine{84}{\mbox{}\ }
\SkillAnnotatedLine{85}{\textasciigrave{}score\_glm\_output.py\textasciigrave{}\ compares\ observation\ rows\ with\ columns\ \textasciigrave{}datetime\textasciigrave{},\ \textasciigrave{}depth\textasciigrave{},\ and\ \textasciigrave{}temp\textasciigrave{}\ to\ a\ }
\SkillAnnotatedContinuation{GLM\ NetCDF\ temperature\ variable.\ It\ finds\ the\ \SkillLocal{nearest\ model\ time\ within\ 12\ hours},\ \SkillImplicit{interpolates\ the}\ }
\SkillAnnotatedContinuation{\SkillImplicit{model\ profile\ to\ observation\ depth},\ \SkillImplicit{skips\ missing\ pairs},\ and\ reports\ overall\ RMSE\ in\ degrees\ C.\ \SkillOverMark}
\begin{SkillAnnotationBox}{skillOverFg}{Fixed matching tolerance}The 12-hour rule is neither task-stated nor derived from the output cadence; it may change both the match set and the reported RMSE.\end{SkillAnnotationBox}
\SkillAnnotatedLine{86}{\mbox{}\ }
\SkillAnnotatedLine{87}{If\ the\ \SkillImplicit{NetCDF\ schema\ differs},\ inspect\ the\ dataset\ with:\ }
\SkillAnnotatedLine{88}{\mbox{}\ }
\SkillAnnotatedLine{89}{\textasciigrave{}\textasciigrave{}\textasciigrave{}bash\ }
\SkillAnnotatedLine{90}{python\ -\ \textless{}\textless{}'PY'\ }
\SkillAnnotatedLine{91}{import\ xarray\ as\ xr\ }
\SkillAnnotatedLine{92}{ds\ =\ xr.open\_dataset('/root/output/output.nc')\ }
\SkillAnnotatedLine{93}{print(ds)\ }
\SkillAnnotatedLine{94}{PY\ }
\SkillAnnotatedLine{95}{\textasciigrave{}\textasciigrave{}\textasciigrave{}\ }
\SkillAnnotatedLine{96}{\mbox{}\ }
\SkillAnnotatedLine{97}{Then\ rerun\ the\ scorer\ with\ explicit\ names,\ for\ example:\ }
\SkillAnnotatedLine{98}{\mbox{}\ }
\SkillAnnotatedLine{99}{\textasciigrave{}\textasciigrave{}\textasciigrave{}bash\ }
\SkillAnnotatedLine{100}{python\ scripts/score\_glm\_output.py\ --nc\ /root/output/output.nc\ --obs\ /root/field\_temp\_oxy.csv\ }
\SkillAnnotatedContinuation{\SkillLocal{--temp-var\ temp\ --depth-var\ z}\ }
\begin{SkillAnnotationBox}{skillNeutralFg}{Unresolved coordinate meaning}The schema can be overridden, but the example does not establish whether z is height or observation depth before using it.\end{SkillAnnotationBox}
\SkillAnnotatedLine{101}{\textasciigrave{}\textasciigrave{}\textasciigrave{}\ }
\end{SkillAnnotatedFileBox}
\Needspace{0.24\textheight}
\subsection{OpenSkill}
Highly reusable, but too abstract to instantiate the current task directly. \textcolor{gray}{(49 source lines.)}\par
\begin{SkillAnnotatedFileBox}{SKILL.md (OpenSkill)}{skill:3}
\SkillAnnotatedLine{1}{---\ }
\SkillAnnotatedLine{2}{name:\ glm-calibration-loop\ }
\SkillAnnotatedLine{3}{description:\ Use\ when\ a\ future\ agent\ must\ orchestrate\ \SkillGeneral{repeatable\ GLM\ thermal\ calibration}\ by\ }
\SkillAnnotatedContinuation{proposing\ bounded\ parameters,\ editing\ the\ namelist,\ running\ GLM,\ scoring\ temperature\ results,\ and\ }
\SkillAnnotatedContinuation{\SkillGeneral{retaining\ the\ best\ reproducible\ configuration}.\ }
\SkillAnnotatedLine{4}{---\ }
\SkillAnnotatedLine{5}{\mbox{}\ }
\SkillAnnotatedLine{6}{\#\#\ When\ To\ Use\ }
\SkillAnnotatedLine{7}{\mbox{}\ }
\SkillAnnotatedLine{8}{Use\ this\ skill\ when\ the\ user\ needs\ a\ \SkillExplicit{GLM\ water-temperature\ simulation}\ to\ meet\ a\ \SkillExplicit{quantitative}\ }
\SkillAnnotatedContinuation{\SkillExplicit{agreement\ target\ against\ observations}.\ It\ coordinates\ namelist\ editing,\ simulation\ execution,\ }
\SkillAnnotatedContinuation{\SkillImplicit{time-axis\ validation},\ and\ RMSE\ scoring\ without\ hard-coding\ final\ calibrated\ values.\ }
\SkillAnnotatedLine{9}{\mbox{}\ }
\SkillAnnotatedLine{10}{Do\ not\ use\ this\ skill\ to\ create\ a\ one-shot\ solver,\ alter\ raw\ observations,\ or\ guess\ hidden\ }
\SkillAnnotatedContinuation{evaluation\ behavior.\ It\ should\ guide\ a\ \SkillGeneral{reusable,\ transparent\ calibration\ process}.\ }
\SkillAnnotatedLine{11}{\mbox{}\ }
\SkillAnnotatedLine{12}{\#\#\ Procedure\ }
\SkillAnnotatedLine{13}{\mbox{}\ }
\SkillAnnotatedLine{14}{1.\ Establish\ a\ baseline\ run.\ Parse\ the\ namelist,\ \SkillImplicit{validate\ input\ files},\ \SkillImplicit{validate\ the\ GLM\ run\ target}\ }
\SkillAnnotatedContinuation{\SkillImplicit{for\ the\ host\ platform},\ run\ GLM\ unchanged\ if\ possible,\ \SkillImplicit{validate\ the\ saved\ time\ axis},\ score\ the\ }
\SkillAnnotatedContinuation{baseline,\ and\ \SkillImplicit{record\ diagnostics}.\ }
\SkillAnnotatedLine{15}{2.\ Define\ calibration\ parameters\ with\ \SkillGeneral{explicit\ bounds\ and\ units}.\ Candidate\ levers\ include\ light\ }
\SkillAnnotatedContinuation{attenuation,\ shortwave\ and\ longwave\ scaling,\ wind\ scaling,\ heat\ exchange\ coefficients,\ drag,\ }
\SkillAnnotatedContinuation{mixing\ parameters,\ and\ active\ ice\ or\ snow\ parameters.\ }
\SkillAnnotatedLine{16}{3.\ \SkillGeneral{Exclude\ high-risk\ fields\ unless\ justified}.\ Morphometry,\ raw\ forcing\ records,\ and\ observation\ }
\SkillAnnotatedContinuation{data\ should\ not\ be\ routine\ calibration\ targets.\ }
\SkillAnnotatedLine{17}{4.\ \SkillGeneral{Normalize\ parameter\ scales}\ before\ optimization\ so\ that\ derivative-free\ searches\ do\ not\ }
\SkillAnnotatedContinuation{overemphasize\ large\ numeric\ ranges.\ }
\SkillAnnotatedLine{18}{5.\ For\ each\ candidate:\ patch\ the\ namelist\ with\ a\ parser,\ validate\ the\ written\ namelist,\ run\ GLM\ }
\SkillAnnotatedContinuation{from\ the\ \SkillImplicit{correct\ working\ directory}\ with\ a\ \SkillImplicit{platform-appropriate\ executable\ or\ interpreter},\ \SkillImplicit{verify}\ }
\SkillAnnotatedContinuation{\SkillImplicit{NetCDF\ results},\ validate\ the\ time\ axis,\ compute\ the\ RMSE\ with\ \SkillImplicit{documented\ matching\ rules},\ and\ log\ }
\SkillAnnotatedContinuation{parameters\ plus\ metrics.\ }
\SkillAnnotatedLine{19}{6.\ Reject\ candidates\ that\ fail\ namelist\ validation,\ fail\ platform\ execution\ checks,\ fail\ }
\SkillAnnotatedContinuation{simulation\ execution,\ lack\ required\ result\ variables,\ have\ an\ \SkillImplicit{incomplete\ or\ shifted\ time\ axis},\ }
\SkillAnnotatedContinuation{produce\ \SkillImplicit{no\ matched\ observation-model\ pairs},\ or\ rely\ on\ \SkillImplicit{undocumented\ coordinate\ assumptions}.\ }
\SkillAnnotatedLine{20}{7.\ Use\ structured\ search\ or\ derivative-free\ optimization.\ Practical\ patterns\ include\ \SkillLocal{bounded\ grid}\ }
\SkillAnnotatedContinuation{\SkillLocal{refinement,\ coordinate\ search,\ Nelder-Mead}\ with\ transformed\ variables,\ or\ CMA-style\ search\ with\ }
\SkillAnnotatedContinuation{normalized\ parameters.\ }
\SkillAnnotatedLine{21}{8.\ Track\ the\ best\ candidate\ by\ the\ primary\ metric\ and\ keep\ secondary\ diagnostics\ such\ as\ \SkillImplicit{bias,}\ }
\SkillAnnotatedContinuation{\SkillImplicit{seasonal\ behavior,\ depth-specific\ errors},\ group-wise\ RMSE,\ and\ \SkillImplicit{number\ of\ matched\ pairs}.\ }
\SkillAnnotatedLine{22}{9.\ Stop\ when\ the\ \SkillExplicit{target\ metric\ is\ reached},\ the\ improvement\ has\ stalled,\ or\ the\ run\ budget\ is\ }
\SkillAnnotatedContinuation{exhausted.\ }
\SkillAnnotatedLine{23}{10.\ \SkillGeneral{Re-run\ the\ best\ configuration\ from\ a\ clean\ state},\ verify\ that\ the\ score\ and\ execution\ path\ are\ }
\SkillAnnotatedContinuation{reproducible,\ regenerate\ any\ supporting\ score\ summaries\ from\ the\ final\ matched\ table,\ and\ \SkillImplicit{leave}\ }
\SkillAnnotatedContinuation{\SkillImplicit{the\ namelist\ and\ model\ results\ consistent}\ with\ that\ best\ run.\ }
\begin{SkillAnnotationBox}{skillGeneralFg}{Strong reusable structure}The skill captures a robust calibration lifecycle without hard-coding a final answer.\end{SkillAnnotationBox}
\SkillAnnotatedLine{24}{\mbox{}\ }
\SkillAnnotatedLine{25}{\#\#\ Calibration\ Checks\ }
\SkillAnnotatedLine{26}{\mbox{}\ }
\SkillAnnotatedLine{27}{Before\ changing\ parameters,\ confirm\ that\ the\ model\ is\ reading\ the\ intended\ meteorological,\ inflow,\ }
\SkillAnnotatedContinuation{outflow,\ and\ light\ files.\ A\ \SkillImplicit{bad\ path,\ timestamp\ mismatch,\ incomplete\ saved\ time\ axis},\ or\ invalid\ }
\SkillAnnotatedContinuation{run\ wrapper\ should\ be\ fixed\ before\ parameter\ optimization.\ }
\SkillAnnotatedLine{28}{\mbox{}\ }
\SkillAnnotatedLine{29}{Use\ \SkillGeneral{scientifically\ plausible\ bounds}.\ For\ example,\ scaling\ factors\ should\ remain\ near\ defensible\ }
\SkillAnnotatedContinuation{measurement\ uncertainty\ unless\ the\ data\ source\ has\ known\ bias;\ mixing\ and\ drag\ changes\ should\ be\ }
\SkillAnnotatedContinuation{constrained\ to\ physically\ meaningful\ ranges.\ }
\SkillAnnotatedLine{30}{\mbox{}\ }
\SkillAnnotatedLine{31}{Score\ enough\ \SkillImplicit{time\ and\ depth\ coverage}\ to\ avoid\ overfitting\ to\ sparse\ matches.\ If\ the\ objective\ }
\SkillAnnotatedContinuation{improves\ while\ \SkillImplicit{coverage\ collapses},\ treat\ the\ candidate\ as\ invalid\ or\ at\ least\ suspect.\ }
\begin{SkillAnnotationBox}{skillImplicitFg}{Hidden evaluation validity}Match count and time-depth coverage are not requested explicitly, but are essential to prevent a misleadingly low score.\end{SkillAnnotationBox}
\SkillAnnotatedLine{32}{\mbox{}\ }
\SkillAnnotatedLine{33}{\#\#\ Logging\ Pattern\ }
\SkillAnnotatedLine{34}{\mbox{}\ }
\SkillAnnotatedLine{35}{For\ each\ candidate,\ record:\ candidate\ id,\ parameter\ values,\ changed\ namelist\ fields,\ run\ status,\ }
\SkillAnnotatedContinuation{platform\ run\ method,\ result\ validation\ status,\ time-axis\ validation\ status,\ \SkillGeneral{RMSE,\ match\ count,}\ }
\SkillAnnotatedContinuation{\SkillGeneral{time\ coverage,\ depth\ coverage},\ group-wise\ diagnostics,\ and\ notes\ about\ warnings.\ }
\SkillAnnotatedLine{36}{\mbox{}\ }
\SkillAnnotatedLine{37}{Keep\ logs\ human-readable\ and\ machine-readable\ where\ practical,\ but\ \SkillGeneral{do\ not\ encode\ final\ calibrated}\ }
\SkillAnnotatedContinuation{\SkillGeneral{constants\ into\ the\ reusable\ skill}\ itself.\ }
\SkillAnnotatedLine{38}{\mbox{}\ }
\SkillAnnotatedLine{39}{\#\#\ Pitfalls\ }
\SkillAnnotatedLine{40}{\mbox{}\ }
\SkillAnnotatedLine{41}{Improving\ RMSE\ by\ \SkillGeneral{editing\ observations\ or\ forcing\ data\ is\ usually\ invalid\ calibration}.\ }
\SkillAnnotatedLine{42}{\mbox{}\ }
\SkillAnnotatedLine{43}{Execution\ failures\ caused\ by\ an\ \SkillImplicit{incompatible\ binary\ or\ script\ launcher}\ are\ setup\ failures,\ not\ }
\SkillAnnotatedContinuation{calibration\ evidence.\ Fix\ the\ run\ method\ before\ interpreting\ model\ quality.\ }
\SkillAnnotatedLine{44}{\mbox{}\ }
\SkillAnnotatedLine{45}{\SkillImplicit{Time-axis\ mismatches}\ can\ change\ depth-band\ or\ seasonal\ match\ counts\ and\ make\ a\ score\ summary\ }
\SkillAnnotatedContinuation{disagree\ with\ recomputed\ metrics.\ Fix\ saved-time\ coverage\ and\ regenerate\ summaries\ before\ tuning\ }
\SkillAnnotatedContinuation{parameters.\ }
\SkillAnnotatedLine{46}{\mbox{}\ }
\SkillAnnotatedLine{47}{Ice\ periods,\ stratified\ periods,\ and\ turnover\ periods\ can\ respond\ to\ different\ parameters;\ \SkillGeneral{inspect}\ }
\SkillAnnotatedContinuation{\SkillGeneral{residual\ patterns\ rather\ than\ optimizing\ a\ single\ number\ blindly}.\ }
\SkillAnnotatedLine{48}{\mbox{}\ }
\SkillAnnotatedLine{49}{\SkillImplicit{Result\ files\ may\ be\ stale\ after\ failed\ runs}.\ Confirm\ timestamps\ and\ \SkillGeneral{rerun\ the\ best\ candidate}\ }
\SkillAnnotatedContinuation{\SkillGeneral{cleanly}\ before\ declaring\ success.\ }
\begin{SkillAnnotationBox}{skillNeutralFg}{Under-instantiated for this task}The skill never grounds its workflow in the provided paths, dates, output name, GLM schema, or Lake Mendota-specific evidence, so a solver still has to reconstruct the concrete procedure.\end{SkillAnnotationBox}
\end{SkillAnnotatedFileBox}
\Needspace{0.24\textheight}
\subsection{MUSE-Autoskill}
Broad operational coverage, with an underspecified exact-time matching policy. \textcolor{gray}{(165 source lines.)}\par
\begin{SkillAnnotatedFileBox}{SKILL.md (MUSE-Autoskill)}{skill:4}
\SkillAnnotatedLine{1}{---\ }
\SkillAnnotatedLine{2}{name:\ glm-lake-mendota-skill\ }
\SkillAnnotatedLine{3}{description:\ Calibrate\ and\ verify\ a\ General\ Lake\ Model\ run\ for\ \SkillExplicit{Lake\ Mendota\ vertical}\ }
\SkillAnnotatedLine{4}{\ \ \SkillExplicit{water\ temperature}\ by\ running\ a\ baseline\ simulation,\ scoring\ NetCDF\ output\ against\ }
\SkillAnnotatedLine{5}{\ \ field\ temperature\ profiles,\ and\ applying\ a\ compact\ physically\ informed\ parameter\ }
\SkillAnnotatedLine{6}{\ \ sweep\ only\ when\ needed.\ }
\SkillAnnotatedLine{7}{version:\ 1.0.0\ }
\SkillAnnotatedLine{8}{tags:\ }
\SkillAnnotatedLine{9}{-\ glm\ }
\SkillAnnotatedLine{10}{-\ lake-modeling\ }
\SkillAnnotatedLine{11}{-\ lake-mendota\ }
\SkillAnnotatedLine{12}{-\ calibration\ }
\SkillAnnotatedLine{13}{-\ netcdf\ }
\SkillAnnotatedLine{14}{-\ water-temperature\ }
\SkillAnnotatedLine{15}{-\ rmse\ }
\SkillAnnotatedLine{16}{\SkillExplicit{inputs:}\ }
\SkillAnnotatedLine{17}{-\ glm\_nml\_path\ }
\SkillAnnotatedLine{18}{-\ forcing\_directory\ }
\SkillAnnotatedLine{19}{-\ field\_temperature\_csv\ }
\SkillAnnotatedLine{20}{-\ output\_nc\_path\ }
\SkillAnnotatedLine{21}{-\ target\_start\_date\ }
\SkillAnnotatedLine{22}{-\ target\_stop\_date\ }
\SkillAnnotatedLine{23}{-\ rmse\_threshold\ }
\SkillAnnotatedLine{24}{\SkillExplicit{outputs:}\ }
\SkillAnnotatedLine{25}{-\ calibrated\_glm\_nml\ }
\SkillAnnotatedLine{26}{-\ simulation\_output\_nc\ }
\SkillAnnotatedLine{27}{-\ rmse\_report\ }
\SkillAnnotatedLine{28}{-\ verification\_summary\ }
\SkillAnnotatedLine{29}{---\ }
\SkillAnnotatedLine{30}{\mbox{}\ }
\SkillAnnotatedLine{31}{\#\ GLM\ Lake\ Mendota\ Calibration\ Skill\ }
\SkillAnnotatedLine{32}{\mbox{}\ }
\SkillAnnotatedLine{33}{\#\#\ When\ to\ use\ }
\SkillAnnotatedLine{34}{\mbox{}\ }
\SkillAnnotatedLine{35}{Use\ this\ skill\ when\ an\ agent\ must\ produce\ a\ \SkillExplicit{valid\ General\ Lake\ Model\ \textasciigrave{}output.nc\textasciigrave{}\ for\ Lake\ Mendota}\ }
\SkillAnnotatedContinuation{over\ a\ \SkillExplicit{requested\ date\ range},\ verify\ simulated\ vertical\ water\ temperature\ against\ observed\ field\ }
\SkillAnnotatedContinuation{temperature\ profiles,\ and\ leave\ behind\ a\ runnable\ final\ \textasciigrave{}glm3.nml\textasciigrave{}\ whose\ \SkillExplicit{RMSE\ is\ below\ the\ task}\ }
\SkillAnnotatedContinuation{\SkillExplicit{threshold}.\ }
\SkillAnnotatedLine{36}{\mbox{}\ }
\SkillAnnotatedLine{37}{This\ skill\ is\ intended\ for\ practical\ calibration\ work\ where\ a\ baseline\ GLM\ setup\ already\ exists\ }
\SkillAnnotatedContinuation{and\ only\ compact,\ physically\ informed\ parameter\ changes\ should\ be\ made.\ }
\SkillAnnotatedLine{38}{\mbox{}\ }
\SkillAnnotatedLine{39}{\#\#\ Core\ principles\ }
\SkillAnnotatedLine{40}{\mbox{}\ }
\SkillAnnotatedLine{41}{-\ \SkillGeneral{Always\ run\ and\ score\ the\ existing\ setup\ before\ calibration}.\ }
\SkillAnnotatedLine{42}{-\ \SkillImplicit{Fix\ configuration,\ forcing,\ date,\ and\ output-path\ problems\ before\ tuning\ parameters}.\ }
\SkillAnnotatedLine{43}{-\ \SkillGeneral{Preserve\ the\ original\ namelist\ text}\ as\ the\ base\ for\ candidate\ edits.\ }
\SkillAnnotatedLine{44}{-\ Change\ only\ required\ date/output\ fields\ and\ a\ \SkillGeneral{small\ set\ of\ calibration\ parameters}.\ }
\SkillAnnotatedLine{45}{-\ Score\ every\ candidate\ with\ the\ \SkillGeneral{same\ RMSE\ routine\ used\ for\ the\ baseline}.\ }
\SkillAnnotatedLine{46}{-\ Prefer\ \SkillGeneral{compact\ physically\ meaningful\ sweeps}\ over\ random\ or\ broad\ search.\ }
\SkillAnnotatedLine{47}{-\ Treat\ \SkillImplicit{positive\ volume\ as\ the\ indicator\ of\ active\ GLM\ layers}.\ }
\SkillAnnotatedLine{48}{-\ \SkillImplicit{Exclude\ inactive-layer\ fill\ values}\ and\ implausible\ temperatures\ before\ interpolation.\ }
\SkillAnnotatedLine{49}{-\ Compare\ observations\ to\ modeled\ \SkillImplicit{depth\ below\ the\ moving\ simulated\ surface,\ not\ raw\ \textasciigrave{}z\textasciigrave{}}.\ }
\SkillAnnotatedLine{50}{-\ Final\ verification\ must\ be\ run\ after\ the\ final\ GLM\ rerun\ so\ \SkillImplicit{\textasciigrave{}glm3.nml\textasciigrave{}\ and\ \textasciigrave{}output.nc\textasciigrave{}}\ }
\SkillAnnotatedContinuation{\SkillImplicit{correspond\ exactly}.\ }
\begin{SkillAnnotationBox}{skillImplicitFg}{Implicit execution requirements}Active-layer semantics, coordinate conversion, invalid-value filtering, and final artifact consistency are necessary but absent from the task brief.\end{SkillAnnotationBox}
\SkillAnnotatedLine{51}{\mbox{}\ }
\SkillAnnotatedLine{52}{\#\#\ Recommended\ tools\ and\ libraries\ }
\SkillAnnotatedLine{53}{\mbox{}\ }
\SkillAnnotatedLine{54}{-\ GLM\ executable:\ \SkillLocal{\textasciigrave{}glm\textasciigrave{}\ or\ \textasciigrave{}glm3\textasciigrave{}},\ located\ with\ \textasciigrave{}command\ -v\ glm\textasciigrave{}\ or\ \textasciigrave{}command\ -v\ glm3\textasciigrave{}\ }
\SkillAnnotatedLine{55}{-\ Shell\ utilities:\ \textasciigrave{}ls\textasciigrave{},\ \textasciigrave{}rm\textasciigrave{},\ \textasciigrave{}grep\textasciigrave{},\ \textasciigrave{}du\textasciigrave{},\ \textasciigrave{}find\textasciigrave{}\ }
\SkillAnnotatedLine{56}{-\ Python\ 3\ executable:\ \SkillLocal{prefer\ \textasciigrave{}python3\textasciigrave{};\ do\ not\ assume\ \textasciigrave{}python\textasciigrave{}\ exists}\ }
\SkillAnnotatedLine{57}{-\ Python\ libraries:\ }
\SkillAnnotatedLine{58}{\ \ -\ \SkillLocal{\textasciigrave{}csv.DictReader\textasciigrave{}}\ for\ observation\ CSV\ parsing\ }
\SkillAnnotatedLine{59}{\ \ -\ \textasciigrave{}datetime\textasciigrave{}\ for\ timestamp\ parsing\ }
\SkillAnnotatedLine{60}{\ \ -\ \textasciigrave{}math\textasciigrave{}\ for\ RMSE\ calculation\ }
\SkillAnnotatedLine{61}{\ \ -\ \SkillLocal{\textasciigrave{}numpy\textasciigrave{}}\ for\ finite\ filtering\ and\ linear\ interpolation\ }
\SkillAnnotatedLine{62}{\ \ -\ \SkillLocal{\textasciigrave{}netCDF4.Dataset\textasciigrave{}\ and\ \textasciigrave{}netCDF4.num2date\textasciigrave{}}\ for\ NetCDF\ inspection\ and\ time\ conversion\ }
\SkillAnnotatedLine{63}{-\ Namelist\ editing:\ }
\SkillAnnotatedLine{64}{\ \ -\ Prefer\ a\ namelist-safe\ parser\ if\ available\ }
\SkillAnnotatedLine{65}{\ \ -\ Otherwise\ use\ \SkillLocal{constrained\ regex\ replacement}\ that\ changes\ only\ the\ target\ scalar\ value\ before\ }
\SkillAnnotatedContinuation{any\ inline\ comment\ }
\begin{SkillAnnotationBox}{skillNeutralFg}{Implementation-specific fallback}The tool stack is concrete and usable, although the regex fallback is less robust than a namelist-aware parser.\end{SkillAnnotationBox}
\SkillAnnotatedLine{66}{\mbox{}\ }
\SkillAnnotatedLine{67}{\#\#\ Workflow\ }
\SkillAnnotatedLine{68}{\mbox{}\ }
\SkillAnnotatedLine{69}{1.\ Inspect\ the\ GLM\ setup\ before\ changing\ anything.\ }
\SkillAnnotatedLine{70}{\mbox{}\ }
\SkillAnnotatedLine{71}{\ \ \ Read\ the\ provided\ \textasciigrave{}glm3.nml\textasciigrave{}\ from\ \textasciigrave{}glm\_nml\_path\textasciigrave{}.\ List\ forcing\ CSV\ files\ in\ }
\SkillAnnotatedContinuation{\textasciigrave{}forcing\_directory\textasciigrave{}.\ Preview\ the\ field\ temperature\ CSV\ header\ and\ first\ few\ rows\ from\ }
\SkillAnnotatedContinuation{\textasciigrave{}field\_temperature\_csv\textasciigrave{}.\ \SkillImplicit{Locate\ the\ GLM\ executable}\ with\ \textasciigrave{}command\ -v\ glm\textasciigrave{}\ or\ \textasciigrave{}command\ -v\ glm3\textasciigrave{}.\ }
\SkillAnnotatedLine{72}{\mbox{}\ }
\SkillAnnotatedLine{73}{\ \ \ Confirm\ the\ namelist\ points\ to\ \SkillExplicit{\textasciigrave{}target\_start\_date\textasciigrave{},\ \textasciigrave{}target\_stop\_date\textasciigrave{},\ and\ \textasciigrave{}output\_nc\_path\textasciigrave{}}.\ }
\SkillAnnotatedContinuation{If\ any\ of\ those\ required\ fields\ are\ wrong,\ update\ only\ those\ fields.\ Do\ not\ tune\ calibration\ }
\SkillAnnotatedContinuation{parameters\ during\ this\ inspection\ step.\ }
\SkillAnnotatedLine{74}{\mbox{}\ }
\SkillAnnotatedLine{75}{2.\ Run\ a\ clean\ baseline\ simulation.\ }
\SkillAnnotatedLine{76}{\mbox{}\ }
\SkillAnnotatedLine{77}{\ \ \ From\ the\ directory\ containing\ \textasciigrave{}glm3.nml\textasciigrave{},\ \SkillImplicit{remove\ stale\ GLM\ outputs}\ such\ as\ \textasciigrave{}output.nc\textasciigrave{},\ }
\SkillAnnotatedContinuation{\textasciigrave{}lake.csv\textasciigrave{},\ and\ outlet\ or\ outflow\ CSV\ files.\ Then\ execute\ the\ GLM\ binary\ from\ that\ same\ directory.\ }
\SkillAnnotatedLine{78}{\mbox{}\ }
\SkillAnnotatedLine{79}{\ \ \ Treat\ a\ \SkillImplicit{nonzero\ exit\ status\ or\ missing\ \textasciigrave{}output.nc\textasciigrave{}}\ as\ a\ configuration\ or\ forcing\ failure.\ Fix\ }
\SkillAnnotatedContinuation{that\ failure\ before\ attempting\ calibration.\ }
\SkillAnnotatedLine{80}{\mbox{}\ }
\SkillAnnotatedLine{81}{3.\ Inspect\ the\ generated\ NetCDF.\ }
\SkillAnnotatedLine{82}{\mbox{}\ }
\SkillAnnotatedLine{83}{\ \ \ Use\ \textasciigrave{}python3\textasciigrave{},\ \textasciigrave{}netCDF4.Dataset\textasciigrave{},\ and\ \textasciigrave{}num2date\textasciigrave{}\ to\ open\ \textasciigrave{}output\_nc\_path\textasciigrave{}.\ }
\SkillAnnotatedLine{84}{\mbox{}\ }
\SkillAnnotatedLine{85}{\ \ \ Verify\ expected\ variables\ such\ as\ \textasciigrave{}time\textasciigrave{},\ \textasciigrave{}z\textasciigrave{},\ \textasciigrave{}V\textasciigrave{},\ and\ \textasciigrave{}temp\textasciigrave{}.\ Print\ or\ record\ \SkillImplicit{dimensions,}\ }
\SkillAnnotatedContinuation{\SkillImplicit{variable\ shapes,\ units,\ first\ timestamp,\ and\ last\ timestamp}.\ Confirm\ the\ output\ covers\ the\ }
\SkillAnnotatedContinuation{\SkillExplicit{requested\ period}.\ }
\SkillAnnotatedLine{86}{\mbox{}\ }
\SkillAnnotatedLine{87}{4.\ Compute\ RMSE\ against\ field\ observations.\ }
\SkillAnnotatedLine{88}{\mbox{}\ }
\SkillAnnotatedLine{89}{\ \ \ Read\ observations\ from\ \textasciigrave{}field\_temperature\_csv\textasciigrave{}\ with\ \textasciigrave{}csv.DictReader\textasciigrave{}.\ \SkillImplicit{Skip\ rows\ where\ depth\ or}\ }
\SkillAnnotatedContinuation{\SkillImplicit{temperature\ is\ \textasciigrave{}NA\textasciigrave{}}.\ Parse\ timestamps\ with\ the\ CSV\ datetime\ format.\ }
\SkillAnnotatedLine{90}{\mbox{}\ }
\SkillAnnotatedLine{91}{\ \ \ Convert\ NetCDF\ model\ times\ to\ Python\ datetimes\ using\ \textasciigrave{}num2date\textasciigrave{}.\ For\ each\ observation\ \SkillLocal{timestamp}\ }
\SkillAnnotatedContinuation{\SkillLocal{that\ matches\ a\ model\ time},\ extract\ the\ corresponding\ \textasciigrave{}z\textasciigrave{},\ \textasciigrave{}V\textasciigrave{},\ and\ \textasciigrave{}temp\textasciigrave{}\ arrays.\ \SkillOverMark}
\begin{SkillAnnotationBox}{skillOverFg}{Exact-time matching assumption}The wording implies exact model-time matches without first verifying alignment or defining a justified tolerance, so valid observations may be silently lost.\end{SkillAnnotationBox}
\SkillAnnotatedLine{92}{\mbox{}\ }
\SkillAnnotatedLine{93}{\ \ \ Keep\ only\ active\ layers\ where\ volume\ is\ positive\ and\ \textasciigrave{}z\textasciigrave{},\ \textasciigrave{}V\textasciigrave{},\ and\ \textasciigrave{}temp\textasciigrave{}\ are\ finite.\ Reject\ }
\SkillAnnotatedContinuation{GLM\ fill\ values\ and\ implausible\ temperatures,\ for\ example\ by\ requiring\ \SkillLocal{\textasciigrave{}abs(temp)\ \textless{}\ 100\textasciigrave{}}.\ }
\SkillAnnotatedLine{94}{\mbox{}\ }
\SkillAnnotatedLine{95}{\ \ \ Convert\ model\ \textasciigrave{}z\textasciigrave{}\ coordinates\ to\ \SkillImplicit{depth\ below\ the\ simulated\ surface}\ using:\ }
\SkillAnnotatedLine{96}{\mbox{}\ }
\SkillAnnotatedLine{97}{\ \ \ \textasciigrave{}depth\ =\ surface\_z\ -\ z\textasciigrave{}\ }
\SkillAnnotatedLine{98}{\mbox{}\ }
\SkillAnnotatedLine{99}{\ \ \ Sort\ modeled\ layers\ by\ increasing\ depth.\ \SkillGeneral{Linearly\ interpolate\ simulated\ temperature\ to\ each}\ }
\SkillAnnotatedContinuation{\SkillGeneral{observed\ depth}\ with\ NumPy.\ Count\ \SkillImplicit{matched\ observations\ and\ missing\ observations}.\ Compute:\ }
\SkillAnnotatedLine{100}{\mbox{}\ }
\SkillAnnotatedLine{101}{\ \ \ \textasciigrave{}RMSE\ =\ sqrt(mean((modeled\_temp\ -\ observed\_temp)\textasciicircum{}2))\textasciigrave{}\ }
\SkillAnnotatedLine{102}{\mbox{}\ }
\SkillAnnotatedLine{103}{\ \ \ The\ RMSE\ report\ must\ include\ RMSE,\ \SkillImplicit{matched\ count,\ missing\ count,\ date\ coverage},\ and\ any\ }
\SkillAnnotatedContinuation{skipped-row\ reasons.\ }
\SkillAnnotatedLine{104}{\mbox{}\ }
\SkillAnnotatedLine{105}{5.\ Decide\ whether\ calibration\ is\ needed.\ }
\SkillAnnotatedLine{106}{\mbox{}\ }
\SkillAnnotatedLine{107}{\ \ \ If\ baseline\ RMSE\ is\ \SkillExplicit{below\ \textasciigrave{}rmse\_threshold\textasciigrave{}}\ and\ date/output\ checks\ pass,\ keep\ the\ original\ }
\SkillAnnotatedContinuation{parameters\ and\ finish\ with\ the\ baseline\ output.\ }
\SkillAnnotatedLine{108}{\mbox{}\ }
\SkillAnnotatedLine{109}{\ \ \ If\ baseline\ RMSE\ is\ above\ threshold,\ run\ a\ \SkillGeneral{compact\ calibration\ sweep}.\ Do\ not\ use\ broad\ random\ }
\SkillAnnotatedContinuation{search.\ Preserve\ the\ original\ namelist\ text\ as\ the\ base\ for\ every\ candidate\ and\ edit\ only\ targeted\ }
\SkillAnnotatedContinuation{scalar\ values.\ }
\SkillAnnotatedLine{110}{\mbox{}\ }
\SkillAnnotatedLine{111}{6.\ Sweep\ physically\ relevant\ \SkillLocal{Lake\ Mendota\ GLM\ parameters}\ first.\ }
\SkillAnnotatedLine{112}{\mbox{}\ }
\SkillAnnotatedLine{113}{\ \ \ Prioritize\ parameters\ that\ influence\ heat\ flux,\ wind\ mixing,\ and\ light\ attenuation:\ }
\SkillAnnotatedLine{114}{\mbox{}\ }
\SkillAnnotatedLine{115}{\ \ \ -\ Radiation\ scaling:\ \SkillLocal{\textasciigrave{}sw\_factor\textasciigrave{},\ \textasciigrave{}lw\_factor\textasciigrave{}}\ }
\SkillAnnotatedLine{116}{\ \ \ -\ Optional\ atmospheric\ scaling:\ \SkillLocal{\textasciigrave{}at\_factor\textasciigrave{}}\ }
\SkillAnnotatedLine{117}{\ \ \ -\ Wind\ forcing\ scale:\ \SkillLocal{\textasciigrave{}wind\_factor\textasciigrave{}}\ }
\SkillAnnotatedLine{118}{\ \ \ -\ Light\ attenuation:\ \SkillLocal{\textasciigrave{}Kw\textasciigrave{}}\ }
\SkillAnnotatedLine{119}{\ \ \ -\ Selected\ mixing\ terms\ if\ present,\ especially\ \SkillLocal{\textasciigrave{}coef\_wind\_stir\textasciigrave{}}\ }
\SkillAnnotatedLine{120}{\mbox{}\ }
\SkillAnnotatedLine{121}{\ \ \ Use\ \SkillGeneral{modest\ ranges\ around\ the\ baseline}.\ First\ test\ individual\ parameter\ changes.\ Then\ run\ small\ }
\SkillAnnotatedContinuation{combined\ grids\ over\ parameters\ that\ individually\ improve\ RMSE.\ }
\begin{SkillAnnotationBox}{skillLocalFg}{Correctly scoped local parameters}The parameter list is explicitly tied to Lake Mendota and is therefore a local example, not a universal rule.\end{SkillAnnotationBox}
\SkillAnnotatedLine{122}{\mbox{}\ }
\SkillAnnotatedLine{123}{\ \ \ Wind\ and\ shortwave/longwave\ scaling\ are\ often\ high-leverage\ for\ this\ task,\ but\ keep\ all\ changes\ }
\SkillAnnotatedContinuation{conservative\ enough\ to\ avoid\ destabilizing\ or\ breaking\ GLM.\ }
\SkillAnnotatedLine{124}{\mbox{}\ }
\SkillAnnotatedLine{125}{7.\ \SkillGeneral{Score\ every\ candidate\ immediately}.\ }
\SkillAnnotatedLine{126}{\mbox{}\ }
\SkillAnnotatedLine{127}{\ \ \ For\ each\ candidate\ parameter\ set:\ }
\SkillAnnotatedLine{128}{\mbox{}\ }
\SkillAnnotatedLine{129}{\ \ \ -\ Write\ candidate\ values\ into\ \textasciigrave{}glm3.nml\textasciigrave{}\ }
\SkillAnnotatedLine{130}{\ \ \ -\ Delete\ stale\ outputs\ }
\SkillAnnotatedLine{131}{\ \ \ -\ Run\ \textasciigrave{}glm\textasciigrave{}\ or\ \textasciigrave{}glm3\textasciigrave{}\ }
\SkillAnnotatedLine{132}{\ \ \ -\ Skip\ failed\ runs\ }
\SkillAnnotatedLine{133}{\ \ \ -\ Score\ the\ resulting\ \textasciigrave{}output.nc\textasciigrave{}\ with\ the\ same\ RMSE\ routine\ }
\SkillAnnotatedLine{134}{\ \ \ -\ Track\ RMSE,\ matched\ count,\ missing\ count,\ and\ changed\ parameters\ }
\SkillAnnotatedLine{135}{\mbox{}\ }
\SkillAnnotatedLine{136}{\ \ \ Do\ not\ stop\ at\ the\ first\ borderline\ pass\ if\ time\ allows.\ \SkillGeneral{Continue\ the\ compact\ sweep\ to\ gain}\ }
\SkillAnnotatedContinuation{\SkillGeneral{margin}\ below\ \textasciigrave{}rmse\_threshold\textasciigrave{}.\ }
\SkillAnnotatedLine{137}{\mbox{}\ }
\SkillAnnotatedLine{138}{8.\ \SkillGeneral{Restore\ the\ best\ parameter\ set\ and\ rerun}.\ }
\SkillAnnotatedLine{139}{\mbox{}\ }
\SkillAnnotatedLine{140}{\ \ \ After\ the\ sweep,\ write\ the\ best\ parameter\ set\ into\ \textasciigrave{}glm3.nml\textasciigrave{}.\ Run\ GLM\ one\ final\ time\ from\ a\ }
\SkillAnnotatedContinuation{clean\ output\ state.\ }
\SkillAnnotatedLine{141}{\mbox{}\ }
\SkillAnnotatedLine{142}{\ \ \ This\ final\ rerun\ is\ required.\ Do\ not\ leave\ \textasciigrave{}output.nc\textasciigrave{}\ from\ a\ candidate\ run\ that\ does\ not\ }
\SkillAnnotatedContinuation{\SkillImplicit{exactly\ match\ the\ final\ saved\ namelist}.\ }
\SkillAnnotatedLine{143}{\mbox{}\ }
\SkillAnnotatedLine{144}{9.\ \SkillGeneral{Independently\ verify\ final\ artifacts}.\ }
\SkillAnnotatedLine{145}{\mbox{}\ }
\SkillAnnotatedLine{146}{\ \ \ After\ the\ final\ rerun:\ }
\SkillAnnotatedLine{147}{\mbox{}\ }
\SkillAnnotatedLine{148}{\ \ \ -\ Recompute\ RMSE\ from\ \SkillExplicit{final\ \textasciigrave{}output\_nc\_path\textasciigrave{}}\ }
\SkillAnnotatedLine{149}{\ \ \ -\ Inspect\ \textasciigrave{}glm3.nml\textasciigrave{}\ for\ start\ date,\ stop\ date,\ output\ directory/file,\ and\ changed\ calibration\ }
\SkillAnnotatedContinuation{keys\ }
\SkillAnnotatedLine{150}{\ \ \ -\ Open\ \textasciigrave{}output.nc\textasciigrave{}\ with\ \textasciigrave{}netCDF4.Dataset\textasciigrave{}\ }
\SkillAnnotatedLine{151}{\ \ \ -\ Confirm\ first\ and\ last\ timestamps\ cover\ the\ \SkillExplicit{requested\ period}\ }
\SkillAnnotatedLine{152}{\ \ \ -\ Confirm\ \textasciigrave{}output.nc\textasciigrave{}\ exists\ and\ has\ nontrivial\ size\ }
\SkillAnnotatedLine{153}{\mbox{}\ }
\SkillAnnotatedLine{154}{\ \ \ The\ final\ verification\ summary\ must\ report\ \SkillExplicit{final\ RMSE},\ \SkillImplicit{matched\ observation\ count,\ missing}\ }
\SkillAnnotatedContinuation{\SkillImplicit{observation\ count},\ output\ path,\ first\ timestamp,\ last\ timestamp,\ and\ the\ small\ set\ of\ changed\ }
\SkillAnnotatedContinuation{parameters.\ }
\SkillAnnotatedLine{155}{\mbox{}\ }
\SkillAnnotatedLine{156}{10.\ Handle\ common\ pitfalls\ explicitly.\ }
\SkillAnnotatedLine{157}{\mbox{}\ }
\SkillAnnotatedLine{158}{\ \ \ -\ Use\ \textasciigrave{}python3\textasciigrave{};\ \textasciigrave{}python\textasciigrave{}\ may\ be\ unavailable.\ }
\SkillAnnotatedLine{159}{\ \ \ -\ Skip\ observation\ rows\ containing\ \textasciigrave{}NA\textasciigrave{}.\ }
\SkillAnnotatedLine{160}{\ \ \ -\ Do\ not\ treat\ all\ NetCDF\ array\ entries\ as\ \SkillImplicit{active\ layers}.\ }
\SkillAnnotatedLine{161}{\ \ \ -\ Identify\ active\ GLM\ layers\ by\ positive\ volume.\ }
\SkillAnnotatedLine{162}{\ \ \ -\ Exclude\ huge\ \SkillImplicit{inactive-layer\ fill\ values}\ before\ interpolation.\ }
\SkillAnnotatedLine{163}{\ \ \ -\ Bound\ plausible\ temperatures,\ for\ example\ \SkillLocal{\textasciigrave{}abs(temp)\ \textless{}\ 100\textasciigrave{}}.\ }
\SkillAnnotatedLine{164}{\ \ \ -\ Convert\ model\ \textasciigrave{}z\textasciigrave{}\ to\ \SkillImplicit{depth\ below\ the\ moving\ simulated\ surface}.\ }
\SkillAnnotatedLine{165}{\ \ \ -\ Verify\ \SkillGeneral{final\ artifacts\ only\ after\ the\ final\ rerun}.\ }
\begin{SkillAnnotationBox}{skillNeutralFg}{Remaining limitation}Coverage is broad, but much of the skill remains a checklist; it gives less evidence-backed explanation for why a failure occurs or which residual should select the next parameter.\end{SkillAnnotationBox}
\end{SkillAnnotatedFileBox}
\Needspace{0.24\textheight}
\subsection{SkillAlchemy}
More implicit requirements and reusable synthesis, with task-local bindings kept in scope. \textcolor{gray}{(254 source lines.)}\par
\begin{SkillAnnotatedFileBox}{SKILL.md (SkillAlchemy)}{skill:5}
\SkillAnnotatedLine{1}{---\ }
\SkillAnnotatedLine{2}{name:\ glm-lake-mendota\ }
\SkillAnnotatedLine{3}{description:\ \textgreater{}-\ }
\SkillAnnotatedLine{4}{\ \ Runs\ the\ General\ Lake\ Model\ (GLM\ 3.x)\ to\ \SkillExplicit{simulate\ vertical\ lake\ water\ temperature}\ and\ }
\SkillAnnotatedLine{5}{\ \ calibrates\ its\ parameters\ until\ sim-vs-observation\ \SkillExplicit{RMSE\ drops\ below\ a\ target}\ (e.g.\ \textless{}\ 2\ degC).\ }
\begin{SkillAnnotationBox}{skillExplicitFg}{Sparse task anchor}Only the core objective is blue. Paths, commands, and validation mechanisms are classified by where they come from, not recolored blue merely because they help execute the task.\end{SkillAnnotationBox}
\SkillAnnotatedLine{6}{\ \ Covers\ the\ whole\ closed\ loop:\ edit\ glm3.nml,\ run\ glm,\ extract\ the\ (time,\ depth)\ temperature\ }
\SkillAnnotatedLine{7}{\ \ field\ from\ output.nc,\ score\ it\ against\ field\ profiles,\ diagnose\ the\ residual,\ pick\ the\ next\ }
\SkillAnnotatedLine{8}{\ \ knob.\ Use\ for\ the\ Lake\ Mendota\ task\ when\ the\ visible\ context\ provides\ glm3.nml,\ output.nc,\ }
\SkillAnnotatedLine{9}{\ \ field\ observations,\ a\ date\ window,\ and\ an\ RMSE\ target.\ For\ another\ GLM\ temperature\ task,\ }
\SkillAnnotatedLine{10}{\ \ \SkillGeneral{reuse\ only\ the\ general\ loop\ after\ its\ task-local\ bindings\ and\ model\ conventions\ are\ revalidated}.\ }
\SkillAnnotatedLine{11}{version:\ 0.1.0\ }
\SkillAnnotatedLine{12}{---\ }
\SkillAnnotatedLine{13}{\mbox{}\ }
\SkillAnnotatedLine{14}{\#\ GLM\ Lake\ Temperature\ Simulation\ \&\ Calibration\ }
\SkillAnnotatedLine{15}{\mbox{}\ }
\SkillAnnotatedLine{16}{A\ GLM\ task\ looks\ like\ "run\ a\ model".\ It\ is\ really\ a\ **\SkillGeneral{closed\ calibration\ loop}**:\ }
\SkillAnnotatedLine{17}{\mbox{}\ }
\SkillAnnotatedLine{18}{\textasciigrave{}\textasciigrave{}\textasciigrave{}\ }
\SkillAnnotatedLine{19}{\SkillGeneral{edit\ glm3.nml\ -\textgreater{}\ run\ glm\ -\textgreater{}\ extract\ temp(time,\ depth)\ -\textgreater{}\ score\ vs\ obs\ -\textgreater{}\ diagnose\ residual\ -\textgreater{}\ next}\ }
\SkillAnnotatedContinuation{\SkillGeneral{knob}\ }
\SkillAnnotatedLine{20}{\textasciigrave{}\textasciigrave{}\textasciigrave{}\ }
\SkillAnnotatedLine{21}{\mbox{}\ }
\SkillAnnotatedLine{22}{The\ commands\ below\ instantiate\ this\ loop\ for\ the\ visible\ Lake\ Mendota\ task.\ Treat\ its\ paths,\ dates,\ }
\SkillAnnotatedLine{23}{target,\ and\ candidate\ parameters\ as\ **\SkillLocal{task-local\ bindings,\ not\ reusable\ defaults}**.\ For\ another\ GLM\ }
\SkillAnnotatedLine{24}{setup,\ \SkillGeneral{re-identify\ and\ validate\ those\ bindings\ before\ following\ the\ general\ loop}.\ }
\begin{SkillAnnotationBox}{skillLocalFg}{Task-scoped instantiation}Concrete paths, dates, targets, and candidate parameters bind the general loop to the visible task and must be re-established before reuse.\end{SkillAnnotationBox}
\SkillAnnotatedLine{25}{\mbox{}\ }
\SkillAnnotatedLine{26}{**The\ three\ facts\ that\ decide\ whether\ you\ succeed.**\ Internalise\ them\ before\ touching\ anything:\ }
\SkillAnnotatedLine{27}{\mbox{}\ }
\SkillAnnotatedLine{28}{1.\ **\SkillImplicit{\textasciigrave{}z\textasciigrave{}\ in\ \textasciigrave{}output.nc\textasciigrave{}\ is\ bottom-up,\ and\ it\ is\ a\ HEIGHT,\ not\ a\ depth}.**\ Layer\ index\ \textasciigrave{}0\textasciigrave{}\ is\ the\ }
\SkillAnnotatedLine{29}{\ \ \ lake\ **bottom**;\ index\ \textasciigrave{}NS[t]-1\textasciigrave{}\ is\ the\ **surface**.\ Reading\ \textasciigrave{}temp[t,0]\textasciigrave{}\ as\ "surface\ }
\SkillAnnotatedLine{30}{\ \ \ temperature"\ is\ a\ likely\ way\ to\ fail\ this\ task.\ The\ mistake\ can\ return\ a\ plausible\ but\ incorrect\ }
\SkillAnnotatedLine{31}{\ \ \ temperature\ and\ RMSE,\ with\ no\ exception,\ no\ warning,\ and\ no\ obvious\ sign\ that\ the\ depth\ }
\SkillAnnotatedLine{32}{\ \ \ convention\ was\ misread.\ }
\SkillAnnotatedLine{33}{2.\ **\SkillImplicit{GLM's\ exit\ code\ is\ not\ a\ success\ signal}.**\ Its\ \textasciigrave{}main()\textasciigrave{}\ ends\ in\ an\ unconditional\ \textasciigrave{}exit(0)\textasciigrave{}.\ }
\SkillAnnotatedLine{34}{\ \ \ Non-zero\ reliably\ means\ failure;\ **zero\ means\ nothing\ at\ all**.\ Success\ has\ to\ be\ *defined*\ as:\ }
\SkillAnnotatedLine{35}{\ \ \ \SkillImplicit{\textasciigrave{}output.nc\textasciigrave{}\ exists}\ **and**\ its\ \SkillImplicit{\textasciigrave{}time\textasciigrave{}\ axis\ spans\ the\ requested\ window}\ **and**\ \SkillImplicit{\textasciigrave{}temp\textasciigrave{}\ is\ finite}.\ }
\SkillAnnotatedLine{36}{3.\ **\SkillImplicit{Relative\ paths\ in\ the\ nml\ bind\ to\ the\ process\ CWD}**,\ not\ to\ the\ nml's\ own\ location\ ---\ GLM's\ C\ }
\SkillAnnotatedLine{37}{\ \ \ source\ contains\ no\ path-resolution\ code\ whatsoever.\ \textasciigrave{}out\_dir='output'\textasciigrave{}\ means\ }
\SkillAnnotatedLine{38}{\ \ \ \textasciigrave{}\$PWD/output\textasciigrave{}.\ **\SkillGeneral{Always\ \textasciigrave{}cd\textasciigrave{}\ to\ the\ nml's\ directory\ before\ running}.**\ }
\begin{SkillAnnotationBox}{skillImplicitFg}{Implicit requirements beyond the prompt}Coordinate semantics, unreliable exit codes, and CWD-dependent paths are not visible in the task brief, yet each can produce plausible-looking false success.\end{SkillAnnotationBox}
\SkillAnnotatedLine{39}{\mbox{}\ }
\SkillAnnotatedLine{40}{---\ }
\SkillAnnotatedLine{41}{\mbox{}\ }
\SkillAnnotatedLine{42}{\#\#\ Activation\ Rules\ }
\SkillAnnotatedLine{43}{\mbox{}\ }
\SkillAnnotatedLine{44}{**Apply\ these\ instructions\ when:**\ }
\SkillAnnotatedLine{45}{-\ The\ visible\ task\ concerns\ \SkillLocal{Lake\ Mendota}\ temperature\ simulation\ or\ vertical-profile\ calibration\ }
\SkillAnnotatedLine{46}{-\ It\ provides\ a\ \textasciigrave{}glm3.nml\textasciigrave{},\ forcing\ directory,\ observation\ file,\ expected\ output,\ and\ date\ window\ }
\SkillAnnotatedLine{47}{-\ It\ specifies\ a\ calibration\ metric\ and\ target\ for\ the\ delivered\ GLM\ artifacts\ }
\SkillAnnotatedLine{48}{-\ For\ another\ GLM\ setup,\ \SkillGeneral{reuse\ only\ the\ loop\ after\ rebinding\ and\ validating\ every\ task-local\ input}\ }
\SkillAnnotatedLine{49}{-\ Post-hoc\ analysis\ of\ a\ GLM\ run:\ extracting\ temperature,\ comparing\ to\ field\ profiles,\ plotting\ }
\SkillAnnotatedContinuation{profiles\ }
\SkillAnnotatedLine{50}{\mbox{}\ }
\SkillAnnotatedLine{51}{**Do\ NOT\ trigger\ on:**\ }
\SkillAnnotatedLine{52}{-\ Generic\ NetCDF\ reading\ with\ no\ lake\ model\ involved\ \ensuremath{\rightarrow}\ just\ use\ \textasciigrave{}netCDF4\textasciigrave{}/\textasciigrave{}xarray\textasciigrave{}\ directly\ }
\SkillAnnotatedLine{53}{-\ Other\ lake/reservoir\ models\ (FLake,\ Simstrat,\ GOTM,\ MyLake,\ CE-QUAL-W2)\ ---\ the\ nml,\ the\ output\ }
\SkillAnnotatedLine{54}{\ \ layout\ and\ the\ calibration\ knobs\ are\ all\ different.\ \SkillGeneral{Only\ the\ *loop\ shape*\ transfers}.\ }
\begin{SkillAnnotationBox}{skillGeneralFg}{Scope boundary}The transferable claim is explicitly limited to the loop structure; task paths, schemas, dates, and calibration values must be revalidated in a new setup.\end{SkillAnnotationBox}
\SkillAnnotatedLine{55}{-\ Water\ **quality**/biogeochemistry\ calibration\ (AED2:\ oxygen,\ nutrients,\ chlorophyll).\ }
\SkillAnnotatedContinuation{Temperature\ }
\SkillAnnotatedLine{56}{\ \ is\ calibrated\ first\ and\ frozen;\ this\ skill\ stops\ there.\ }
\SkillAnnotatedLine{57}{-\ Statistical\ /\ ML\ lake-temperature\ prediction\ (LSTM,\ process-guided\ DL)\ ---\ no\ GLM\ binary\ in\ the\ }
\SkillAnnotatedContinuation{loop\ }
\SkillAnnotatedLine{58}{-\ Pure\ hydrology\ (streamflow,\ rainfall-runoff)\ with\ no\ lake\ thermal\ structure\ }
\SkillAnnotatedLine{59}{\mbox{}\ }
\SkillAnnotatedLine{60}{---\ }
\SkillAnnotatedLine{61}{\mbox{}\ }
\SkillAnnotatedLine{62}{\#\#\ Agentic\ Protocol\ }
\SkillAnnotatedLine{63}{\mbox{}\ }
\SkillAnnotatedLine{64}{\SkillGeneral{Work\ the\ loop\ in\ this\ order}.\ **Do\ not\ skip\ Step\ 1\ or\ Step\ 3**\ ---\ they\ are\ the\ two\ places\ where\ }
\SkillAnnotatedLine{65}{silence\ is\ mistaken\ for\ success.\ }
\SkillAnnotatedLine{66}{\mbox{}\ }
\SkillAnnotatedLine{67}{**Step\ 0\ ---\ \SkillGeneral{Read\ the\ model\ card\ before\ acting}.**\ Read\ \textasciigrave{}references/sop\_models.md\textasciigrave{}\ and\ match\ your\ }
\SkillAnnotatedLine{68}{current\ stage\ to\ a\ card\ (H1--H9).\ The\ cards\ carry\ the\ evidence,\ the\ failure\ modes,\ and\ the\ exact\ }
\SkillAnnotatedLine{69}{commands.\ Come\ back\ here\ for\ the\ sequencing.\ }
\SkillAnnotatedLine{70}{\mbox{}\ }
\SkillAnnotatedLine{71}{**Step\ 1\ ---\ \SkillImplicit{Establish\ ground\ truth\ about\ the\ environment}\ (never\ assume\ it).**\ }
\SkillAnnotatedLine{72}{\textasciigrave{}\textasciigrave{}\textasciigrave{}bash\ }
\SkillAnnotatedLine{73}{\SkillLocal{ldd\ /usr/local/bin/glm}\ |\ grep\ -i\ "not\ found"\ \ \ \ \ \#\ missing\ libnetcdf.so.X?\ -\textgreater{}\ troubleshooting.md\ }
\SkillAnnotatedLine{74}{\SkillLocal{head\ -3\ /root/bcs/meteo.csv\ /root/field\_temp\_oxy.csv}\ }
\SkillAnnotatedLine{75}{\SkillLocal{python3\ scripts/nml\_edit.py\ dump\ --nml\ /root/glm3.nml}\ \ \ \ \ \#\ what\ is\ actually\ in\ the\ config\ }
\SkillAnnotatedLine{76}{\textasciigrave{}\textasciigrave{}\textasciigrave{}\ }
\SkillAnnotatedLine{77}{A\ missing\ shared\ library\ is\ an\ **environment**\ \SkillGeneral{problem,\ not\ a\ modelling\ problem}.\ Fix\ it\ before\ }
\SkillAnnotatedLine{78}{anything\ else,\ or\ every\ run\ will\ "succeed"\ and\ write\ nothing.\ }
\SkillAnnotatedLine{79}{\mbox{}\ }
\SkillAnnotatedLine{80}{**Step\ 2\ ---\ Get\ one\ \SkillGeneral{baseline\ run\ to\ completion,\ from\ the\ right\ directory}.**\ }
\SkillAnnotatedLine{81}{\textasciigrave{}\textasciigrave{}\textasciigrave{}bash\ }
\SkillAnnotatedLine{82}{python3\ scripts/run\_glm.py\ --nml\ /root/glm3.nml\ --glm\ /usr/local/bin/glm\ }
\SkillAnnotatedLine{83}{\textasciigrave{}\textasciigrave{}\textasciigrave{}\ }
\SkillAnnotatedLine{84}{This\ \textasciigrave{}cd\textasciigrave{}s\ to\ \textasciigrave{}/root\textasciigrave{}\ for\ you\ and\ then\ **\SkillImplicit{verifies\ the\ artefact\ instead\ of\ the\ exit\ code}**\ ---\ it\ }
\SkillAnnotatedLine{85}{recomputes\ the\ \SkillImplicit{expected\ record\ count}\ from\ \textasciigrave{}start\textasciigrave{}/\textasciigrave{}stop\textasciigrave{}/\textasciigrave{}dt\textasciigrave{}/\textasciigrave{}nsave\textasciigrave{}\ and\ checks\ the\ \SkillImplicit{actual\ \textasciigrave{}time\textasciigrave{}}\ }
\SkillAnnotatedLine{86}{\SkillImplicit{axis\ against\ it}.\ Do\ not\ hand-roll\ \textasciigrave{}glm\textasciigrave{}\ invocations;\ the\ CWD\ trap\ is\ why\ this\ script\ exists.\ }
\SkillAnnotatedLine{87}{\mbox{}\ }
\SkillAnnotatedLine{88}{**Step\ 3\ ---\ \SkillGeneral{Introspect\ \textasciigrave{}output.nc\textasciigrave{}\ before\ you\ trust\ any\ recipe}\ about\ it.**\ }
\SkillAnnotatedLine{89}{\textasciigrave{}\textasciigrave{}\textasciigrave{}bash\ }
\SkillAnnotatedLine{90}{python3\ scripts/extract\_temp.py\ --nc\ /root/output/output.nc\ --header\ \ \ \ \#\ or:\ ncdump\ -h\ }
\SkillAnnotatedLine{91}{\textasciigrave{}\textasciigrave{}\textasciigrave{}\ }
\SkillAnnotatedLine{92}{Confirm\ you\ see\ \SkillImplicit{\textasciigrave{}NS(time)\textasciigrave{},\ \textasciigrave{}z(time,z,lat,lon)\textasciigrave{},\ \textasciigrave{}temp(time,z,lat,lon)\textasciigrave{},\ and\ \textasciigrave{}time\textasciigrave{}}\ with\ }
\SkillAnnotatedLine{93}{\textasciigrave{}units\ =\ "hours\ since\ ..."\textasciigrave{}.\ If\ the\ names\ differ,\ your\ GLM\ is\ a\ different\ build\ ---\ consult\ the\ }
\SkillAnnotatedLine{94}{version\ table\ in\ \textasciigrave{}references/sop\_models.md\textasciigrave{}\ (H3)\ \SkillGeneral{rather\ than\ guessing}.\ }
\begin{SkillAnnotationBox}{skillGeneralFg}{Evidence-to-protocol synthesis}Concrete environment and schema findings are converted into a reusable order of operations rather than copied as universal constants.\end{SkillAnnotationBox}
\SkillAnnotatedLine{95}{\mbox{}\ }
\SkillAnnotatedLine{96}{**Step\ 4\ ---\ Score\ the\ baseline\ under\ the\ fixed\ metric.**\ }
\SkillAnnotatedLine{97}{\textasciigrave{}\textasciigrave{}\textasciigrave{}bash\ }
\SkillAnnotatedLine{98}{python3\ \SkillLocal{scripts/eval\_rmse.py}\ --nc\ /root/output/output.nc\ --obs\ /root/field\_temp\_oxy.csv\ --target\ }
\SkillAnnotatedContinuation{2.0\ }
\SkillAnnotatedLine{99}{\textasciigrave{}\textasciigrave{}\textasciigrave{}\ }
\SkillAnnotatedLine{100}{This\ prints\ the\ pooled\ RMSE,\ the\ bias,\ and\ the\ breakdown\ **\SkillImplicit{by\ depth\ band\ and\ by\ season}**.\ The\ }
\SkillAnnotatedLine{101}{breakdown\ is\ not\ decoration\ ---\ it\ is\ the\ input\ to\ Step\ 5.\ A\ whole-column\ RMSE\ that\ looks\ fine\ while\ }
\SkillAnnotatedLine{102}{the\ hypolimnion\ is\ 3\ \textdegree{}C\ off\ is\ the\ normal\ state\ of\ an\ uncalibrated\ GLM,\ and\ the\ aggregate\ hides\ }
\SkillAnnotatedContinuation{it.\ }
\SkillAnnotatedLine{103}{\mbox{}\ }
\SkillAnnotatedLine{104}{**Step\ 5\ ---\ \SkillGeneral{Diagnose\ the\ residual,\ then\ pick\ the\ knob}.**\ Use\ the\ mapping\ table\ below\ (full\ version,\ }
\SkillAnnotatedLine{105}{with\ the\ physics\ and\ the\ citations,\ in\ H7).\ **\SkillGeneral{Do\ not\ grid-search}.**\ Each\ residual\ pattern\ points\ }
\SkillAnnotatedContinuation{at\ }
\SkillAnnotatedLine{106}{a\ specific,\ \SkillGeneral{physically-motivated\ parameter};\ move\ that\ one.\ }
\begin{SkillAnnotationBox}{skillGeneralFg}{Diagnostic generalization}The skill turns residual structure into a principled next action instead of promoting one local parameter setting.\end{SkillAnnotationBox}
\SkillAnnotatedLine{107}{\mbox{}\ }
\SkillAnnotatedLine{108}{**Step\ 6\ ---\ \SkillGeneral{Calibrate,\ staged\ and\ bounded}.**\ }
\SkillAnnotatedLine{109}{\textasciigrave{}\textasciigrave{}\textasciigrave{}bash\ }
\SkillAnnotatedLine{110}{python3\ scripts/calibrate.py\ --nml\ /root/glm3.nml\ --obs\ /root/field\_temp\_oxy.csv\ \textbackslash{}\ }
\SkillAnnotatedLine{111}{\ \ \ \ --params\ \SkillLocal{ch,lw\_factor,sw\_factor,sed\_temp\_mean}\ --target\ 2.0\ --calib-frac\ 0.5\ }
\SkillAnnotatedLine{112}{\textasciigrave{}\textasciigrave{}\textasciigrave{}\ }
\SkillAnnotatedLine{113}{For\ this\ Lake\ Mendota\ task,\ the\ command\ uses\ four\ parameters\ selected\ by\ a\ \SkillLocal{Morris\ sensitivity\ screen}\ }
\SkillAnnotatedLine{114}{(Ladwig\ et\ al.,\ HESS\ 2021).\ It\ iterates\ in\ a\ \SkillGeneral{scratch\ directory},\ then\ **\SkillGeneral{writes\ the\ winning}\ }
\SkillAnnotatedLine{115}{\SkillGeneral{parameters\ back\ into\ the\ real\ nml\ and\ re-runs\ in\ place}**\ ---\ so\ the\ delivered\ nml\ genuinely\ }
\SkillAnnotatedLine{116}{reproduces\ the\ delivered\ \textasciigrave{}output.nc\textasciigrave{}.\ Runs\ are\ seconds-scale;\ dozens\ of\ iterations\ are\ cheap.\ }
\SkillAnnotatedLine{117}{\mbox{}\ }
\SkillAnnotatedLine{118}{**Step\ 7\ ---\ \SkillGeneral{Gate\ the\ delivery}.\ This\ is\ the\ last\ thing\ you\ do.**\ }
\SkillAnnotatedLine{119}{\textasciigrave{}\textasciigrave{}\textasciigrave{}bash\ }
\SkillAnnotatedLine{120}{python3\ scripts/verify\_delivery.py\ --nc\ /root/output/output.nc\ --nml\ /root/glm3.nml\ \textbackslash{}\ }
\SkillAnnotatedLine{121}{\ \ \ \ --obs\ /root/field\_temp\_oxy.csv\ --start\ \SkillExplicit{2009-01-01}\ --stop\ \SkillExplicit{2015-12-30}\ --\SkillExplicit{target\ 2.0}\ }
\SkillAnnotatedLine{122}{\textasciigrave{}\textasciigrave{}\textasciigrave{}\ }
\SkillAnnotatedLine{123}{Five\ checks:\ the\ \SkillImplicit{file\ exists};\ the\ \SkillImplicit{time\ axis\ covers\ the\ window};\ \SkillImplicit{\textasciigrave{}temp\textasciigrave{}\ is\ finite\ and\ physical};\ }
\SkillAnnotatedContinuation{**the\ }
\SkillAnnotatedLine{124}{\SkillImplicit{final\ nml\ re-runs\ cleanly}\ and\ reproduces\ that\ time\ axis**;\ \SkillExplicit{RMSE\ is\ under\ target}.\ Exit\ 0\ only\ if\ }
\SkillAnnotatedContinuation{all\ }
\begin{SkillAnnotationBox}{skillImplicitFg}{Delivery requirements discovered}The task asks for a runnable final nml, but not the full artifact, coverage, physical-validity, and clean-rerun gate needed to establish that claim.\end{SkillAnnotationBox}
\SkillAnnotatedLine{125}{five\ pass.\ If\ you\ did\ any\ iterating\ outside\ \textasciigrave{}/root\textasciigrave{},\ this\ is\ what\ catches\ the\ fact\ that\ you\ never\ }
\SkillAnnotatedLine{126}{copied\ the\ answer\ back.\ }
\SkillAnnotatedLine{127}{\mbox{}\ }
\SkillAnnotatedLine{128}{---\ }
\SkillAnnotatedLine{129}{\mbox{}\ }
\SkillAnnotatedLine{130}{\#\#\ Core\ Operation\ Models\ }
\SkillAnnotatedLine{131}{\mbox{}\ }
\SkillAnnotatedLine{132}{|\ \#\ |\ Model\ |\ Core\ proposition\ |\ Source\ |\ }
\SkillAnnotatedLine{133}{|---|-------|------------------|--------|\ }
\SkillAnnotatedLine{134}{|\ H1\ |\ **Run\ GLM\ Without\ Losing\ Your\ Output**\ |\ \SkillGeneral{Relative\ nml\ paths\ bind\ to\ CWD},\ never\ to\ the\ nml.\ }
\SkillAnnotatedContinuation{\textasciigrave{}cd\textasciigrave{}\ to\ the\ nml\ dir.\ \textasciigrave{}out\_dir\textasciigrave{}\ is\ created\ with\ \textasciigrave{}mkdir(2)\textasciigrave{}\ ---\ one\ level\ only\ |\ GLM\ C\ source\ }
\SkillAnnotatedContinuation{(\textasciigrave{}glm\_main.c\textasciigrave{},\ no\ \textasciigrave{}chdir\textasciigrave{}/\textasciigrave{}realpath\textasciigrave{}\ anywhere)\ |\ }
\SkillAnnotatedLine{135}{|\ H2\ |\ **Never\ Trust\ GLM's\ Exit\ Code**\ |\ \textasciigrave{}main()\textasciigrave{}\ ends\ in\ unconditional\ \textasciigrave{}exit(0)\textasciigrave{}.\ \SkillGeneral{Define\ success}\ }
\SkillAnnotatedContinuation{\SkillGeneral{as\ artefact\ +\ full\ time\ axis\ +\ finite\ temp}\ |\ GLM\ \textasciigrave{}src/glm\_main.c\textasciigrave{};\ LakeEnsemblR\ issues\ \#279,\ \#288\ }
\SkillAnnotatedContinuation{|\ }
\SkillAnnotatedLine{136}{|\ H3\ |\ **The\ Lagrangian\ Layer\ Trap**\ \textbf{!}\ |\ \SkillGeneral{\textasciigrave{}z\textasciigrave{}\ =\ layer-TOP\ height\ above\ the\ **bottom**,\ bottom-up}.\ }
\SkillAnnotatedContinuation{Index\ 0\ =\ bottom,\ \textasciigrave{}NS[t]-1\textasciigrave{}\ =\ surface.\ Slice\ by\ \textasciigrave{}NS[t]\textasciigrave{};\ padding\ is\ \ensuremath{\geq}1e30\ and\ \textasciigrave{}\_FillValue\textasciigrave{}\ is\ }
\SkillAnnotatedContinuation{compile-time\ optional\ |\ GLM\ \textasciigrave{}glm\_ncdf.c\textasciigrave{}\ +\ glmtools\ +\ glm-py,\ verified\ numerically\ on\ a\ real\ }
\SkillAnnotatedContinuation{\textasciigrave{}output.nc\textasciigrave{}\ |\ }
\SkillAnnotatedLine{137}{|\ H4\ |\ **\SkillGeneral{Midpoint\ Interpolation\ to\ a\ Depth\ Grid}**\ |\ Temperature\ attaches\ to\ layer\ **midpoints**,\ }
\SkillAnnotatedContinuation{not\ tops.\ Nodes\ \textasciigrave{}[0]+mids+[z\_surf]\textasciigrave{},\ values\ \textasciigrave{}[T\ensuremath{_{0}}]+T+[T\_last]\textasciigrave{},\ linear\ interp,\ out-of-column\ \ensuremath{\rightarrow}\ }
\SkillAnnotatedContinuation{**NaN\ not\ clamped**\ |\ glmtools\ \textasciigrave{}resample\_depth()\textasciigrave{}\ in\ \textasciigrave{}R/get\_var.R\textasciigrave{}\ |\ }
\SkillAnnotatedLine{138}{|\ H5\ |\ **The\ \SkillGeneral{Fixed\ RMSE\ Convention}**\ |\ Pooled\ over\ all\ matched\ (time,\ depth);\ sim\ interpolated\ }
\SkillAnnotatedContinuation{**in\ depth\ onto\ obs\ depths**;\ time\ matched\ at\ calendar-day\ precision;\ unmatched\ dropped;\ ice\ and\ }
\SkillAnnotatedContinuation{surface\ **not**\ excluded\ |\ glmtools\ \textasciigrave{}dot\_compare\_to\_field.R\textasciigrave{}\ +\ \textasciigrave{}calib\_helpers.R\textasciigrave{};\ LakeEnsemblR\ }
\SkillAnnotatedContinuation{\textasciigrave{}calc\_fit.R\textasciigrave{}\ |\ }
\SkillAnnotatedLine{139}{|\ H6\ |\ **\SkillGeneral{Staged\ Calibration,\ Not\ Grid\ Search}**\ |\ Screen\ \ensuremath{\rightarrow}\ cap\ at\ 4--6\ params\ \ensuremath{\rightarrow}\ heat\ budget\ \ensuremath{\rightarrow}\ }
\SkillAnnotatedContinuation{light/structure\ (\textasciigrave{}Kw\textasciigrave{})\ \ensuremath{\rightarrow}\ bottom\ (\textasciigrave{}sed\_temp\_mean\textasciigrave{})\ \ensuremath{\rightarrow}\ mixing\ efficiencies\ last\ |\ Ladwig\ et\ al.\ HESS\ }
\SkillAnnotatedContinuation{2021;\ Feldbauer\ et\ al.\ HESS\ 2025;\ Bruce\ et\ al.\ 2018\ |\ }
\SkillAnnotatedLine{140}{|\ H7\ |\ **\SkillGeneral{Residual\ Diagnosis\ \ensuremath{\rightarrow}\ Knob\ Selection}**\ |\ Each\ residual\ pattern\ maps\ to\ a\ specific\ knob\ }
\SkillAnnotatedContinuation{with\ a\ physical\ justification.\ Read\ the\ residual,\ don't\ sweep\ the\ space\ |\ R03\ synthesis\ of\ HESS\ }
\SkillAnnotatedContinuation{2025\ /\ Bruce\ 2018\ /\ glmGUI\ 2020\ |\ }
\SkillAnnotatedLine{141}{|\ H8\ |\ **Know\ When\ To\ Stop**\ |\ \ensuremath{\geq}3\ \textdegree{}C\ =\ broken.\ \textasciitilde{}2\ \textdegree{}C\ =\ passing\ but\ mediocre.\ \SkillLocal{1.3--1.6\ \textdegree{}C\ =\ good\ for}\ }
\SkillAnnotatedContinuation{\SkillLocal{Mendota}.\ **\textless{}1.0\ \textdegree{}C\ =\ suspicious**\ |\ Bruce\ (Mendota\ 1.60);\ Ladwig\ (1.96\ total,\ surface\ 1.30,\ bottom\ }
\SkillAnnotatedContinuation{2.43)\ |\ }
\SkillAnnotatedLine{142}{|\ H9\ |\ **\SkillGeneral{Editing\ the\ nml\ Safely}**\ |\ Fortran\ namelist,\ \textasciigrave{}!\textasciigrave{}\ comments,\ array-length\ rule\ is\ \textasciigrave{}len\ \textgreater{}=\ }
\SkillAnnotatedContinuation{count\textasciigrave{}.\ Surgical\ key-value\ edits\ +\ round-trip\ verification.\ Never\ regex-rewrite\ wholesale\ |\ GLM\ }
\SkillAnnotatedContinuation{\textasciigrave{}glm\_init.c\textasciigrave{};\ AED\ config\ docs\ |\ }
\begin{SkillAnnotationBox}{skillGeneralFg}{Operation models}The evidence is consolidated into reusable models for execution, scoring, calibration, and safe editing.\end{SkillAnnotationBox}
\SkillAnnotatedLine{143}{\mbox{}\ }
\SkillAnnotatedLine{144}{Full\ cards\ ---\ with\ inputs,\ exact\ actions,\ evidence\ URLs,\ failure\ modes\ and\ confidence\ ---\ in\ }
\SkillAnnotatedLine{145}{\textasciigrave{}references/sop\_models.md\textasciigrave{}.\ **H3\ is\ the\ one\ that\ decides\ the\ task.**\ }
\SkillAnnotatedLine{146}{\mbox{}\ }
\SkillAnnotatedLine{147}{\#\#\#\ Residual\ \ensuremath{\rightarrow}\ knob\ (the\ short\ version)\ }
\SkillAnnotatedLine{148}{\mbox{}\ }
\SkillAnnotatedLine{149}{|\ \SkillGeneral{What\ the\ residual\ looks\ like\ |\ Move\ this\ |\ Which\ way}\ |\ }
\SkillAnnotatedLine{150}{|---|---|---|\ }
\SkillAnnotatedLine{151}{|\ \SkillGeneral{Warm/cold\ bias\ at\ **all**\ depths}\ |\ \textasciigrave{}sw\_factor\textasciigrave{},\ \textasciigrave{}lw\_factor\textasciigrave{}\ |\ down\ if\ warm,\ up\ if\ cold\ |\ }
\SkillAnnotatedLine{152}{|\ **\SkillGeneral{Surface/epilimnion}**\ too\ warm\ in\ summer\ |\ \textasciigrave{}ce\textasciigrave{}\ (latent\ heat\ loss),\ then\ \textasciigrave{}ch\textasciigrave{};\ \textasciigrave{}wind\_factor\textasciigrave{}\ |\ }
\SkillAnnotatedContinuation{up\ |\ }
\SkillAnnotatedLine{153}{|\ **Surface**\ too\ cold\ |\ \textasciigrave{}ch\textasciigrave{}/\textasciigrave{}ce\textasciigrave{}\ down,\ \textasciigrave{}sw\_factor\textasciigrave{}\ up\ |\ ---\ |\ }
\SkillAnnotatedLine{154}{|\ **\SkillGeneral{Thermocline\ too\ deep}**\ /\ not\ enough\ stratification\ |\ \textasciigrave{}Kw\textasciigrave{}\ **first**;\ then\ \textasciigrave{}wind\_factor\textasciigrave{},\ \textasciigrave{}cd\textasciigrave{}\ }
\SkillAnnotatedContinuation{down\ |\ \textasciigrave{}Kw\textasciigrave{}\ up\ |\ }
\SkillAnnotatedLine{155}{|\ **\SkillGeneral{Thermocline\ too\ shallow}**\ /\ won't\ mix\ down\ |\ \textasciigrave{}Kw\textasciigrave{}\ down;\ then\ \textasciigrave{}wind\_factor\textasciigrave{},\ \textasciigrave{}coef\_mix\_shear\textasciigrave{}\ }
\SkillAnnotatedContinuation{up\ |\ ---\ |\ }
\SkillAnnotatedLine{156}{|\ **\SkillGeneral{Hypolimnion/bottom\ too\ cold}**\ |\ \textasciigrave{}sed\_temp\_mean\textasciigrave{}\ (deep\ zone,\ 3--8\ \textdegree{}C);\ then\ \textasciigrave{}coef\_mix\_hyp\textasciigrave{}\ |\ up\ }
\SkillAnnotatedContinuation{|\ }
\SkillAnnotatedLine{157}{|\ **\SkillGeneral{Hypolimnion\ too\ warm}**\ |\ \textasciigrave{}sed\_temp\_mean\textasciigrave{}\ down;\ \textasciigrave{}coef\_mix\_hyp\textasciigrave{}\ down\ |\ down\ |\ }
\SkillAnnotatedLine{158}{|\ **\SkillGeneral{Winter\ /\ under-ice}**\ bias\ |\ \textasciigrave{}sed\_temp\_mean\textasciigrave{}\ (dominant\ under\ ice),\ \textasciigrave{}lw\_factor\textasciigrave{}\ |\ ---\ |\ }
\SkillAnnotatedLine{159}{|\ \SkillGeneral{Turnover\ too\ early/late}\ |\ \textasciigrave{}wind\_factor\textasciigrave{},\ \textasciigrave{}coef\_mix\_conv\textasciigrave{}\ |\ up\ \ensuremath{\Rightarrow}\ earlier\ |\ }
\begin{SkillAnnotationBox}{skillGeneralFg}{Residual-to-action abstraction}Error patterns are mapped to physically motivated parameter families, increasing transfer beyond the original run.\end{SkillAnnotationBox}
\SkillAnnotatedLine{160}{\mbox{}\ }
\SkillAnnotatedLine{161}{Physics,\ directions,\ bounds\ and\ citations\ for\ every\ row:\ H7\ in\ \textasciigrave{}references/sop\_models.md\textasciigrave{}.\ }
\SkillAnnotatedLine{162}{\mbox{}\ }
\SkillAnnotatedLine{163}{---\ }
\SkillAnnotatedLine{164}{\mbox{}\ }
\SkillAnnotatedLine{165}{\#\#\ Output\ Style\ }
\SkillAnnotatedLine{166}{\mbox{}\ }
\SkillAnnotatedLine{167}{-\ Lead\ with\ the\ metric\ and\ verdict:\ "Pooled\ RMSE\ is\ below\ the\ task-defined\ target;\ the\ dominant\ }
\SkillAnnotatedLine{168}{\ \ bias\ is\ in\ the\ hypolimnion."\ Then\ explain.\ }
\SkillAnnotatedLine{169}{-\ Show\ the\ command\ you\ ran\ and\ the\ output\ that\ matters.\ Not\ the\ whole\ log.\ }
\SkillAnnotatedLine{170}{-\ Name\ the\ knob\ and\ the\ reason\ together:\ "the\ bottom\ is\ too\ warm,\ so\ \textasciigrave{}sed\_temp\_mean\textasciigrave{}\ comes\ down"\ ---\ }
\SkillAnnotatedLine{171}{\ \ never\ "let\ me\ tune\ some\ parameters".\ }
\SkillAnnotatedLine{172}{-\ When\ you\ cite\ a\ fact,\ cite\ it\ the\ way\ a\ person\ would:\ "Ladwig's\ Mendota\ calibration\ got\ surface\ }
\SkillAnnotatedLine{173}{\ \ 1.30\ \textdegree{}C\ but\ bottom\ 2.43\ \textdegree{}C"\ ---\ not\ "according\ to\ H8\ in\ sop\_models.md".\ }
\SkillAnnotatedLine{174}{-\ State\ uncertainty\ where\ the\ evidence\ is\ thin\ (the\ under-ice\ knob\ mapping\ is\ physics-derived,\ not\ }
\SkillAnnotatedLine{175}{\ \ measured).\ \SkillGeneral{Do\ not\ launder\ inference\ into\ fact}.\ }
\SkillAnnotatedLine{176}{-\ Banned:\ "let\ me\ systematically\ analyse...",\ "according\ to\ the\ model\ card...",\ "would\ you\ like\ me\ }
\SkillAnnotatedLine{177}{\ \ to\ continue?".\ Answer,\ then\ stop.\ }
\SkillAnnotatedLine{178}{\mbox{}\ }
\SkillAnnotatedLine{179}{---\ }
\SkillAnnotatedLine{180}{\mbox{}\ }
\SkillAnnotatedLine{181}{\#\#\ Output\ Modes\ }
\SkillAnnotatedLine{182}{\mbox{}\ }
\SkillAnnotatedLine{183}{|\ Mode\ |\ Triggered\ when\ |\ Structure\ |\ }
\SkillAnnotatedLine{184}{|------|---------------|-----------|\ }
\SkillAnnotatedLine{185}{|\ **\SkillGeneral{Diagnose}**\ |\ A\ run\ finished\ and\ the\ RMSE\ is\ too\ high\ |\ Pooled\ RMSE\ +\ bias\ \ensuremath{\rightarrow}\ depth-band\ \&\ }
\SkillAnnotatedContinuation{seasonal\ breakdown\ \ensuremath{\rightarrow}\ the\ one\ residual\ pattern\ that\ dominates\ \ensuremath{\rightarrow}\ the\ knob\ it\ implies\ \ensuremath{\rightarrow}\ the\ next\ }
\SkillAnnotatedContinuation{command\ |\ }
\SkillAnnotatedLine{186}{|\ **\SkillGeneral{Bootstrap}**\ |\ Fresh\ environment,\ nothing\ has\ run\ yet\ |\ \textasciigrave{}ldd\textasciigrave{}\ +\ forcing-file\ survey\ +\ nml\ dump\ }
\SkillAnnotatedContinuation{\ensuremath{\rightarrow}\ baseline\ run\ \ensuremath{\rightarrow}\ \textasciigrave{}ncdump\ -h\textasciigrave{}\ \ensuremath{\rightarrow}\ baseline\ RMSE.\ No\ parameter\ changes\ yet\ |\ }
\SkillAnnotatedLine{187}{|\ **\SkillGeneral{Calibrate}**\ |\ Baseline\ is\ sound,\ RMSE\ needs\ to\ come\ down\ |\ The\ parameter\ set\ +\ the\ bounds\ +\ }
\SkillAnnotatedContinuation{why\ those\ params\ \ensuremath{\rightarrow}\ the\ loop\ \ensuremath{\rightarrow}\ the\ iteration\ log\ \ensuremath{\rightarrow}\ best\ params,\ with\ the\ calibration/validation\ }
\SkillAnnotatedContinuation{split\ reported\ |\ }
\SkillAnnotatedLine{188}{|\ **\SkillGeneral{Debug}**\ |\ GLM\ "ran"\ but\ something\ is\ wrong\ or\ missing\ |\ Which\ of\ the\ three\ silent-failure\ }
\SkillAnnotatedContinuation{classes\ it\ is\ (no\ file\ /\ truncated\ /\ all-NaN)\ \ensuremath{\rightarrow}\ the\ confirming\ check\ \ensuremath{\rightarrow}\ the\ fix\ from\ }
\SkillAnnotatedContinuation{\textasciigrave{}troubleshooting.md\textasciigrave{}\ |\ }
\SkillAnnotatedLine{189}{|\ **\SkillGeneral{Verify}**\ |\ About\ to\ declare\ done\ |\ The\ five\ acceptance\ checks,\ each\ PASS/FAIL\ with\ its\ reason.\ }
\SkillAnnotatedContinuation{Never\ claim\ success\ on\ an\ unverified\ artefact\ |\ }
\SkillAnnotatedLine{190}{|\ **Explain**\ |\ User\ asks\ how\ GLM\ or\ the\ layer\ scheme\ works\ |\ Prose,\ grounded\ in\ the\ source.\ Point\ }
\SkillAnnotatedContinuation{to\ \textasciigrave{}references/\textasciigrave{}\ rather\ than\ pasting\ cards\ |\ }
\SkillAnnotatedLine{191}{\mbox{}\ }
\SkillAnnotatedLine{192}{---\ }
\SkillAnnotatedLine{193}{\mbox{}\ }
\SkillAnnotatedLine{194}{\#\#\ Boundary\ Rules\ }
\SkillAnnotatedLine{195}{\mbox{}\ }
\SkillAnnotatedLine{196}{1.\ **\SkillGeneral{No\ answer\ key\ exists\ in\ this\ skill},\ by\ design.**\ No\ calibrated\ parameter\ set\ for\ Lake\ Mendota\ }
\SkillAnnotatedLine{197}{\ \ \ is\ supplied\ anywhere\ in\ this\ pack.\ The\ literature\ gives\ *search\ bounds\ and\ starting\ points*,\ }
\SkillAnnotatedContinuation{not\ }
\SkillAnnotatedLine{198}{\ \ \ optima.\ The\ only\ published,\ *used*\ Mendota\ value\ citable\ here\ is\ \SkillLocal{\textasciigrave{}Kw\ =\ 0.69\ m\ensuremath{^{-}}\ensuremath{^{1}}\textasciigrave{}}\ (Bruce\ et\ al.\ }
\SkillAnnotatedLine{199}{\ \ \ 2018).\ If\ you\ find\ yourself\ pasting\ parameters\ from\ memory,\ stop\ ---\ you\ are\ inventing\ them.\ }
\SkillAnnotatedLine{200}{2.\ **\SkillGeneral{Never\ tune\ the\ metric\ to\ hit\ the\ target}.**\ The\ RMSE\ convention\ (H5)\ is\ fixed.\ Different\ }
\SkillAnnotatedLine{201}{\ \ \ aggregation\ choices\ can\ materially\ change\ RMSE\ on\ the\ same\ simulation,\ so\ the\ requested\ }
\SkillAnnotatedLine{202}{\ \ \ convention\ is\ a\ \SkillLocal{task-local\ binding}\ rather\ than\ a\ tunable\ choice.\ Excluding\ the\ ice\ season\ or\ }
\SkillAnnotatedLine{203}{\ \ \ surface\ layer\ is\ not\ a\ result;\ no\ such\ exclusion\ exists\ in\ glmtools\ or\ LakeEnsemblR.\ }
\SkillAnnotatedLine{204}{3.\ **\SkillGeneral{Calibrate\ at\ most\ 4--6\ parameters}.**\ Across\ 73\ lakes,\ only\ 2--3\ of\ 6\ were\ sensitive\ }
\SkillAnnotatedLine{205}{\ \ \ (HESS\ 2025).\ Ladwig\ used\ 4\ for\ Mendota.\ More\ free\ parameters\ buys\ equifinality,\ not\ accuracy.\ }
\SkillAnnotatedLine{206}{4.\ **A\ \SkillGeneral{parameter\ pinned\ to\ its\ bound\ is\ a\ diagnostic},\ not\ a\ victory.**\ It\ usually\ means\ the\ }
\SkillAnnotatedContinuation{forcing\ }
\SkillAnnotatedLine{207}{\ \ \ data\ is\ wrong\ and\ the\ parameter\ is\ absorbing\ the\ error.\ Investigate\ rather\ than\ widening\ the\ }
\SkillAnnotatedContinuation{bound.\ }
\SkillAnnotatedLine{208}{5.\ **RMSE\ \textless{}\ 1.0\ \textdegree{}C\ whole-column\ should\ be\ disbelieved.**\ No\ published\ GLM\ study\ on\ a\ \SkillLocal{Mendota-like}\ }
\SkillAnnotatedLine{209}{\ \ \ lake\ reports\ one.\ Suspect\ a\ broken\ obs\ensuremath{\leftrightarrow}sim\ match\ (unit\ error,\ wrong\ depth\ convention,\ }
\SkillAnnotatedContinuation{near-empty\ }
\SkillAnnotatedLine{210}{\ \ \ match\ set)\ before\ celebrating.\ Check\ N.\ }
\SkillAnnotatedLine{211}{6.\ **Report\ by\ depth\ band,\ always.**\ \SkillLocal{Mendota's\ published\ surface/bottom\ gap\ is\ 1.30\ vs\ 2.43\ \textdegree{}C}\ }
\SkillAnnotatedLine{212}{\ \ \ (NSE\ 0.97\ vs\ 0.20).\ An\ aggregate\ that\ hides\ a\ broken\ hypolimnion\ is\ not\ a\ passing\ result.\ }
\SkillAnnotatedLine{213}{7.\ **Scope:\ \SkillGeneral{water\ temperature\ only,\ GLM\ only}.**\ Not\ AED2\ water\ quality,\ not\ other\ lake\ models.\ The\ }
\SkillAnnotatedLine{214}{\ \ \ \textasciigrave{}coef\_mix\_hyp\textasciigrave{}\ bounds\ here\ assume\ the\ Weinstock\ \textasciigrave{}deep\_mixing\textasciigrave{}\ option\ ---\ **its\ units\ change\ with\ }
\SkillAnnotatedLine{215}{\ \ \ that\ setting**;\ check\ it\ before\ applying\ any\ bound.\ }
\SkillAnnotatedLine{216}{8.\ **\SkillGeneral{Evidence\ boundary}.**\ Everything\ here\ traces\ to\ the\ GLM\ C\ source,\ the\ glmtools/glm-py\ reader\ }
\SkillAnnotatedLine{217}{\ \ \ source,\ or\ peer-reviewed\ literature\ ---\ all\ URLs\ in\ \textasciigrave{}references/sources.md\textasciigrave{}.\ Claims\ that\ are\ }
\SkillAnnotatedLine{218}{\ \ \ *inferred*\ rather\ than\ read\ are\ labelled\ as\ such\ in\ the\ cards;\ keep\ the\ label.\ Information\ }
\SkillAnnotatedLine{219}{\ \ \ current\ as\ of\ the\ sources\ cited\ (GLM\ 3.x,\ literature\ through\ 2025).\ }
\begin{SkillAnnotationBox}{skillLocalFg}{Local evidence remains scoped}Mendota measurements, literature ranges, commands, and parameter examples are retained as scoped evidence. The reusable claim is the validation or diagnostic principle, so no over-specific marker is needed.\end{SkillAnnotationBox}
\SkillAnnotatedLine{220}{\mbox{}\ }
\SkillAnnotatedLine{221}{---\ }
\SkillAnnotatedLine{222}{\mbox{}\ }
\SkillAnnotatedLine{223}{\#\#\ Examples\ }
\SkillAnnotatedLine{224}{\mbox{}\ }
\SkillAnnotatedLine{225}{\textasciigrave{}examples/worked\_session.md\textasciigrave{}\ walks\ four\ scenarios\ end\ to\ end.\ Read\ it\ if\ you\ want\ to\ see\ the\ loop\ }
\SkillAnnotatedLine{226}{before\ running\ it\ ---\ especially\ the\ two\ where\ the\ correct\ move\ is\ to\ **distrust\ the\ result**:\ }
\SkillAnnotatedLine{227}{\mbox{}\ }
\SkillAnnotatedLine{228}{-\ **\SkillLocal{Bootstrap}**\ ---\ cold\ start:\ \textasciigrave{}ldd\textasciigrave{}\ \ensuremath{\rightarrow}\ nml\ dump\ \ensuremath{\rightarrow}\ baseline\ run\ \ensuremath{\rightarrow}\ \textasciigrave{}ncdump\ -h\textasciigrave{}\ \ensuremath{\rightarrow}\ baseline\ RMSE,\ in\ }
\SkillAnnotatedContinuation{that\ }
\SkillAnnotatedLine{229}{\ \ order,\ touching\ no\ parameters\ until\ the\ last\ step.\ }
\SkillAnnotatedLine{230}{-\ **\SkillLocal{Diagnose}**\ ---\ a\ residual\ concentrated\ in\ the\ hypolimnion\ (+2.97\ \textdegree{}C,\ worst\ in\ summer)\ points\ at\ }
\SkillAnnotatedLine{231}{\ \ exactly\ one\ knob\ (\textasciigrave{}sed\_temp\_mean\textasciigrave{}),\ and\ explicitly\ *not*\ at\ \textasciigrave{}sw\_factor\textasciigrave{},\ which\ would\ trade\ away\ }
\SkillAnnotatedContinuation{a\ }
\SkillAnnotatedLine{232}{\ \ surface\ fit\ that\ is\ already\ good.\ }
\SkillAnnotatedLine{233}{-\ **The\ \SkillLocal{suspiciously\ good\ result}**\ ---\ RMSE\ comes\ back\ at\ 0.62\ \textdegree{}C.\ It\ is\ not\ a\ win:\ \textasciigrave{}N\ =\ 37\textasciigrave{}\ of\ }
\SkillAnnotatedContinuation{\textasciitilde{}2800\ }
\SkillAnnotatedLine{234}{\ \ observations,\ because\ the\ obs\ensuremath{\leftrightarrow}sim\ time\ match\ silently\ collapsed.\ A\ number\ that\ beats\ the\ target\ }
\SkillAnnotatedLine{235}{\ \ for\ the\ wrong\ reason\ is\ worse\ than\ one\ that\ misses\ it.\ }
\SkillAnnotatedLine{236}{-\ **\SkillLocal{Delivery}**\ ---\ the\ trap\ people\ actually\ lose\ on:\ iterating\ in\ a\ scratch\ dir,\ then\ shipping\ an\ }
\SkillAnnotatedContinuation{nml\ }
\SkillAnnotatedLine{237}{\ \ that\ reproduces\ something\ other\ than\ the\ delivered\ \textasciigrave{}output.nc\textasciigrave{}.\ Check\ 4\ of\ \textasciigrave{}verify\_delivery.py\textasciigrave{}\ }
\SkillAnnotatedLine{238}{\ \ is\ what\ catches\ it.\ }
\begin{SkillAnnotationBox}{skillLocalFg}{Worked local cases}Concrete sessions illustrate the generalized protocol without being presented as universal solutions.\end{SkillAnnotationBox}
\SkillAnnotatedLine{239}{\mbox{}\ }
\SkillAnnotatedLine{240}{\#\#\ References\ }
\SkillAnnotatedLine{241}{\mbox{}\ }
\SkillAnnotatedLine{242}{|\ File\ |\ What\ it\ holds\ |\ Read\ it\ when\ |\ }
\SkillAnnotatedLine{243}{|------|--------------|--------------|\ }
\SkillAnnotatedLine{244}{|\ \textasciigrave{}references/sop\_models.md\textasciigrave{}\ |\ The\ 9\ operation\ cards\ (H1--H9),\ 8\ fields\ each,\ with\ evidence\ URLs\ |\ }
\SkillAnnotatedContinuation{Always\ ---\ Step\ 0\ of\ the\ protocol\ |\ }
\SkillAnnotatedLine{245}{|\ \textasciigrave{}references/nml\_reference.md\textasciigrave{}\ |\ Every\ nml\ block\ +\ the\ Mendota\ starting\ config;\ parameter\ tiers,\ }
\SkillAnnotatedContinuation{bounds,\ units;\ CSV\ forcing\ contracts;\ the\ literature\ inconsistencies\ |\ Before\ changing\ any\ }
\SkillAnnotatedContinuation{parameter\ |\ }
\SkillAnnotatedLine{246}{|\ \textasciigrave{}references/troubleshooting.md\textasciigrave{}\ |\ 22\ failure\ modes:\ symptom\ \ensuremath{\rightarrow}\ cause\ \ensuremath{\rightarrow}\ confirm\ \ensuremath{\rightarrow}\ fix.\ "First\ 60\ }
\SkillAnnotatedContinuation{seconds"\ checklist\ |\ GLM\ misbehaves,\ or\ produces\ nothing\ |\ }
\SkillAnnotatedLine{247}{|\ \textasciigrave{}references/sources.md\textasciigrave{}\ |\ Every\ URL\ used,\ what\ it\ establishes,\ confidence.\ How\ to\ re-verify\ this\ }
\SkillAnnotatedContinuation{skill\ yourself\ |\ You\ doubt\ a\ claim\ here\ (you\ should)\ |\ }
\SkillAnnotatedLine{248}{|\ \textasciigrave{}references/research\_notes.md\textasciigrave{}\ |\ The\ evidence\ summary,\ the\ contradictions\ found,\ what\ stays\ }
\SkillAnnotatedContinuation{uncertain\ |\ You\ want\ the\ reasoning,\ not\ the\ rule\ |\ }
\SkillAnnotatedLine{249}{|\ \textasciigrave{}references/R01\textasciigrave{}--\textasciigrave{}R05\textasciigrave{}\ |\ Raw\ research\ reports\ (GLM\ source\ reads,\ literature,\ numerical\ }
\SkillAnnotatedContinuation{verification)\ |\ Full\ provenance\ for\ any\ single\ claim\ |\ }
\SkillAnnotatedLine{250}{|\ \textasciigrave{}scripts/\textasciigrave{}\ |\ \textasciigrave{}nml\_edit\textasciigrave{}\ ·\ \textasciigrave{}run\_glm\textasciigrave{}\ ·\ \textasciigrave{}extract\_temp\textasciigrave{}\ ·\ \textasciigrave{}eval\_rmse\textasciigrave{}\ ·\ \textasciigrave{}calibrate\textasciigrave{}\ ·\ }
\SkillAnnotatedContinuation{\textasciigrave{}verify\_delivery\textasciigrave{}\ |\ Throughout\ ---\ they\ encode\ the\ traps\ |\ }
\SkillAnnotatedLine{251}{\mbox{}\ }
\SkillAnnotatedLine{252}{\SkillGeneral{Three\ checks\ a\ skeptical\ solver\ should\ personally\ re-run}:\ \textasciigrave{}ncdump\ -h\ output.nc\textasciigrave{}\ (confirm\ the\ dims\ }
\SkillAnnotatedLine{253}{and\ vars),\ glmtools'\ \textasciigrave{}resample\_depth()\textasciigrave{}\ (confirm\ the\ midpoint\ convention),\ GLM's\ \textasciigrave{}glm\_main.c\textasciigrave{}\ }
\SkillAnnotatedLine{254}{(confirm\ the\ unconditional\ \textasciigrave{}exit(0)\textasciigrave{}).\ Do\ not\ take\ this\ skill's\ word\ for\ any\ of\ them.\ }
\end{SkillAnnotatedFileBox}

\endgroup

\else

\begin{abstract}
Agent skills are reusable procedural artifacts that extend language agents with specialized workflows, tool conventions, and domain behaviors at inference time.
However, creating reliable skills still depends largely on human authorship, model priors, or execution traces. 
These sources are often unavailable for unfamiliar tasks, suggesting the need to create skills from open-world materials.
In this paper, we study \emph{open-world skill creation}: given an underspecified skill brief and a source-access specification, a creator must discover behavior-relevant requirements omitted by the brief and determine how broadly each source-derived procedure is justified.
We propose \sysname{}, an admission-centered framework for source-grounded skill creation. 
\sysname{} identifies implicit requirements through contrastive evidence, admits candidate procedures based on evidence-supported scope, and compiles the admitted content into a grammar-guided skill package.
Extensive experiments across 87 \textsc{SkillsBench} v1.1 tasks demonstrate that our \sysname{} improves pass rate over no-skill execution by 19.9pp and the strongest automated baseline by 8.6pp and is comparable to human-curated skills.
\end{abstract}


\section{Introduction}
\label{sec:intro}

    Agent skills are reusable procedural artifacts that enable language agents to dynamically load and execute specialized workflows and domain-specific behaviors at inference time.
    In current agent ecosystems, a skill is commonly packaged as a filesystem-based
    artifact centered on a \texttt{SKILL.md} file with metadata (\eg skill descriptions) and instructions and may bundle scripts, references, assets, or other resources that an agent can load on demand~\citep{anthropic-agent-skills,openai-codex-skills}.

    Equipping agents with skills enables them to perform a wide range of practical tasks beyond model priors, including code-generation workflows, data-analysis pipelines and document-processing routines ~\citep{survey-comprehensive,skillsbench,skills-in-the-wild}.
    Despite their effectiveness and flexibility as deployment-time extensions, existing skills are typically created by experts, generated from model priors, or distilled from reasoning traces \citep{expel, trace2skill, skillrevise, skillevolver, autoskill}.

    
    However, these routes rely on a distinct procedural knowledge source that is not always accessible.
    Expert-crafted skills demand extensive manual labor, model priors limit self-generated skills, and trace-based skills require archived execution traces.
    These assumptions break down most severely for unfamiliar tasks or capabilities, precisely when new custom skills are urgently required.
    In such cases, useful procedural knowledge may already be in open-world materials (including documentation, repositories and issue reports) yet remains largely under-exploited for reusable agent skill specifications.


    \begin{figure}[!t]
        \vspace{-1em}
        \centering
        \setlength{\tabcolsep}{0pt}
    
        \begin{tabular}{cc}
    
            \multicolumn{2}{c}{
                \includegraphics[width=0.97\columnwidth]{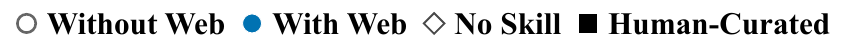}
            }\\[-0.5em]
    
            \includegraphics[width=0.52\columnwidth]{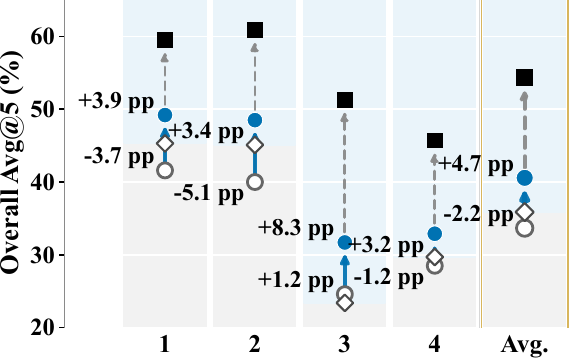}
            &
            \includegraphics[width=0.45\columnwidth]{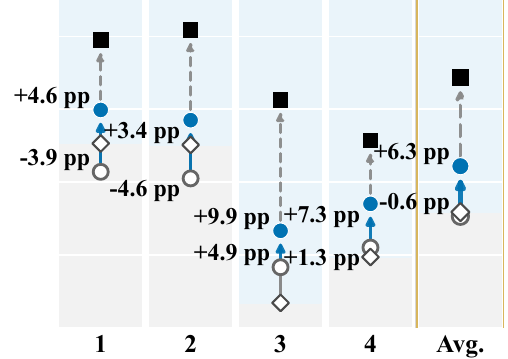}
            \\[-0.5em]
    
            {\small {(a) Anthropic-Skill-Creator}}
            &
            {\small {(b) OpenAI-Skill-Creator}}
            \\
            [-0.1em]
    
            \multicolumn{2}{c}{%
                \footnotesize
                \begin{tabular}{@{}r@{\,}l@{\qquad}r@{\,}l@{}}
                    1: & Claude Code + Opus-4.8 
                    & 2: & Codex + GPT-5.5  \\
                    3: & Claude Code + DeepSeek-V4
                    & 4: &  Codex + DeepSeek-V4
                \end{tabular}%
            }
        \end{tabular}
        \vspace{-0.8em}
        \caption{\textbf{\small Pilot Study}: {\small Open-world source access improves skill creation but does not close the gap to human-curated skills}.}
        \vspace{-1.8em}
        \label{fig:web-access}
    \end{figure}

    \begin{figure*}[!t]
    \centering
        \includegraphics[width= \linewidth]{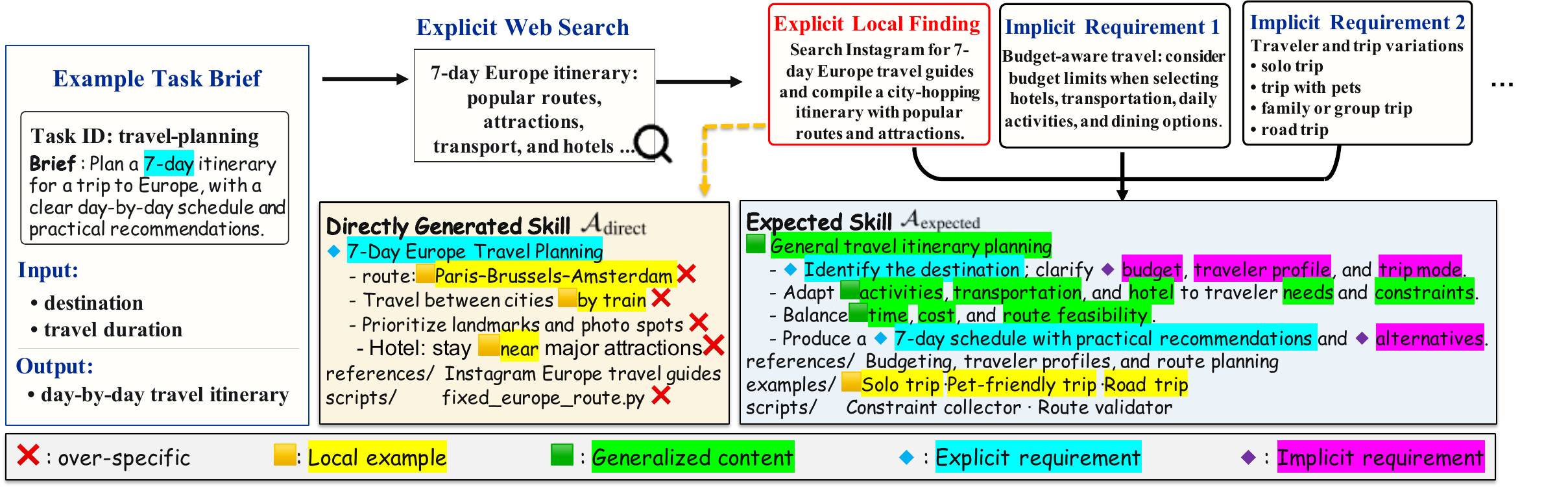}
        \vspace{-1.0em}

        \caption{\small \textbf{Example illustrating why open-world skill creation is non-trivial}: \ding{172} task briefs underspecify implicit requirements, and \ding{173} directly adopting open-world findings without scope justification may lead to over-specific practices being mistaken for reusable instructions.}
        \label{fig:example}
        \vspace{-1.5em}
    \end{figure*}

    \fakeparagraph{Pilot Study} We examine two questions on \textsc{SkillsBench}     v1.1~\citep{skillsbench}: \textit{whether access to open-world sources improves automatic skill creation and whether such access alone closes the gap to human-curated skills?}
    We use two representative official skill creators released by Anthropic and OpenAI~\citep{anthropic-skill-creator,openai-codex-skills}.
    Each creator builds skills for the same tasks with and without web access, and the resulting skills are executed by the same downstream agents (as in \figref{fig:web-access}).
    Without web access, the generated skills perform 1.4 percentage points below no-skill execution on average across four agent--model configurations.
    With web access, every configuration improves by 6.9 points on average over its without-web counterpart and by 5.5 points over no-skill execution.
    However, the stronger web-grounded creator remains about 14 points behind human-curated skills.
    These results show that open-world sources are beneficial but insufficient for reliable skill creation, motivating the central question of this work: {how can reliable agent skills be created from open-world sources?}
    Even with web grounding, the better web-access creator remains about 14 points behind human-curated skills, motivating our diagnosis: \textit{open-world sources are informative but not skill-ready as they contain implicit decisions, local examples, and context-dependent practices rather than validated reusable procedures.}
    Specifically, converting open-world sources into reusable skills is challenging for two reasons (as illustrated in \figref{fig:example}).

    \begin{itemize}
        \item \textbf{Task briefs under-specify operational requirements.} 
        A task brief typically states the immediate objective but leaves implicit
        the requirements, failure modes, and operational boundaries needed for
        a reusable skill.
        Using the brief directly as a retrieval query preserves these blind spots.
        Individual sources are also organized around their own subjects rather than around the complete target capability.
        \textit{Thus, reliable skill creation must discover missing requirements beyond the brief before acquiring evidence.}
        
        \item \textbf{Open-world findings do not justify their reusable scope.}
        A source-specific finding often mixes a reusable practice with local
        details, such as hardcoded parameters, preferred tools, fixed inputs, or
        environment-specific assumptions.
        The single occurrence is insufficient to justify promoting the entire finding into a persistent instruction.
        \textit{The skill creation therefore needs to determine scope across cases, promoting consistent practices into general instructions, retaining
        context-bound ones as scoped examples, and excluding candidates whose support is weak or conflicting.}
    \end{itemize}

    We formulate the open-world skill creation as a source-grounded procedure-admission problem, \ie given an underspecified skill brief and a set of heterogeneous sources, the skill creator must first recover absent requirements from the brief and then decide whether each candidate is licensed as a reusable instruction, remains a scoped example, or should be excluded.
    To address these challenges, we propose \textsc{SkillAlchemy}, a framework that transforms an underspecified description of a target task or capability into an agent-usable skill. It discovers implicit requirements and admits instructions only when open-world evidence justifies their scope.
    \textsc{SkillAlchemy} is not merely a retrieval pipeline, but an admission-centered framework operates in three stages.
    \textit{(i) Implicit Requirement Discovery} (\S\ref{sec:lens}) lifts the brief to its underlying capability, identifies omitted operational dimensions, and converts them into focused research questions. 
    \textit{(ii) Grounded Procedure Admission} (\S\ref{sec:admission}) aggregates relevant findings and admits a candidate as a reusable instruction only when the evidence supports both the action and its scope.
    \textit{(iii) Skill Package Compilation} (\S\ref{sec:compile}) organizes admitted procedures and scoped examples into an installable skill package using the skill grammar and task-relevant exemplars.

    \noindent Our main contributions are summarized as follows.
    \begin{itemize}
        \item We formulate open-world skill creation as a source-grounded procedure-admission problem, identifying two key challenges in converting an underspecified brief and heterogeneous sources into a reusable skill: implicit requirement discovery and procedure-scope justification.
        \item We propose \sysname{}, a skill-creation framework that turns implicit requirements in briefs into focused targets for open-world knowledge acquisition and determines whether source findings warrant general instructions, local examples, or exclusions of an installable skill package.

        \item We evaluate \sysname{} over 87 tasks from \textsc{SkillsBench} across four agent--model configurations. 
        Our framework improves {pass rate} by {19.9 percentage points} over no-skill execution and by {8.6 percentage points} over the strongest automatic skill-creation baseline, comparable to the human-curated skills.
        Ablation studies further examine requirement coverage and unsupported procedure admission.

    \end{itemize}

\section{Related Work}
\label{sec:related}

\paragraph{Skills for Agents.}
Agent skills are typically treated as reusable procedural artifacts rather than ordinary prompts or atomic tool calls~\citep{sok-agentic-skills,survey-comprehensive, skillsh}.
SkillAct~\citep{skillact} shows that adding reusable skill abstractions to existing prompting methods (\eg ReAct~\citep{react}) improves agent performance on interactive tasks such as ALFWorld~\citep{alfworld}.
Skills-in-the-Wild further examines whether agents can effectively leverage skills in realistic settings, where useful skills need to be retrieved from a large and noisy collection rather than being manually curated~\citep{skills-in-the-wild}.
    Together, these studies establish an artifact-centric view of the skills lifecycle,  in which skills can be constructed, represented, retrieved, invoked, and evaluated.
    Within such a lifecycle, skill construction determines what procedural knowledge is conceptualized into the repository in the first place, thereby directly shaping the  utility of all downstream skill use.
Aligned with this line of research, our \sysname{} studies how to produce the skill artifacts from open-world source materials.

\paragraph{Skill Creation.}
Skill creation methods can be categorized by the source of candidate skill content.
\textit{(i) Human-authored skills}.
Expert-written or benchmark-provided skills~\cite{skillsbench} encode human procedural knowledge and serve as strong reference artifacts.
\textit{(ii) Interaction traces} based skill creation.
Voyager~\citep{voyager} builds an executable skill library from open-ended embodied interaction.
ExpeL~\citep{expel} learns reusable lessons from task experience without updating model weights.
SkillGen~\citep{skillgen} synthesizes auditable skills from successful and failed traces and checks their net intervention effect.
CoEvoSkills~\citep{coevoskills} iteratively evolves multi-file skill packages using surrogate verification and execution feedback.
Together, these methods show how execution records can be transformed into reusable procedural knowledge, where the main evidence comes from observed attempts, failures, successes, or intervention effects.
\textit{(iii) Broad-source and scaffolded skill creation}.
Official skill creation workflows provide general-purpose scaffolds for authoring and improving skills from available context~\citep{anthropic-skill-creator,openai-skill-creator}.
OpenSkill~\citep{openskill} retrieves documentation, repositories, and web resources to build transferable skills and verification anchors.
SkillGenBench~\citep{skillgenbench} benchmarks skill generation from repository- and document-grounded sources.

\sysname{} is closest to the broad-source creation line.
Rather than treating retrieved or provided source content as skill content directly, it performs knowledge acquisition and evidence aggregation before skill creation, which focus complements human-authored and trace-based creation.

\begin{figure*}[t]
\centering
\includegraphics[width=\textwidth]{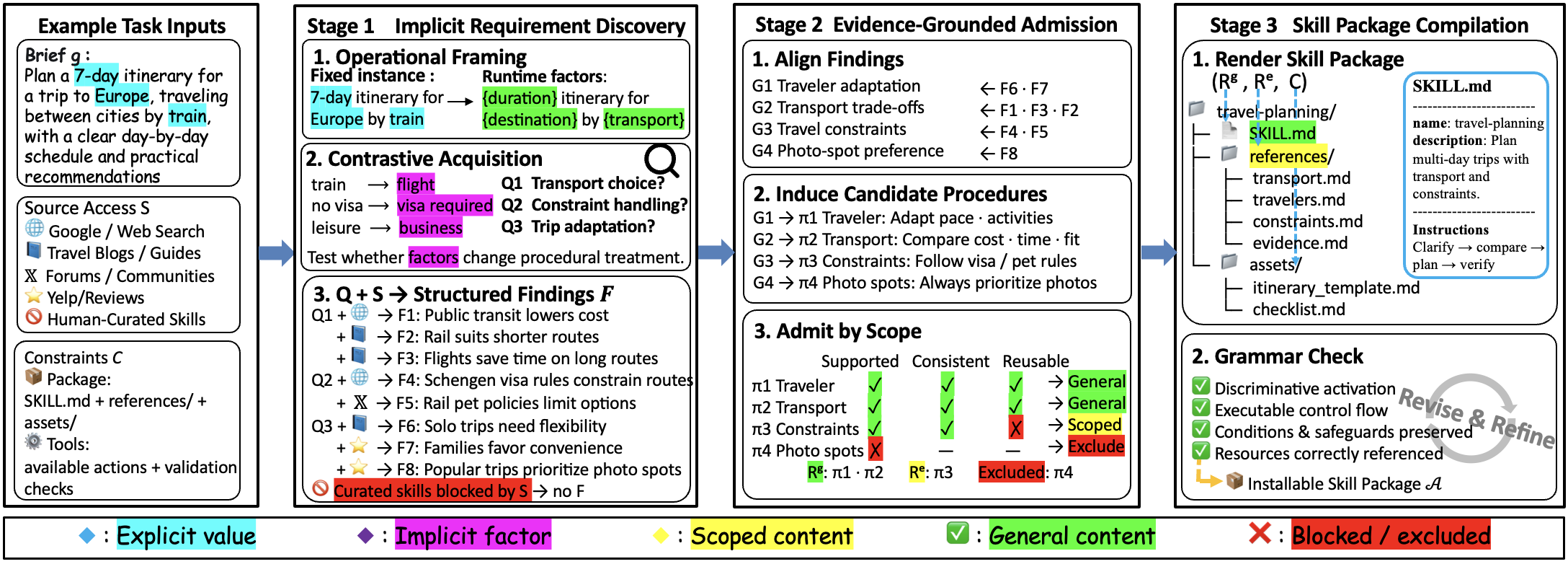}
\caption{\small \textbf{Overview of \sysname{}}: $\Phi_1$ converts an underspecified brief into focused questions and structured findings,
$\Phi_2$ induces candidate procedures and admits them via the evidence-supported scope, and $\Phi_3$ compiles the admitted content into an agent skill package.
}
\label{fig:example_overview}
\vspace{-1.5em}
\end{figure*}




\section{Method}
\label{sec:method}

\subsection{Problem Formulation}
\label{sec:preliminaries}

We study \emph{open-world skill creation}, where evidence needed to construct a reusable skill is not assumed to be organized as a task-complete corpus but must be identified and acquired from heterogeneous sources  (\eg repositories, documents or agent experience) under a specified access policy.

The input is a tuple $(g,\mathcal{S},\mathcal{C})$:
\textit{(i)} an underspecified skill brief $g$, namely a short
natural-language description of the task or capability the skill
should support,
\textit{(ii)} a source-access specification $\mathcal{S}$ defining permitted source types, retrieval channels, and exclusions (\eg documentation, repositories, or existing skills), and
\textit{(iii)} execution and packaging constraints $\mathcal{C}$
(\eg available tools and required artifact structure).
A skill creator $\Phi$ maps these inputs to an installable skill package,
\begin{equation}
\label{eq:construction}
    \mathcal{A}:=\langle \texttt{SKILL.md},\mathcal{X}\rangle = \Phi(g,\mathcal{S},\mathcal{C}),
\end{equation}
In this work, the skill package $\mathcal{A}$ follows the filesystem-based skill
convention centered on \texttt{SKILL.md} and optionally bundlings $\mathcal{X}$ (\eg references, examples and scripts).


\fakeparagraph{Source-Grounded Procedure Admission}
Unlike a document summarization, which preserves what the sources state, an agent skill must specify \textit{reusable procedures}: when an instruction applies, what the agent should do and produce, and the conditions under which it fails or falls outside scope.
The aim of \textit{open-world skill creation} is therefore not to compress or summarize a fixed source collection but to acquire evidence materials under $\mathcal{S}$ and decide whether each candidate procedure should be admitted as a general instruction, retained as a scoped example or notes, or just be excluded from the skill.

\subsection{Framework Overview}
\label{sec:method-overview}

\fakeparagraph{Design Rationale}
Existing official general-purpose skill creation workflows typically author a skill directly from the task brief and available context~\citep{anthropic-skill-creator,openai-skill-creator}.
When the brief is underspecified, this process entangles three distinct
decisions, \ie what omitted by the brief should be investigated, whether a source-derived candidate is supported beyond its local context, and how the admitted content should be expressed in the final artifact.
\sysname{} separates these decisions explicit and resolves them sequentially.
It first discovers implicit requirements and acquires evidence for them,
then determines which candidate procedures are supported at which scope, and finally compiles the admitted content as an installable skill package.
This separation is the central design choice of our framework.


\paragraph{Three-Stage Creation Process.}
Figure~\ref{fig:example_overview} illustrates the following three-stage workflow through a running example.
\begin{itemize}
\item 
\emph{Stage 1: implicit requirement discovery} identifies behavior-relevant distinctions omitted in briefs, turns them into focused research questions, and acquires structured findings.

\item 
\emph{Stage 2: evidence-grounded procedure admission} aggregates findings that address the same situations, makes conflicts and missing support explicit, and distills supported content into general instructions or scoped examples.

\item 
\emph{Stage 3: skill package compilation} renders admitted instructions and scoped examples as an installable package under a corpus-derived skill grammar.
\end{itemize}

Writing $\Phi_1,\Phi_2,\Phi_3$ for the three stages, they jointly compose
the whole open-world skill creation pipeline of \equref{eq:construction},
\begin{equation}
\label{eq:framework}
    (g,\mathcal{S})
    \xrightarrow{\Phi_1}
    (Q,F)
    \xrightarrow{\Phi_2}
    (R^g,R^e)
    \xrightarrow[\mathcal{C}]{\Phi_3}
    \mathcal{A}.
\end{equation}
where $Q$ is the set of research questions, $F$ the structured findings, $R^{g}$ the admitted general instructions, and $R^{e}$ the retained scoped examples.
Candidates admitted to neither $R^{g}$ nor $R^{e}$ are excluded.
These intermediate outputs are preserved as explicit provenance records, whereas $R^{g}$ and $R^{e}$ are compiled into the skill content under $\mathcal{C}$.
The following three subsections describe these creation stages subsequently.

\subsection{Implicit Requirement Discovery}
\label{sec:lens}

\fakeparagraph{Operational Framing}
A task brief may describe a requested instance, whereas a reusable skill must operate over a family of instances.
Given a brief $g$ and source-access specification $\mathcal{S}$, $\Phi_1$ maps $g$ to an operational frame $L_g=\operatorname{Frame}(g)$.
The frame treats brief-specific values as candidate operational factors and exposes the procedural decisions left unspecified for producing the requested output, handling failures, and verifying the result.
For example, a brief requesting ``a 7-day family trip to Europe'' contains instance-specific values such as \textsf{7-day}, \textsf{family trip}, and \textsf{Europe}.
The operational frame instead represents them as candidate factors such as \textsf{duration}, \textsf{traveler type}, and \textsf{destination}, rather than hard-coding them into a reusable procedure.
The frame therefore proposes operational factors for investigation rather than treating them as requirements and a factor becomes an implicit operational factor only when acquired evidence shows that varying it changes procedural behavior.

\fakeparagraph{Contrastive Evidence Acquisition}
Directly searching with the original brief tends to retrieve repeated evidence about the same local instance, whereas open-ended requirement expansion may introduce many conditions unrelated to skill behavior.
Thus, \sysname{} constructs paired acquisition targets $h=\langle d,x,x'\rangle$, where $x$ and $x'$ are matched task contexts that differ along one candidate operational factor $d$.
Here, a source-stated applicability boundary along $d$ counts as evidence that the corresponding component has different applicability across the matched contexts while missing evidence for either context alone does not establish such a difference.
A target may vary a declared factor to test whether a procedure transfers beyond the seed instance, or an omitted factor to test whether the brief lacks a behavior-changing condition.
Specifically, \textit{(i) a substitution probe} varies a subject, method, or tool within the same capability family. 
For example, a substitution probe may replace a family traveler with a solo traveler while keeping the destination and duration fixed.
\textit{(ii) A boundary probe} introduces an omitted precondition, failure, or operating constraint.
\textit{(iii) A neighbor probe} compares the target with a sibling capability under the same output interface.
Each target is converted into a focused question asking whether the matched contexts require different treatment in any component $k\in\mathcal{K}=\{c,a,r,v\}$, where $c$, $a$, $r$, and $v$ denote conditions, action, recovery, and verification, respectively.
These components characterizes procedural behavior rather than just prescribing the layout of the \texttt{\small SKILL.md}.
For findings $F_h$ acquired for targets $h$, let
\begin{equation}
\label{eq:behavior-gap}
    K_F(h)=
    \left\{
    k\in\mathcal{K}
    \;\middle|\;
    F_h \models
    \operatorname{Treat}_{k}(x)
    \neq
    \operatorname{Treat}_{k}(x')
    \right\}.
\end{equation}
Only when $K_F(h)\neq\emptyset$ does \sysname{} record an implicit requirement to condition procedural behavior on $d$, together with the affected components and any evidence-stated boundary.
Evidence supporting the same treatment across non-equivalent contexts is retained as cross-context invariance evidence.
Conflicting findings remain explicit, whereas insufficient evidence leaves the contrast unresolved.
Finally, $\Phi_1$ returns questions $Q$ and structured findings $F$, which remain source-grounded observations rather than executable procedures.
Next, Stage~2 $\Phi_2$ determines whether these findings are justified to be general instructions, scoped examples, or exclusions.

\subsection{Evidence-Grounded Procedure Admission}
\label{sec:evidence}
\label{sec:admission}

\fakeparagraph{Decision-Aligned Induction}
Since the findings $F$ remain tied to local source contexts, $\Phi_2$ first groups findings by the procedural decision they inform, rather than by their source topic or document,
\begin{equation}
\label{eq:procedure-induction}
\{G_1,\ldots,G_m\}\leftarrow\operatorname{Align}(F),
\qquad
\pi_j\leftarrow\operatorname{Induce}(G_j).
\end{equation}
Each Stage-1 finding retains the acquisition target that produced it, its operating context, and the treatment reported by the source.
Using this information, $\operatorname{Align}(\cdot)$ places findings about the same decision point, such as route selection, accommodation choice, or failure handling, into one group $G_j$, while preserving their recorded conditions.
The induced candidate is represented as $\pi_j=\langle c_j,a_j,r_j,v_j,P_j\rangle$, where $c_j$ records its
applicability conditions.
$a_j$, $r_j$, and $v_j$ denote its action, recovery, and verification components and $P_j$ maps each populated component to the findings that support it.
Induction includes only content directly supported by findings in $G_j$.
It may canonicalize synonymous source terms, but does not generalize
named entities or fixed choices beyond their observed contexts unless
cross-context evidence supports doing so.
Candidates without supporting evidence are left unspecified.
When different treatments are supported under distinct recorded conditions, the candidate retains them as separate conditional cases, and incompatible treatments under matched conditions remain unresolved conflicts.

\fakeparagraph{Scope-Aware Admission}
For each candidate procedure $\pi_j$, \sysname{} constructs an admission record as follows,
\[
\mathcal{D}(\pi_j)
=
\langle F_j^{+},F_j^{-},\sigma_j\rangle,
\]
where $\sigma_j$ is the widest applicability scope justified by the current evidence, $F_j^{+}$ contains findings supporting the components specified in $\pi_j$ within $\sigma_j$, and $F_j^{-}$ contains findings prescribing incompatible treatment under overlapping operating conditions.
$c_j$ is part of the procedure, whereas $\sigma_j$ records how broadly the available evidence licenses that procedure.
A candidate is \emph{supported} if every component specified in $\pi_j$ is backed by evidence under the conditions for which it is claimed.
It is \emph{consistent} if no unresolved conflicting finding applies under matched conditions within $\sigma_j$.
When the evidence supports the candidate only under a narrower scope,
\sysname{} restricts $\sigma_j$ to that scope before admission.
A supported and consistent candidate is \emph{reusable} only when the procedure is justified beyond a single source-local case: either an eligible source under $\mathcal{S}$ explicitly states broader applicability, or compatible evidence supports the same treatment across non-equivalent contexts identified by $\Phi_1$.
Let $S_j$, $C_j$, and $U_j$ denote whether $\pi_j$ is supported, consistent, and reusable, respectively.
The admission decision is,
\begin{equation}
\label{eq:procedure-admission}
\operatorname{Admit}(\pi_j)=
\begin{cases}
\textsc{General}, & S_j\land C_j\land U_j,\\
\textsc{Scoped},  & S_j\land C_j\land\neg U_j,\\
\textsc{Exclude}, & \text{otherwise}.
\end{cases}
\end{equation}
The general candidates form $R^{g}$ and are compiled as reusable
instructions.
Scoped candidates form $R^{e}$ and are retained as context-bound
examples.
Excluded candidates remain in the audit record and are not passed to
$\Phi_3$ as the skill content.

\subsection{Skill Package Compilation}
\label{sec:compile}
Package compilation does not create new procedures but maps the admitted procedures into an executable skill artifact.
Given admitted instructions $R^{g}$, scoped examples $R^{e}$, and constraints $\mathcal{C}$, $\Phi_3$ renders them, without changing admitted scope, as a standard skill package containing \texttt{\small SKILL.md} and optional bundled resources $\mathcal{X}$,
\begin{equation}
\label{eq:compile}
\mathcal{A} =
\operatorname{Render}(R^{g},R^{e};\mathcal{C}) = \langle \texttt{SKILL.md},\mathcal{X}\rangle.
\end{equation}
Specifically, \sysname{} writes the skill name and a description of what the skill does and when it should be used to the {\small YAML} frontmatter of \texttt{\small SKILL.md}.
It then renders the admitted procedures, together with their applicability conditions and safeguards, as executable instructions in its body.
Supporting content not needed in the initially loaded \texttt{\small SKILL.md} context is externalized as optional bundled resources $\mathcal{X}$ and referenced from \texttt{\small SKILL.md}, \eg detail notes and scoped examples in \texttt{\small references/*}, executable routines in \texttt{\small scripts/*}, and templates or static resources in \texttt{\small assets/*}.

Finally, \sysname{} uses a corpus-derived skill grammar $\mathcal{G}_{\mathrm{skill}}$, distilled from numerous public qualified skills, to guide package organization.
The grammar supplies recurrent presentation patterns for descriptions, executable sequences or conditional structures, applicability conditions, safeguards, and progressive disclosure through package-relative references.
It affects how admitted content is rendered, but does not add new procedures or broaden the admitted scope.
(\textit{The complete skill grammar is provided in supplementary materials}.)

\section{Experiments}
\label{sec:exp}


\begin{table*}[!t]
\centering

\setlength{\tabcolsep}{1.9pt}
\renewcommand{\arraystretch}{0.98}
\resizebox{0.85\textwidth}{!}{%
\begin{tabular}{@{}ccccccccccccc@{}}
\toprule
\textbf{Agent} &
\textbf{Model} &
\textbf{Skill Setting} &
\shortstack{\textbf{Overall}\\($n=87$)} &
\shortstack{\textbf{$\Delta$}\\\textbf{(pp)}} &
\shortstack{\textbf{Soft.}\\($n=16$)} &
\shortstack{\textbf{Office}\\($n=14$)} &
\shortstack{\textbf{Sci.}\\($n=14$)} &
\shortstack{\textbf{Media}\\($n=5$)} &
\shortstack{\textbf{Cyber}\\($n=7$)} &
\shortstack{\textbf{Fin.}\\($n=9$)} &
\shortstack{\textbf{Ind.}\\($n=14$)} &
\shortstack{\textbf{Math}\\($n=8$)} \\
\midrule

Claude Code & DeepSeek-V4-Pro & No Skill
& 23.4 & -- & 22.5 & 32.9 & 18.6 & 20.0 & 14.3 & 26.7 & 25.7 & 20.0 \\


Claude Code & DeepSeek-V4-Pro & Anthropic Skill-Creator
& 31.7 & +8.3 & 35.0 & 35.7 & 32.9 & 44.0 & 22.9 & 13.3 & 34.3 & 32.5 \\


Claude Code & DeepSeek-V4-Pro & OpenAI Skill-Creator
& 33.3 & +9.9 & 33.8 & 41.4 & 44.3 & 32.0 & 22.9 & 17.8 & 28.6 & 35.0 \\

Claude Code & DeepSeek-V4-Pro & Human-Curated Skill
& 51.3 & +27.9 & 43.8 & 47.1 & \textbf{72.9} & \textbf{80.0} & 42.9 & 40.0 & \textbf{44.3} & \textbf{50.0} \\

Claude Code & DeepSeek-V4-Pro & OpenSkill
& 42.3 & +18.9 & 35.0 & 38.6 & 61.4 & 68.0 & 37.1 & 28.9 & 37.1 & 42.5 \\

Claude Code & DeepSeek-V4-Pro & MUSE-Autoskill
& 43.2 & +19.8 & 37.5 & 38.6 & 61.4 & 64.0 & 40.0 & 31.1 & 37.1 & 45.0 \\

\noalign{\vskip 0.7pt}
\arrayrulecolor{black!32}\cdashline{1-13}[0.32pt/1.8pt]\arrayrulecolor{black}
\noalign{\vskip 2.2pt}
Claude Code & DeepSeek-V4-Pro & \textbf{\sysname}
& \textbf{54.7} & \textbf{+31.3} & \textbf{53.8} & \textbf{54.3} & 65.7 & 72.0 & \textbf{57.1} & \textbf{48.9} & \textbf{44.3} & \textbf{50.0} \\

\midrule

Claude Code & Opus 4.8 & No Skill
& 45.3 & -- & 56.3 & 55.7 & 57.1 & 32.0 & 57.1 & 15.6 & 32.9 & 37.5 \\


Claude Code & Opus 4.8 & Anthropic Skill-Creator
& 49.2 & +3.9 & 56.3 & 55.7 & 62.9 & 56.0 & 57.1 & 17.8 & 42.9 & 35.0 \\


Claude Code & Opus 4.8 & OpenAI Skill-Creator
& 49.9 & +4.6 & 55.0 & 57.1 & 65.7 & 52.0 & 54.3 & 17.8 & 45.7 & 37.5 \\

Claude Code & Opus 4.8 & Human-Curated Skill
& 59.5 & +14.2 & 57.5 & 68.6 & 75.7 & \textbf{80.0} & \textbf{60.0} & \textbf{51.1} & 45.7 & \textbf{40.0} \\

Claude Code & Opus 4.8 & OpenSkill
& 51.5 & +6.2 & 51.3 & 60.0 & 68.6 & 72.0 & 51.4 & 44.4 & 40.0 & 22.5 \\

Claude Code & Opus 4.8 & MUSE-Autoskill
& 53.3 & +8.0 & 51.3 & 64.3 & 68.6 & 76.0 & 57.1 & 44.4 & 41.4 & 25.0 \\

\noalign{\vskip 0.7pt}
\arrayrulecolor{black!32}\cdashline{1-13}[0.32pt/1.8pt]\arrayrulecolor{black}
\noalign{\vskip 2.2pt}
Claude Code & Opus 4.8 & \textbf{\sysname}
& \textbf{60.9} & \textbf{+15.6} & \textbf{63.8} & \textbf{71.4} & \textbf{81.4} & 64.0 & 57.1 & 48.9 & \textbf{47.1} & \textbf{40.0} \\

\midrule

Codex & DeepSeek-V4-Pro & No Skill
& 29.7 & -- & 42.5 & 35.7 & 28.6 & 8.0 & 22.9 & 28.9 & 21.4 & 30.0 \\


Codex & DeepSeek-V4-Pro & Anthropic Skill-Creator
& 32.9 & +3.2 & 45.0 & 38.6 & 24.3 & 20.0 & 40.0 & 20.0 & 30.0 & 35.0 \\


Codex & DeepSeek-V4-Pro & OpenAI Skill-Creator
& 37.0 & +7.3 & 42.5 & 45.7 & 42.9 & 12.0 & 42.9 & 24.4 & 30.0 & 37.5 \\

Codex & DeepSeek-V4-Pro & Human-Curated Skill
& \textbf{45.7} & \textbf{+16.0} & \textbf{52.5} & \textbf{52.9} & 41.4 & \textbf{52.0} & \textbf{62.9} & 28.9 & 32.9 & \textbf{50.0} \\

Codex & DeepSeek-V4-Pro & OpenSkill
& 40.7 & +11.0 & 47.5 & 45.7 & 37.1 & 48.0 & 54.3 & 26.7 & 30.0 & 42.5 \\

Codex & DeepSeek-V4-Pro & MUSE-Autoskill
& 40.2 & +10.5 & 43.8 & 48.6 & 35.7 & 48.0 & 60.0 & 24.4 & 28.6 & 42.5 \\

\noalign{\vskip 0.7pt}
\arrayrulecolor{black!32}\cdashline{1-13}[0.32pt/1.8pt]\arrayrulecolor{black}
\noalign{\vskip 2.2pt}
Codex & DeepSeek-V4-Pro & \textbf{\sysname}
& 43.9 & +14.2 & 46.2 & 48.6 & \textbf{47.1} & 36.0 & 48.6 & \textbf{42.2} & \textbf{34.3} & 45.0 \\

\midrule

Codex & GPT-5.5 & No Skill
& 45.1 & -- & 57.5 & 58.6 & 48.6 & 20.0 & 51.4 & 11.1 & 34.3 & 57.5 \\


Codex & GPT-5.5 & Anthropic Skill-Creator
& 48.5 & +3.4 & 58.8 & 60.0 & 51.4 & 20.0 & 51.4 & 17.8 & 48.6 & 52.5 \\


Codex & GPT-5.5 & OpenAI Skill-Creator
& 48.5 & +3.4 & 60.0 & 62.9 & 48.6 & 28.0 & 51.4 & 20.0 & 42.9 & 52.5 \\

Codex & GPT-5.5 & Human-Curated Skill
& 60.9 & +15.8 & 60.0 & 62.9 & \textbf{75.7} & \textbf{60.0} & 65.7 & 33.3 & \textbf{62.9} & 57.5 \\

Codex & GPT-5.5 & OpenSkill
& 49.4 & +4.3 & 50.0 & 48.6 & 64.3 & 44.0 & 57.1 & 28.9 & 47.1 & 47.5 \\

Codex & GPT-5.5 & MUSE-Autoskill
& 52.0 & +6.9 & 53.8 & \textbf{70.0} & 54.3 & 24.0 & 65.7 & 17.8 & 45.7 & 67.5 \\

\noalign{\vskip 0.7pt}
\arrayrulecolor{black!32}\cdashline{1-13}[0.32pt/1.8pt]\arrayrulecolor{black}
\noalign{\vskip 2.2pt}
Codex & GPT-5.5 & \textbf{\sysname}
& \textbf{63.7} & \textbf{+18.6} & \textbf{66.3} & \textbf{70.0} & 74.3 & 52.0 & \textbf{71.4} & \textbf{37.8} & 57.1 & \textbf{70.0} \\

\bottomrule
\end{tabular}
}
\caption{\small \textbf{Main results on SkillsBench.}
$n$ is the task number and
$\Delta=p_{\mathrm{skill}}-p_{\mathrm{no\text{-}skill}}$
reports the difference from no-skill in percentage points.}
\label{tab:main-results}
\vspace{-1em}
\end{table*}

\subsection{Experimental Setup}
\label{sec:exp-setup}

\fakeparagraph{Evaluation benchmark}
We follow \textsc{SkillsBench} v1.1~\citep{skillsbench} and evaluate all 87 tasks across 8 domains.
We report avg@5 pass rate, computed as the mean verified success rate over 5 independent runs for each task.
Domain scores are averaged over tasks of each domain, while the overall score is averaged over all tasks.


\fakeparagraph{Compared baselines}
We compare \sysname{} against six baselines, covering a no-skill setting, a human-authored reference, and four automated skill-construction methods. 
More implementation and configuration details are provided in the supplementary material.
\begin{itemize}
    \item \textbf{No Skill}.
    The agent uses only task descriptions and visible context, without installed skills or task-specific procedural guidance.

    \item \textbf{Human-Curated Skill}~\citep{skillsbench}.
    The agent uses the original human-authored task-specific skills released with \textsc{SkillsBench} v1.1 without any modification.
    
    \item \textbf{Anthropic Skill-Creator} \citep{anthropic-skill-creator}.
    It an official skill creator released by Anthropic, which drafts skill instructions, evaluates them on representative prompts, and iteratively revises the skill based on evaluation feedback.
    
    \item  \textbf{OpenAI Skill-Creator} \citep{openai-skill-creator}.
    It an official skill creator from OpenAI, which scaffolds a modular skill package, adds relevant resources, and validates the package before evaluation.

    \item \textbf{OpenSkill} \citep{openskill}.
    It retrieves open-world knowledge and verification anchors, synthesizes a skill, and refines it against the self-constructed virtual tasks.
    
    \item \textbf{MUSE-Autoskill} \citep{muse-autoskill}.
    It distills task-solving experience into reusable procedures, validation steps, and common failure modes for reliable downstream execution.
\end{itemize}

\fakeparagraph{Configurations}
To provide a fair comparison, we also enable web access for skill creators from Anthropic and OpenAI, as with other baselines.
We evaluate four configurations spanning two agent runtimes and three models:
Claude Code~\citep{anthropic-claude-code} with DeepSeek-V4-Pro~\citep{deepseek-v4} and Claude Opus 4.8~\citep{anthropic-claude-opus-4-8}, and Codex~\citep{openai-codex} with DeepSeek-V4-Pro and
GPT-5.5~\citep{openai-gpt-5-5}.
More implementation details about the evaluation protocol are provided in supplementary materials.

\subsection{Main Results}
\label{sec:main-results}

\tabref{tab:main-results} reports the complete overall performance and the domain-level results across all four agent-model configurations.

\fakeparagraph{Overall Performance}
\sysname{} achieves the highest overall performance in three of the four agent-model configurations.
This exceeds no-skill execution by 19.9 percentage points and the strongest automated baseline, MUSE-Autoskill, by 8.6 percentage points.
At the aggregate level, \sysname{} reaches an observed avg@5 of 55.8\%,
1.5 percentage points above the Human-Curated Skill.
We also report 95\% confidence intervals for both conditions in \S\ref{sec:more_exps}.


\fakeparagraph{Domain-Level Analysis}
\figref {fig:main-results} reveals a clear domain-level asymmetry between \sysname {} and the human-curated, with configurations weighted equally within each domain.
\sysname {} performs better in Finance and Economics, Software Engineering, and Office Tasks, whereas Media is the only domain with a substantial deficit.
A task-level examination of all Media tasks identifies a recurring content gap: \textit {the generated skills capture the overall solution procedure but less consistently preserve the specific steps and calibrated parameter choices used at failure-prone stages}.
This refines the task-level variability reported in prior work by identifying execution-critical detail preservation as a plausible source of the remaining gap~\citep {openskill,muse-autoskill, skillevolver}.

\begin{figure}[htb]
\centering
\includegraphics[width=0.85\columnwidth]{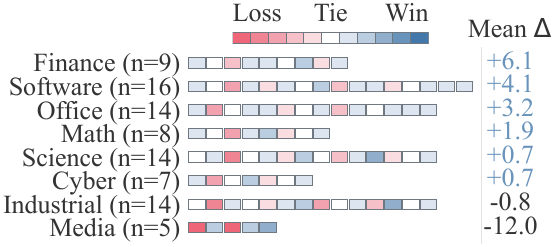}
\caption{\small \textbf{Task-level comparison with Human-Curated Skills.}
Each tile represents one task and is grouped by the sign of its avg@5 difference after averaging equally over the four agent--model configurations.
Color intensity encodes the absolute difference, and the rightmost column reports the category mean in percentage points.
}
\label{fig:main-results}
\end{figure}

\subsection{Evaluation Diagnostics} \label{sec:more_exps}
\fakeparagraph{Statistical Uncertainty} \tabref{tab:pooled-confidence} pools 1,740 binary evaluations per skill setting across 87 tasks, five runs, and four agent--model configurations.
\sysname{} achieves the highest observed aggregate avg@5 at 55.8\%, compared with 54.4\% for the Human-Curated Skill, 47.2\% for MUSE-Autoskill, and 46.0\% for OpenSkill.
Its 95\% Wilson interval is [53.5, 58.1], versus [52.0, 56.7] for the Human-Curated Skill.
These intervals quantify pooled within-setting uncertainty rather than pairwise superiority.
Accordingly, the observed 1.4-point margin over the Human-Curated Skill is interpreted descriptively.


\begin{table}[thb]
\centering
\footnotesize
\setlength{\tabcolsep}{3.0pt}
\renewcommand{\arraystretch}{1.08}
\resizebox{0.9\linewidth}{!}{
    \begin{tabular}{@{}lccc@{}}
    \toprule
    Skill Setting & Passes / Trials & avg@5 & 95\% CI \\
    \midrule
    No Skill & 624 / 1740 & 35.9 & [33.6, 38.1] \\
    Anthropic Skill-Creator & 706 / 1740 & 40.6 & [38.3, 42.9] \\
    OpenAI Skill-Creator & 734 / 1740 & 42.2 & [39.9, 44.5] \\
    Human-Curated Skill & 946 / 1740 & 54.4 & [52.0, 56.7] \\
    OpenSkill & 800 / 1740 & 46.0 & [43.6, 48.3] \\
    MUSE-Autoskill & 821 / 1740 & 47.2 & [44.8, 49.5] \\
    \sysname{} & 971 / 1740 & \textbf{55.8} & [53.5, 58.1] \\
    \bottomrule
    \end{tabular}
}
\caption{\small \textbf{Combined results on four agent-model configurations.}
We combine all runs from the configurations, giving 1,740 binary outcomes per condition (over 87 tasks $\times$ 5 runs $\times$ 4 configurations).
The 95\% CI denotes the 95\% Wilson Confidence Intervals.}
\label{tab:pooled-confidence}
\vspace{-1em}
\end{table}

\begin{table}[thb]
\centering
\small
\setlength{\tabcolsep}{1pt}
\renewcommand{\arraystretch}{1.15}
\resizebox{0.85\linewidth}{!}{
    \begin{tabular*}{\columnwidth}{@{\extracolsep{\fill}}lcccc@{}}
    \toprule
    & \multicolumn{2}{c}{Creation}
    & \multicolumn{2}{c}{Execution} \\
    \cmidrule(lr){2-3}\cmidrule(lr){4-5}
    Skill Setting
    & \shortstack{Tok./Call\\(K)}
    & \shortstack{Time/Task\\(min)}
    & \shortstack{Tok./Run\\(K)}
    & \shortstack{Time/Run\\(min)} \\
    \midrule
    No Skill                & N/A  & N/A   & 661 & 6.44 \\
    Anthropic Skill-Creator & 59.4 & 6.47  & 612 & 5.79 \\
    OpenAI Skill-Creator    & 58.7 & 6.88  & 583 & 5.61 \\
    Human-Curated Skill     & N/A  & N/A   & 716 & 6.69 \\
    OpenSkill               & 65.9 & 36.37 & 694 & 6.78 \\
    MUSE-Autoskill          & 68.3 & 35.21 & 578 & 6.45 \\
    \sysname{}              & 69.1 & 23.21 & 709 & 6.39 \\
    \bottomrule
    \end{tabular*}
}
\caption{\small \textbf{Skill creation and downstream execution resource use.}
Creation token is averaged per LLM call, the execution-token usage is per
evaluation run and time is end-to-end per task or run.}
\label{tab:cost-budget}
\vspace{-1em}
\end{table}

\fakeparagraph{Creation and Execution Cost} 
\tabref{tab:cost-budget} reports costs for the Codex--GPT-5.5 configuration.
Across all automated skill creation methods, average creation-token usage per LLM call is similar, ranging from 58.7K to 69.1K.
\sysname{} requires 23.21 minutes per task, less than OpenSkill and MUSE-Autoskill (35.21--36.37 minutes) but more than the two Skill-Creator baselines (6.47--6.88 minutes).
Its parallel research subagents reduce end-to-end creation latency. During downstream execution, \sysname{} uses 709K tokens and 6.39 minutes per run, with latency comparable to the other methods despite moderately higher token usage.
The supplementary materials further compare skill-package length and file composition.



\begin{figure}[thb]
\centering
\includegraphics[width=\columnwidth]{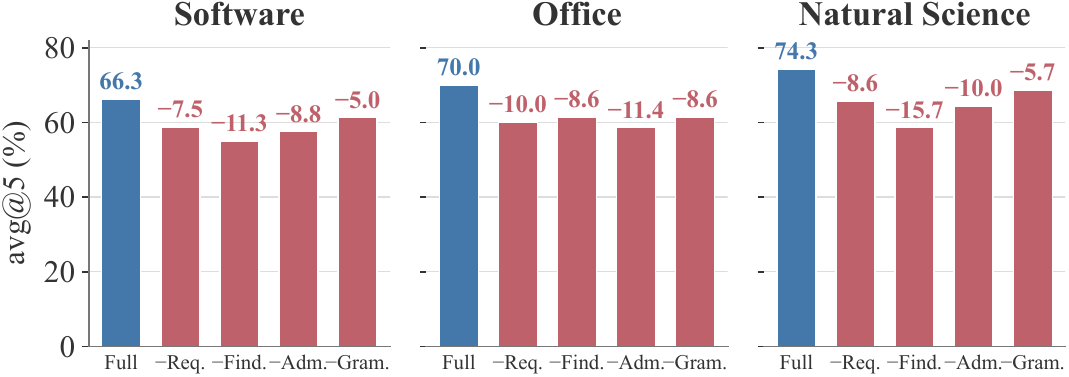}
\caption{\small
\textbf{Component ablation across three \textsc{SkillsBench} domains.}
Bars report avg@5 for the full method and variants without implicit requirement discovery (\textsc{Req.}), structured findings (\textsc{Find.}), procedure admission (\textsc{Adm.}), or the skill grammar (\textsc{Gram.}).
Labels report within-domain decrease in pp relative to the full method.
}
\label{fig:ablation-results}
\vspace{-1em}
\end{figure}

\subsection{Ablation Study} \label{sec:component-analysis}
We evaluate four one-component ablations on the Software Engineering, Office, and Natural Science domains of \textsc{SkillsBench} v1.1 using Codex with GPT-5.5.
The variants remove the key component of each stage, including implicit requirement discovery, structured findings, procedure admission, or grammar-guided rendering, respectively.
All conditions use the same source-access scope and creation budget.
For each task and condition, we create one skill and evaluate it over five independent execution runs.
As shown in \figref{fig:ablation-results}, every ablation reduces avg@5 in all three domains, with observed drops of 5.0--15.7 percentage points. Removing structured findings causes the largest decrease in Software Engineering and Natural Science, while removing procedure admission has the largest effect in Office.
Grammar-guided rendering yields smaller but consistent gains across the three domains.
Overall, the results suggest that requirement discovery, evidence structuring, scope-aware admission, and grammar-guided artifact organization provide complementary benefits.

\subsection{Robustness under Source Perturbations} \label{sec:source-robustness}
\fakeparagraph{Setups} 
Open-world sources may contain irrelevant noise, conflict procedures, or adversarially framed claims.
We conduct three perturbation tests on four tasks by adding one task-specific document to the otherwise identical creation context.
Specifically, \textit{(i)} \textit{Irrelevant} is topically related but does not support the target decision.
\textit{(ii)} \textit{Conflict} prescribes incompatible treatment under overlapping operating conditions.
\textit{(iii)} \textit{Adversarial} combines a misleading claim with directive-like language targeting the creator.
All other creation and evaluation settings remain fixed.
We create the skill for each (task, skill-creator, condition) tuple and run each skill five times downstream.
Thus, the source-propagation unit is the generated skill package ($n=4$ per method and perturbation) while five executions measure downstream variation.
An injected payload is counted as \textit{promoted} only when it is copied or semantically paraphrased as an affirmative runtime-facing instruction.
Quotations, warnings, explicit rejections, and conflict records do not count as promotion.

\fakeparagraph{Results of Perturbation} 
Conflict is the strongest perturbation for existing creators.
Across the four baselines, 9/16 conflicting payloads are promoted, compared with 5/16 irrelevant and 4/16 adversarial payloads, while their pooled pass count decreases from 58/80 under clear evidence to 35/80 under conflict.
\textsc{SkillAlchemy} {does not promote any of the 12 injected payloads and retains 17--18/20 downstream passes across all conditions}.
The single additional failure under conflict occurs without payload promotion and is therefore as  a downstream execution variation rather than evidence contamination.
This experiment evaluates the containment of pre-specified undesirable claims under exposure to perturbation sources.

\begin{table}[htb]

\resizebox{\linewidth}{!}{
\setlength{\tabcolsep}{5pt} 
\renewcommand{\arraystretch}{1.08}
\begin{tabular}{lccc|rrrr}
\toprule
\multirow{2}{*}{Skill Setting}
& \multicolumn{3}{c|}{Payload promotion $\downarrow$}
& \multicolumn{4}{c}{Downstream passes $\uparrow$} \\
\cmidrule(lr){2-4}\cmidrule(lr){5-8}
& Irr. & Conf. & Adv.
& Clean & Irr. & Conf. & Adv. \\
\midrule
Anthropic Skill-Creator
& 1/4 & 3/4 & 1/4
& 14/20 & 11/20 & 6/20 & 13/20 \\
OpenAI Skill-Creator
& 1/4 & 2/4 & 1/4
& 13/20 & 12/20 & 8/20 & 12/20 \\
OpenSkill
& 1/4 & 3/4 & 1/4
& 16/20 & 15/20 & 11/20 & 15/20 \\
MUSE-Autoskill
& 2/4 & 1/4 & 1/4
& 15/20 & 14/20 & 10/20 & 15/20 \\
\midrule
\textbf{SkillAlchemy}
& \textbf{0/4} & \textbf{0/4} & \textbf{0/4}
& \textbf{18/20} & \textbf{18/20}
& \textbf{17/20} & \textbf{18/20} \\
\bottomrule
\end{tabular}
} 
\centering
\caption{\small \textbf{Robustness test under three-type source perturbations}. 
Promotion reports created skills where injected payload be a runtime-facing instruction.
Passes report successful executions of all $20$ runs.}
\label{tab:source-robustness}

\end{table}

\subsection{Case Study: Reusable Skill Artifacts}
\label{sec:reuse-case}

We examine skill reusability from two complementary perspectives:
whether artifacts encode operations at a reusable level, and whether a
skill created for one task transfers to related tasks without revision.

\fakeparagraph{Matched Artifact Audit}
We audit a shared PDF-redaction operation: permanently removing sensitive
text while optionally preserving an allowed fragment.
\tabref{tab:reuse-case} reports the shortest semantically complete instruction unit, with boilerplate compressed but scope preserved.
Human-curated and generated skills often mix reusable principles with task-specific examples or execution templates.
\sysname{} more clearly separates decision logic from runtime facts by binding the current requirement to observable PDF structure, selecting a supported operation, and rejecting unjustified execution.

\fakeparagraph{Frozen-Skill Reuse}
We evaluate unchanged skill reuse on the \texttt{threejs-to-obj} task
family.
The original task exports a \texttt{Three.js} hierarchy as an OBJ;
\emph{filtered-export} adds ancestor-aware exclusion, while \emph{semantic-parts} assigns each geometry to its nearest semantic owner
and emits per-part files with a manifest.
Both variants are evaluated on held-out scenes.
For each condition, the skill is created or selected only for the original task, frozen, and reused unchanged.
Across five Codex--GPT-5.5 runs, \sysname{} achieves $5/5$ on the original task and $4/5$ on both variants, yielding the highest observed
score under each requirement shift.
Its cumulative degradation is only $-2$, compared with $-3$ for No Skill and between $-2$ and $-5$ for the other skill conditions.
These results provide controlled evidence that \sysname{} transfers beyond its seed task while remaining robust to distinct requirement changes.

\begin{table}[t]
\centering
\scriptsize
\setlength{\tabcolsep}{1.6pt}
\renewcommand{\arraystretch}{0.92}

\begin{tabularx}{\columnwidth}{
@{}
l
>{\raggedright\arraybackslash}X
cccc
@{}
}
\toprule
\textbf{Setting}
&
\textbf{Representative Instruction}
&
\textbf{Orig.}
&
\textbf{Filt.}
&
\textbf{Sem.}
&
$\boldsymbol{\Delta}_{\mathrm{sum}}$
\\
\midrule

No Skill
&
---
&
5/5 & 4/5 & 3/5 & $-3$
\\

\mbox{Human-Curated}
&
Redact sensitive data (e.g., student IDs) and insert an allowed mask.
&
5/5 & 4/5 & 1/5 & $-5$
\\

\mbox{Ant. Skill-Creator}
&
Redact rather than cover text; replace content and validate.
&
4/5 & 3/5 & 3/5 & $-2$
\\

\mbox{OpenAI Skill-Creator}
&
Select redaction, execute the edit plan, and validate.
&
4/5 & 3/5 & 2/5 & $-3$
\\

OpenSkill
&
Classify the edit, redact the target, and verify the result.
&
4/5 & 3/5 & 3/5 & $-2$
\\

\mbox{MUSE-Autoskill}
&
Remove text-layer content and verify deletion.
&
5/5 & 2/5 & 3/5 & $-5$
\\

\midrule

\textbf{\sysname{}}
&
\textbf{Bind PDF evidence to edit operations without hard-coded
runtime facts.}
&
\textbf{5/5}
&
\textbf{4/5}
&
\textbf{4/5}
&
$\mathbf{-2}$
\\

\bottomrule
\end{tabularx}

\caption{\small
\textbf{Artifact scope audit and frozen-skill reuse.}
Representative instructions summarize the operational scope of matched PDF-redaction artifacts, while scores report unchanged reuse on the \texttt{threejs-to-obj}.
Each skill is created for the original task, frozen, and evaluated on two variants.
$\Delta_{\mathrm{sum}} =(\mathrm{Filt.}-\mathrm{Orig.})
+(\mathrm{Sem.}-\mathrm{Orig.})$ reports cumulative change across the variants.
}
\label{tab:reuse-case}
\vspace{-1em}
\end{table}



\section{Conclusion}
\label{sec:conclusion}


We present \sysname{}, an admission-centered framework for open-world agent skill creation. 
\sysname{} addresses two central challenges in open-world skill creation: recovering behavior-changing requirements omitted by underspecified briefs and restricting each source-derived procedure to its evidence-supported scope. 
\sysname{} operationalizes this view through implicit requirement discovery, evidence-grounded procedure admission, and scope-preserving skill package compilation. 
Across 87 SkillsBench v1.1 tasks and four agent--model configurations, \sysname{} improves pass rate by 19.9pp over no-skill execution and by 8.6pp over the strongest automated baseline, while reaching aggregate performance comparable to human-curated skills. 
Overall, these results suggest that reliable skill creation should treat open-world knowledge as evidence to be admitted under explicit scope, rather than as instructions to be copied directly.


\bibliography{refs}

\clearpage
\appendix
\setcounter{table}{0}
\setcounter{figure}{0}
\setcounter{equation}{0}
\setcounter{principle}{0}
\setcounter{observation}{0}
\setcounter{proposition}{0}
\setcounter{procedure}{0}
\renewcommand{\thetable}{A.\arabic{table}}
\renewcommand{\thefigure}{A.\arabic{figure}}
\renewcommand{\theequation}{A.\arabic{equation}}
\renewcommand{\theprinciple}{A.\arabic{principle}}
\renewcommand{\theobservation}{A.\arabic{observation}}
\renewcommand{\theproposition}{A.\arabic{proposition}}

\fi
\end{document}